\documentclass[10pt,twocolumn,letterpaper]{article}
\usepackage{cvpr} 
\usepackage{cite}
\usepackage{amsmath,amssymb,amsfonts}
\usepackage{algorithmic}
\usepackage{graphicx}
\usepackage{multirow}
\usepackage{soul}
\usepackage{siunitx}
\usepackage{physunits}

\usepackage[colorlinks,bookmarksopen,bookmarksnumbered,citecolor=green,urlcolor=red]{hyperref}

\usepackage{makecell}
\usepackage{enumitem}
\usepackage{subcaption}
\usepackage[size=small]{caption}
\usepackage{booktabs}
\date{}
\usepackage{tabularx}
\usepackage{balance}

\title{\Large \bf
BlueME: Robust Underwater Robot-to-Robot Communication \\ Using Compact Magnetoelectric Antennas 
}

\author{Mehron Talebi, Sultan Mahmud, Adam Khalifa$^{\Join}$, and Md Jahidul Islam$^{\Join}$ \\
Department of ECE, University of Florida, USA. \\
\thanks{$^{\Join}$Represents equal contribution. {\small Corresponding Email: \\ \tt $\quad \text{ }$ $\quad \text{ }$ a.khalifa@ufl.edu, jahid@ece.ufl.edu}}
\thanks{This pre-print is accepted for publication at the IEEE Journal of Oceanic Engineering (JOE). DOI: 10.1109/JOE.2026.3675822.}
\vspace{-3mm}
}

\begin{document}

\maketitle

\begin{abstract}
We present the design, development, and experimental validation of \textbf{BlueME}, a compact magnetoelectric (ME) antenna array system for underwater robot-to-robot communication. BlueME employs ME antennas operating at their natural mechanical resonance frequency to efficiently transmit and receive very-low and low-frequency (VLF/LF) electromagnetic signals underwater. 
We detail the system’s design, simulation, fabrication, and integration onto low-power embedded platforms, emphasizing portability and scalability. 
For performance evaluation, we deployed BlueME on mobile robot platforms in freshwater (lake) and saltwater (ocean) trials. Our tests demonstrate reliable signal transmission and detection using the BlueME antenna system at distances beyond 700 meters while consuming less than 10 watts of power. Ocean trials demonstrate that the system operates effectively under challenging conditions, such as turbidity, obstacles, and multipath interference, which are factors that typically degrade acoustic and optical methods. 
Our analysis also examines the impact of complete submersion on system performance and identifies key deployment considerations. This work represents the first practical underwater deployment of ME antennas outside the laboratory and implements the largest VLF/LF ME array system to date. BlueME demonstrates significant potential for marine robotics and automation in multi-robot cooperative systems and remote sensor networks.

\vspace{2mm}
\end{abstract}
\vspace{-3mm}
\section{Introduction}
Underwater robot-to-robot communication is essential for ensuring the coordinated operation of autonomous underwater vehicles (AUVs) in a wide range of applications including environmental monitoring, search and rescue, and scientific expeditions~\cite{zhou2021survey,islam2021robot}. In subsea environments where human intervention is limited, AUVs rely on seamless communication to share information, synchronize tasks, and respond to dynamic changes in real time~\cite{jawhar2020secure,trawny_new,ghosh1992neural} in multi-robot missions or when communicating with surface vessels~\cite{irawan2019robot,abdullah2024ego2exo,gu2022communication}. This capability is particularly crucial in surveying or mapping deep-sea areas~\cite{trawny_new}, navigating overhead structures~\cite{abdullah2023caveseg}, and collecting samples from areas inaccessible to human divers~\cite{abdullah2024ego2exo}.

\begin{figure}[t]
    \centering
    \includegraphics[width=\columnwidth]{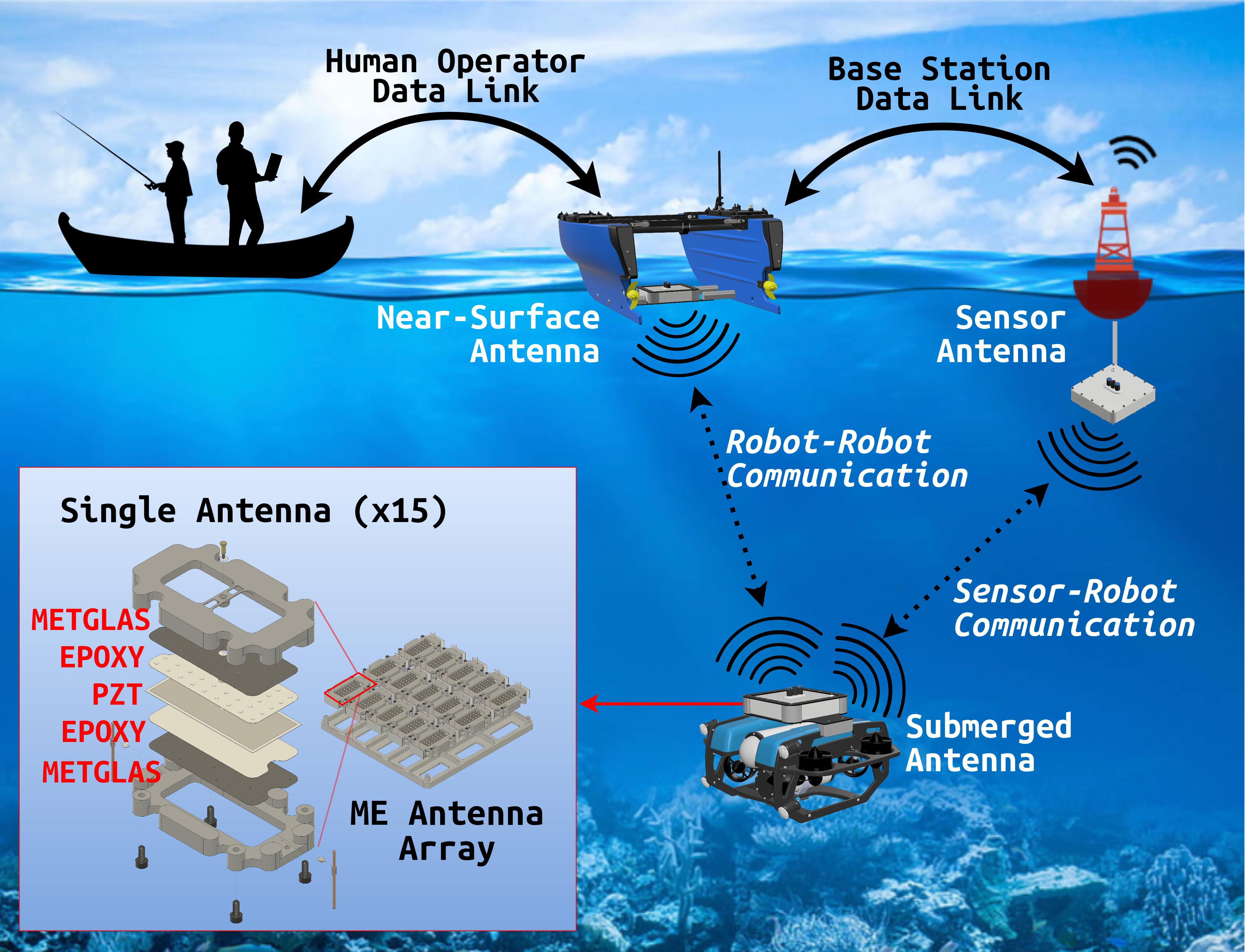}%
    \vspace{-2mm}
    \caption{The proposed BlueME system includes a novel ME antenna design; we use a $3$$\times$$5$ array of these antennas to enable real-time communications between underwater robots.}%
    \label{fig:big_picture}
    \vspace{-2mm}
\end{figure}

Conventional electromagnetic and radio frequency (RF) signals are severely attenuated underwater~\cite{zhou2021survey,jaafar2022overview}, limiting their effective communication range to only a few meters. While propagation improves at very-low and low-frequencies (VLFs/LFs)~\cite{dong2022vlf}, it typically comes at the cost of high transmission power and impractically large antenna sizes~\cite{slevin2021broadband,kumara2021underwater}. In seawater, electromagnetic signal attenuation ranges from approximately 1 to 10 dB per meter depending on both frequency and conductivity~\cite{moore_radio_1967}. This high attenuation severely limits the viability of RF communication for marine robotics applications~\cite{che_re-ev_new}.

Acoustic communication has been the most commonly used modality for underwater robot communications, offering omnidirectional transmission and acceptable signal attenuation over longer distances~\cite{min2019development,bourre2013robust}. Long-range low-frequency systems are used in ocean-based communication~\cite{freitag_basin-scale_2001}, while short-range high-frequency systems are preferred for robot positioning and data transfer~\cite{watanabe2005design,tuci2008evolving}. Acoustic communications suffer from Doppler effects, phase and magnitude fluctuations, and multi-path interference~\cite{kilfoyle_state_2000,llor2012underwater,stojanovic_underwater_2009}, compromising reliability and efficiency. Acoustic signals can be disruptive to marine ecosystems as well, limiting their long-term deployments. Optical communication systems (\eg, visible light, blue-green lasers) offer higher data rates~\cite{schirripa2020underwater,khalighi2014underwater,wang2016long}, but are limited by line-of-sight requirements. They are susceptible to biofouling~\cite{joslin2015demonstration}, a process where microorganisms and biological buildup degrade optical sensors and signal quality over time~\cite{sun2020review}.

Magnetoelectric (ME) antennas offer a promising solution to overcome the limitations of acoustic and optical communication in underwater environments~\cite{du2023very,dong2022vlf,lu_acoustically_2023,luo2024magnetoelectric,lv2021investigation}. By leveraging the magnetoelectric effect, these antennas efficiently transmit and receive electromagnetic signals with a more compact design than traditional electrically small antennas (ESAs)~\cite{lin_future_2018}. This efficiency, combined with their resilience to multi-path interference, non-line-of-sight requirements, and Doppler effects, makes ME antennas well-suited for space- and power-constrained underwater applications.

In this paper, we present the design, development, and validation of BlueME, a novel magnetoelectric (ME) antenna system for real-time data transmission between underwater robots. As shown in Fig.~\ref{fig:big_picture}, we use a compact array of ME antennas integrated into pressure-compensated enclosures for real-time data transmission and retrieval between robots and sensors deployed underwater.
While previous ME-based underwater communication efforts have primarily focused on basic connectivity tests in controlled water-tank environments~\cite{du2023very,xu2019low,xiang2016subsea,dong2022vlf}, we validate BlueME in a natural lake and ocean, demonstrating its potential as a deployable communication system for practical marine robotics applications.

In particular, we deploy the proposed system on ASVs (Autonomous Surface Vehicles) and underwater ROVs (Remotely Operated Vehicles) for comprehensive performance evaluation and feasibility analysis. Our field experimental results demonstrate that the BlueME system offers a more robust, long-range, scalable, and efficient alternative to the traditional acoustic and optical communication modalities~\cite{angara2024performance,stojanovic1994phase}. We also analyze the antenna characteristics, underlying theory of operations, and relevant engineering constructs for practical deployments of ME antenna arrays underwater. 

\vspace{1mm}
\noindent
Overall, we make the following contributions in this work:
\begin{enumerate}[label={$\arabic*$.},nolistsep,leftmargin=*]
    \item We design, simulate, and fabricate an ME antenna array system (BlueME) for real-time underwater communication between mobile robots and/or sensor nodes. This is the first ME antenna array system tested in practical environments beyond controlled water-tank setups.
    \item We present a seamless integration of BlueME into pressure-compensated enclosures to ensure reliable underwater operations in real-time. We demonstrate that the enclosed system is compact and end-to-end portable for use on low-power embedded devices. 
    \item We validate the system through simulation as well as field experimental trials in freshwater (lake) and saltwater (ocean) environments, demonstrating its effectiveness for underwater communication. The system performs reliably when fully submerged and is unaffected by turbidity, line-of-sight obstacles, and shallow-water interference, which limit acoustic and optical systems.
    \item Our field deployments reveal that we can achieve a high-fidelity communication link between mobile robots for up to $730$\,meters with a power footprint of only $1$-$10$\,watt. These capabilities demonstrate its potential for robust and low-latency communication underwater. 
\end{enumerate}

\section{Background \& Related Work}
\subsection{Underwater Communication Technologies}
Underwater communication is crucial for data exchange among autonomous underwater vehicles (AUVs), submarines, and subsea sensor networks~\cite{jaafar2022overview,min2019development,hao2023survery}. The most prevalent technology is acoustic communication, operating within the 10 Hz to 100 kHz frequency bands and supporting data rates up to several hundred Kbps. Despite its widespread use, acoustic communication faces significant limitations, including variable propagation delays and Doppler shifts due to the relatively slow speed of sound underwater (approximately 1500 m/s)~\cite{llor2012underwater, headrick2009growth}. Phase and magnitude fluctuations, along with multi-path interference, further degrade signal reliability\cite{kumara2021underwater, pompili2006deployment}. Additionally, acoustic signals can adversely affect marine life and are susceptible to environmental noise~\cite{song2019editorial,stojanovic1996recent}.

Optical communication offers higher data rates and lower latency compared to acoustic methods~\cite{schirripa2020underwater, khalighi2014underwater}. Notable systems include the BlueComm system, achieving 20\,Mbps over distances up to 200\,m~\cite{BlueComm}; the Ambalux system, providing 10 Mbps over 40\,m~\cite{khalighi2014underwater}; and LUMA modems, supporting 10\,Mbps across distances up to 50 m~\cite{Luma}. Underwater visible light communication (UVLC) has also been explored for extended ranges using wavelengths between 450--550\,nm, which experience lower attenuation in water~\cite{wang2016long, sun2020review}. However, optics are limited by line-of-sight requirements and are susceptible to scattering, absorption due to turbidity, and biofouling effects that degrade performance over time~\cite{gilbert1966underwater,neuner2019multi}.

Radio frequency (RF) electromagnetic communication underwater has historically been challenging due to high attenuation of EM waves in conductive media like seawater~\cite{moore_radio_1967}. Attenuation increases with frequency, limiting effective communication to short distances~\cite{che_re-ev_new}. However, RF communication offers advantages such as the ability to cross air-water and water-sediment boundaries, operation in turbid conditions, and non-line-of-sight capabilities~\cite{jaafar2022overview, wang_seawater_2019}. Recent research has revisited underwater RF communication using very low frequency (VLF, 3--30 kHz) and ultra-low frequency (ULF, 300--3000 Hz) bands to mitigate attenuation~\cite{du2023very, jaafar2022overview}. Despite these advances, practical deployment is hindered by the need for large antennas and/or high transmission power, making it less suitable for mobile underwater applications~\cite{dong2020portable, slevin2021broadband}.

\begin{figure}[t]
    \vspace{-2mm}
    \centering
    \includegraphics[width=\linewidth]{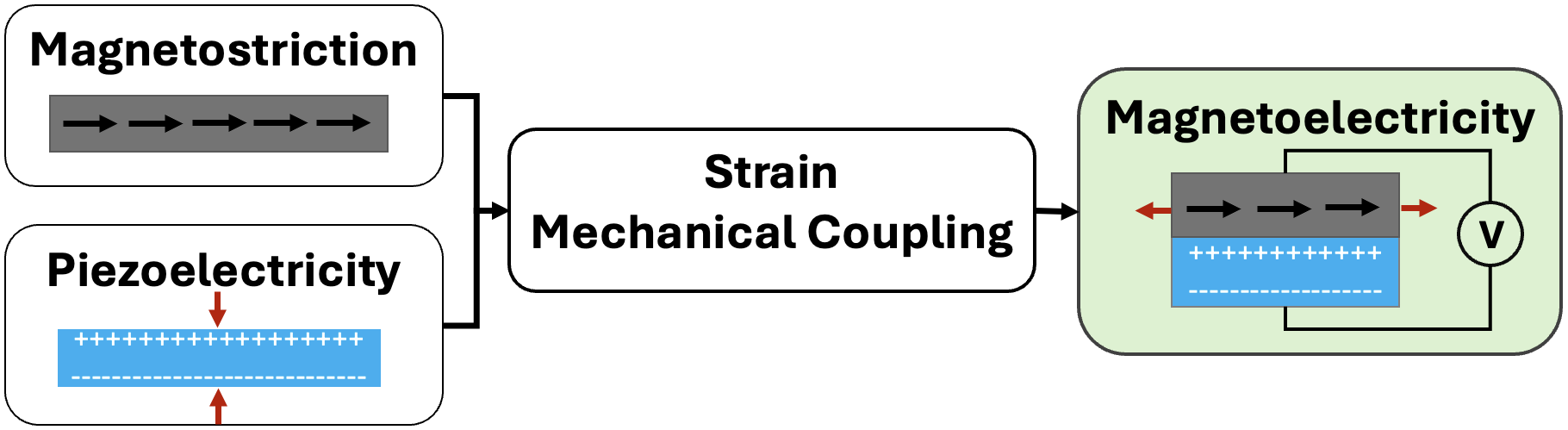}%
    \vspace{-1mm}
    \caption{The Magnetoelectric (ME) effect illustrating the coupling between magnetostrictive and piezoelectric layers. This mechanical coupling enables direct conversion between the magnetic and electric fields.}
    \label{fig:me_principle}
    \vspace{-3mm}
\end{figure}

\vspace{-1mm} 
\subsection{ME Antennas for Low-Power Communication} 
\vspace{-1mm}
ME antennas have emerged as a prospective solution for underwater communication at low frequencies with reduced antenna sizes~\cite{dong2022vlf}. ME antenna structures utilize the magnetoelectric effect, where a magnetic field induces strain in a magnetostrictive material, which is transferred to a piezoelectric material to generate an electric field, and vice versa~\cite{Wang2024,Fu2023}; see Fig.~\ref{fig:me_principle}. This mechanism enables efficient energy conversion in compact configurations, addressing the size limitations of traditional electrically small antennas (ESAs).

Research on ME communication systems has primarily focused on biomedical applications such as implantable devices~\cite{mukherjee2023miniaturized,Wu2021}. More recently, studies have explored their potential for underwater applications~\cite{du2023very,xu2019low,xiang2016subsea}. Fabrication typically involves layering magnetostrictive materials like Metglas with piezoelectric materials (\eg, PZT, PVDF), bonded using structural epoxy, and integrating conductive elements~\cite{yang2021progress}. ME antennas operating in VLF range can overcome some limitations of traditional RF communication by providing efficient transmission without requiring impractically large antennas. They are less susceptible to multi-path interference and can function in non-line-of-sight conditions.

However, significant challenges remain in designing and fabricating compact, low-power ME antennas suitable for integration with mobile robotic platforms~\cite{Fang2024,Leung2024}. Key issues include optimizing antenna performance, developing feasible deployment strategies, and characterizing communication properties in dynamic underwater conditions. This work aims to advance the field by developing a novel ME antenna array system tailored for marine robotics applications, addressing current gaps in practical deployment and performance evaluation.

\begin{figure}[h]
    \centering
    \includegraphics[width=0.85\columnwidth]{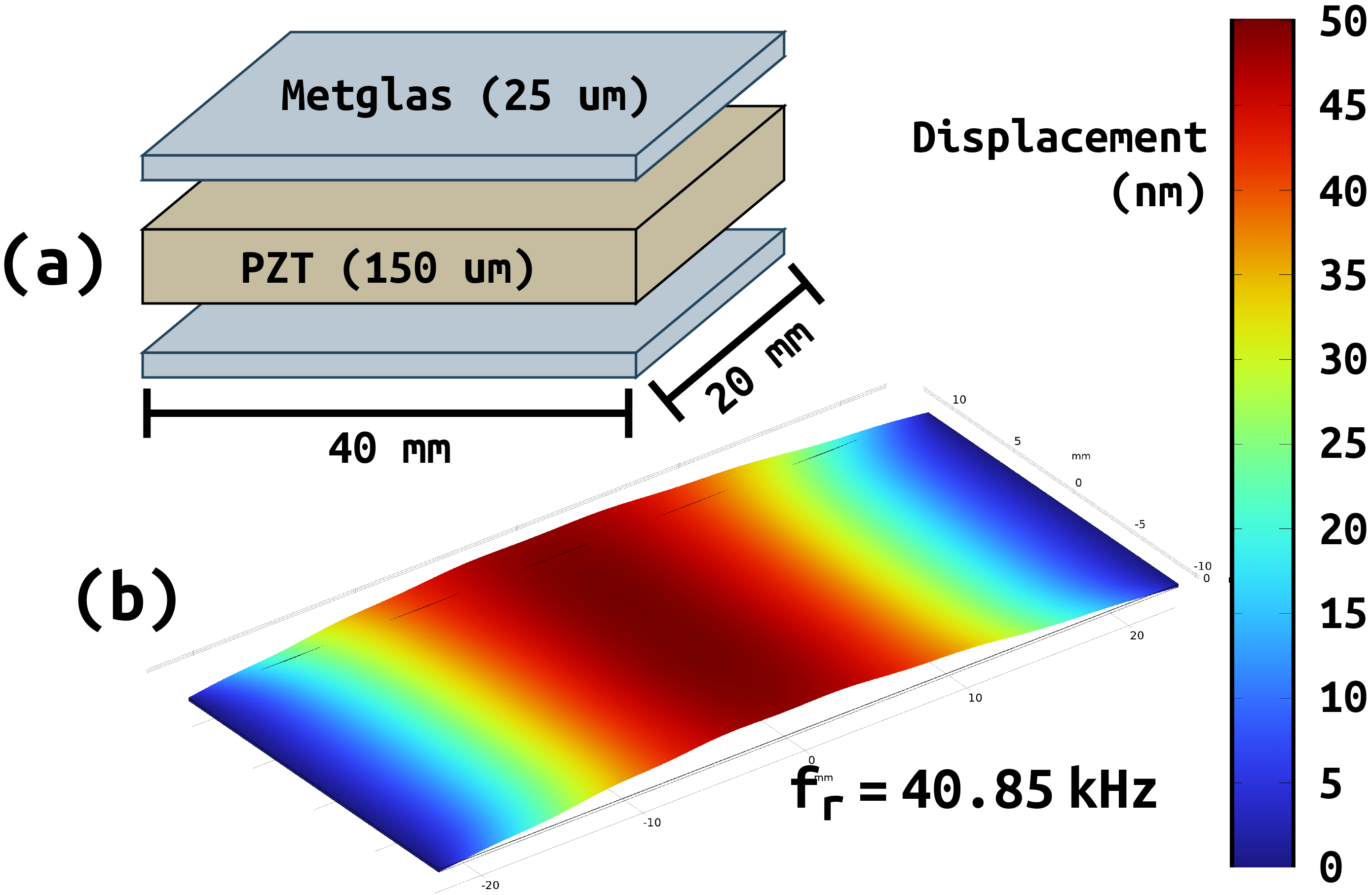}
    \includegraphics[width=0.82\columnwidth]{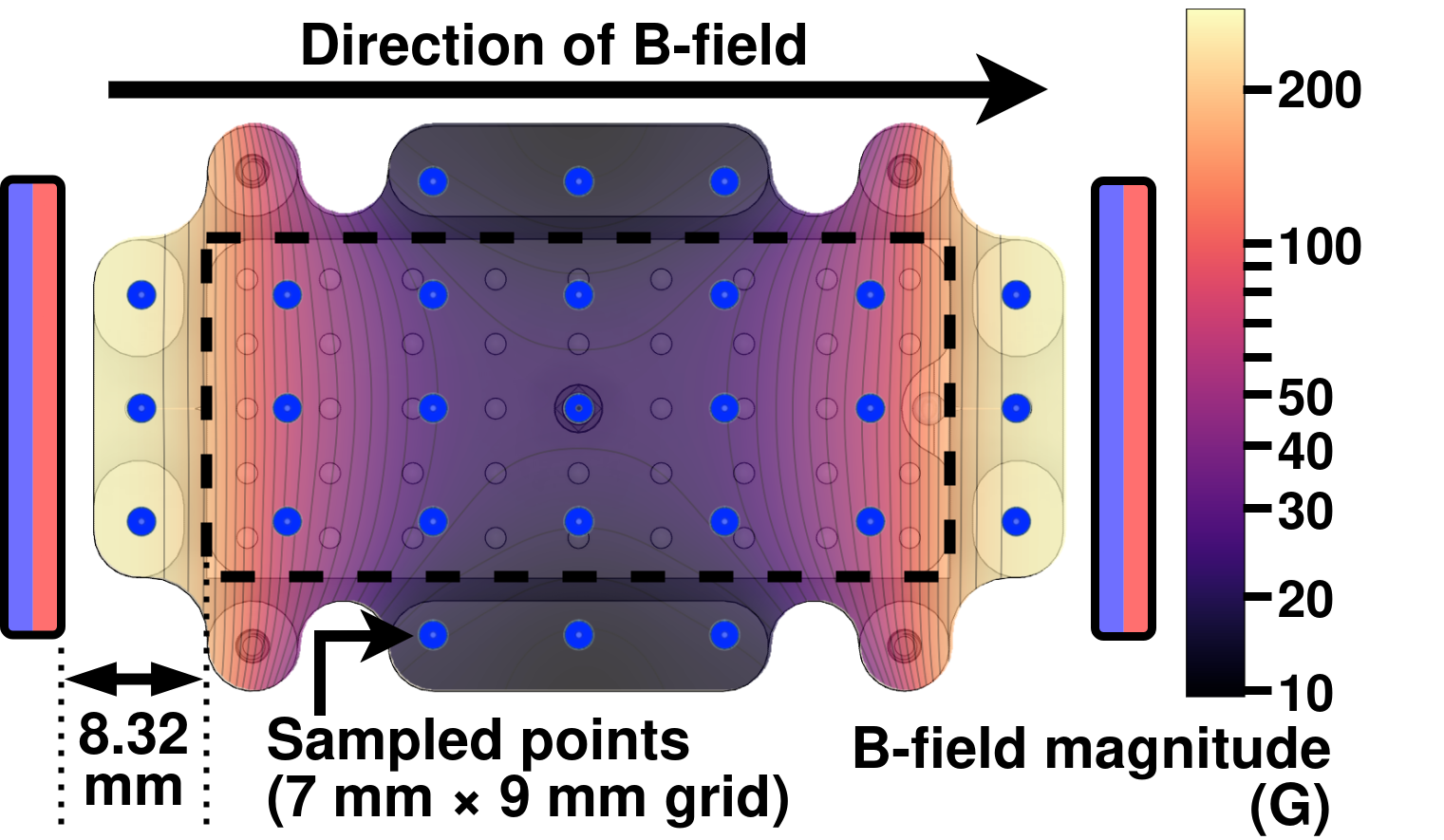}%
    \vspace{-1mm}
    \caption{(\textbf{Top}) Illustrations for (a) Cross-sectional schematic showing the three-layer structure: two \SI{25}{\micro\meter} Metglas layers with a \SI{150}{\micro\meter} PZT layer, dimensions $40\times20$\,mm$^2$; (b) COMSOL simulation showing displacement at the predicted fundamental resonance frequency of $40.85$\,kHz. (\textbf{Bottom}) Magnetic flux density across a single fabricated ME antenna (dashed outline) under applied bias. Measurements are taken at a uniform height, with the Gaussmeter positioned 1.8\,mm above the antenna surface.}
    \label{fig:simulation_view}
\end{figure}

\section{BlueME System Design}
\subsection{ME Antenna Design: Theory and Simulation}
\vspace{-1mm}
We first investigate the fundamental mechanical resonance frequency of the ME antenna for underwater communication. Considering the requirements of compact size, low attenuation characteristics, and long propagation distances, we calculate a frequency of approximately $35$\,kHz as follows~\cite{Popov2008}:
\begin{equation}
f_r = \frac{1}{2L} \sqrt{\frac{v_{\text{PZT}} \, Y_{\text{PZT}} + v_{\text{Metglas}} \, Y_{\text{Metglas}}}{v_{\text{PZT}} \,\rho_{\text{PZT}} + v_{\text{Metglas}} \, \rho_{\text{Metglas}}}}~.
\label{eq:1}
\end{equation}
Here, $L$=$45.7$\,mm is the sample length; $Y_{\text{PZT}}$=$51$\,GPa and $Y_{\text{Metglas}}$=$110$\,GPa are the Young's moduli~\cite{Popov2008}; $v_{\text{PZT}}$=$0.6$ and $v_{\text{Metglas}}$=$0.4$ are the volume fractions; $\rho_{\text{PZT}}$=$7800$\,Kg/m$^3$ and $\rho_{\text{Metglas}}$= $7180$\,Kg/m$^3$ are the densities of PZT and Metglas, respectively~\cite{metglas_inc_2605sa1,mide_technology_pzt5j}. In our design, PZT-5J is chosen as the \emph{piezoelectric} layer, and Metglas as the \emph{magnetostrictive} layer.

Fig.~\ref{fig:me_principle} illustrates an outline of our design showing how magnetostriction and piezoelectricity are coupled together in ME antennas. The ME effect is characterized by the magnetoelectric coupling coefficient $\alpha_{\text{ME}}$, defined as the induced electric field per unit applied magnetic field~\cite{Hosur2021, Karan2024}:
\begin{equation}
\alpha_{\text{ME}} = \frac{\partial E}{\partial H} = \frac{1}{t} \frac{\partial V}{\partial H},
\label{eq:alphaME}
\end{equation}
where $E$ is the electric field, $H$ is the magnetic field, $V$ is the voltage generated across the piezoelectric layer, and $t$ is its thickness. The ME coupling arises from the product of the magnetostrictive and piezoelectric effects, and our objective is to maximize the strain transfer between the magnetostrictive and piezoelectric layers to increase $\alpha_{\text{ME}}$.

To validate this theoretical analysis, we use COMSOL Multiphysics~\cite{Comsol} to simulate a three-layer ME structure. As shown in Fig.~\ref{fig:simulation_view}, we evaluate the displacement magnitude of our antenna with a Metglas-PZT-Metglas thickness of $25$-$150$-$25$\,$\mu$m. By simulation, we predict the eigenfrequency for the bias-free fundamental displacement mode to be $40.85$\,kHz.

\begin{figure}[t]
    \centering
    \includegraphics[width=\columnwidth]{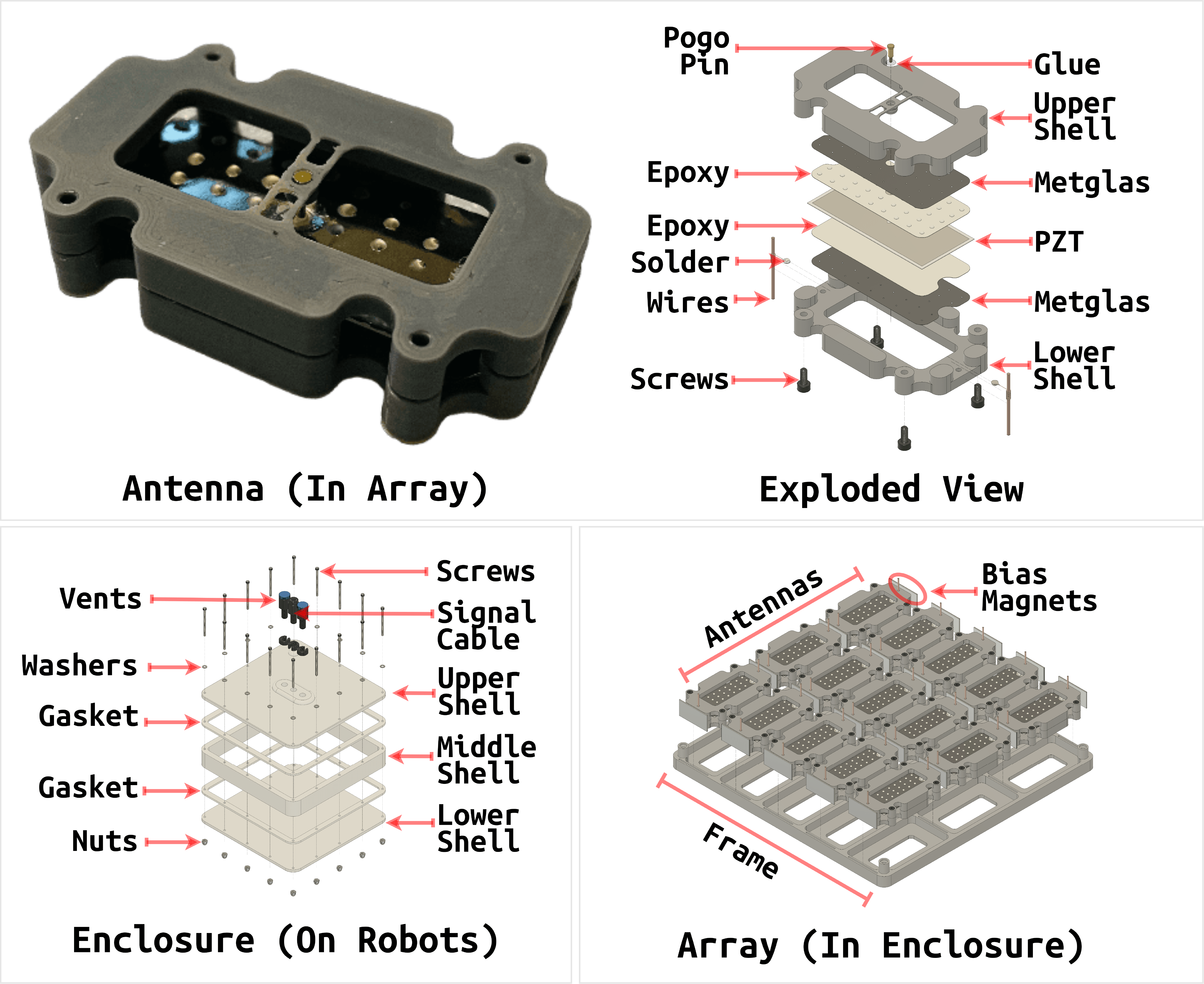}%
    \vspace{-1mm}
    \caption{A fabricated ME antenna and its components (top); the packaged enclosure of an array frame with $15$ antennas (bottom).}%
    \label{fig:packaging}
\end{figure}

\vspace{1mm}
\noindent 
\textbf{Antenna Enclosure and Array Assembly}. As shown in Fig.~\ref{fig:packaging}, each ME antenna is housed in a custom 3D-printed enclosure. The upper shell of the enclosure contains a pogo pin and two vertical $18$\,AWG copper wires; the pogo pin is bonded to a designated antenna hole using a UV-curable cyanoacrylate adhesive. The $36$\,AWG wires from the antenna are twisted and soldered to the $18$\,AWG wires, which extend through the lower shell. The two shells are then enclosed and arranged in a \(3 \times 5\) grid within a frame that includes twenty \(1 \times 0.5 \times \tfrac{1}{32}\,\mathrm{in}\) N52 neodymium magnets. Each magnet is positioned \SI{8.32}{\milli\meter} from the antenna edge, all oriented with the same north–south polarity. The magnetic bias distribution across a representative antenna was measured as shown earlier in Fig.~\ref{fig:simulation_view}. Finally, the receiver and transmitter antennas are connected in series and parallel configurations, respectively.

\vspace{-2mm}
\subsection{Pressure Compensation \& System Integration}
The antenna arrays and associated wiring are enclosed within a lidded resin-printed shell; see Fig.~\ref{fig:packaging}. A laser-cut silicone gasket ensures a watertight seal, secured by sixteen stainless steel screws torqued to $10$\,inch-lbs. The enclosure is then filled with a pressure-compensating oil to mitigate hydrostatic pressure effects. 

\begin{figure}[h]
    \vspace{-2mm}
    \centering
    \includegraphics[width=\linewidth]{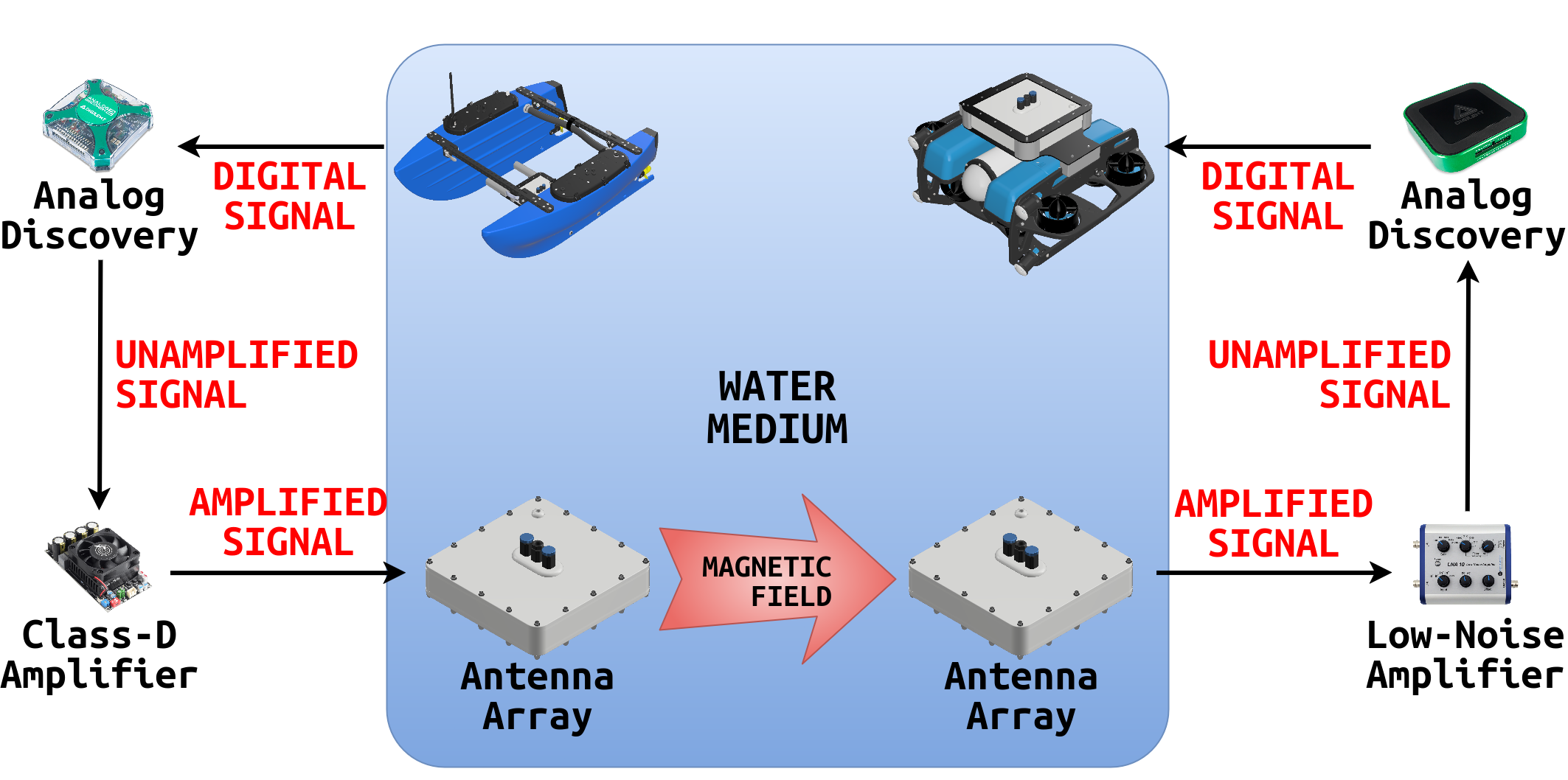}%
    \caption{Data communication and signal processing flow between a BlueME transmitter (ASV) and receiver (ROV) antenna arrays.}
    \label{fig:datacom}
\end{figure}

In our experiments, a transmitter antenna array is mounted on an autonomous surface vehicle (ASV) ({\tt BlueRobotics BlueBoat}), while a receiver array is mounted on an underwater remotely operated vehicle (ROV) ({\tt BlueRobotics BlueROV2}). As shown in Fig.~\ref{fig:datacom}, the transmitter array connects to a power amplifier driven by a signal from a Digilent Analog Discovery 2 (AD2) board.

The receiver array connects to a low-noise amplifier ({\tt AlphaLab LNA10}) with configured gain and filtering, which is interfaced with a Digilent Analog Discovery 3 (AD3) board. Signal modulation and recording are performed on host computers tethered to the ASV and ROV via AD2 and AD3 devices. We capture signal strength, represented by the induced voltage on the ME receiving array, as a function of distance. This data was recorded using Digilent's WaveForms software on the host computers.

\section{System Integration \& Experimental Setup}
\begin{figure}[t]
    \centering
    \includegraphics[width=0.5\textwidth]{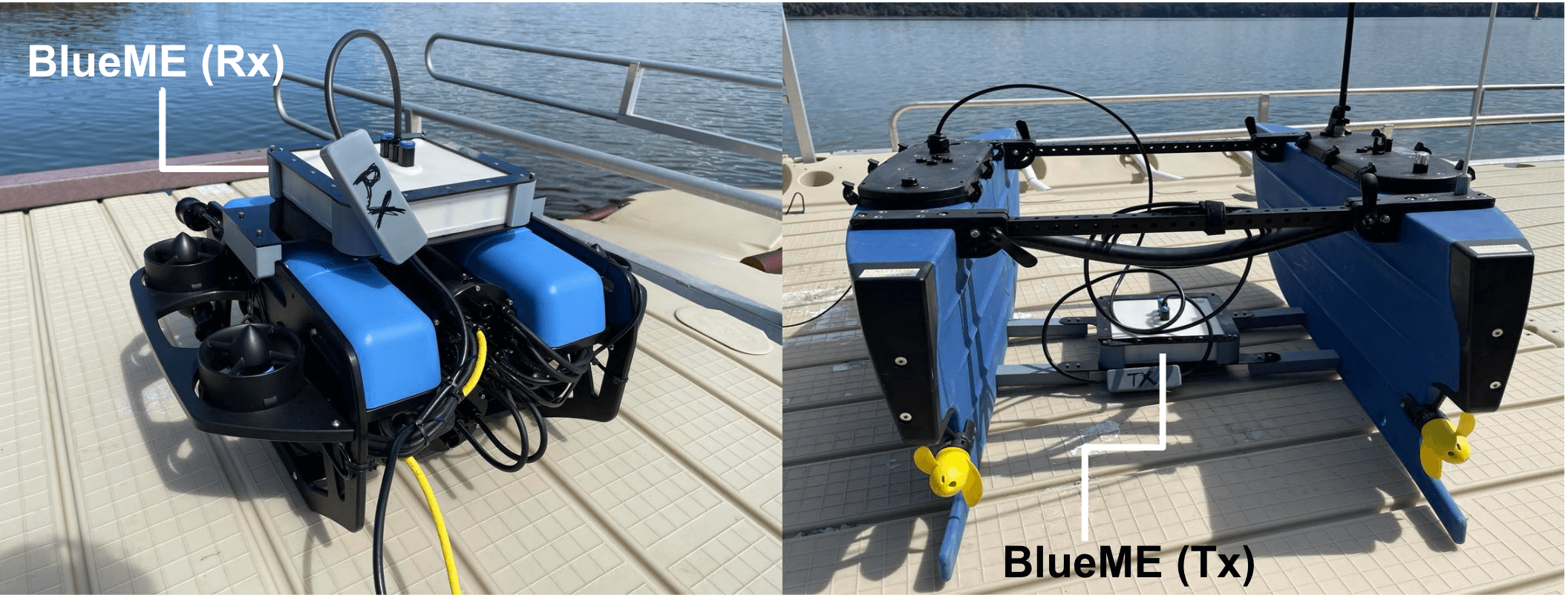} \\
    \includegraphics[width=0.5\textwidth]{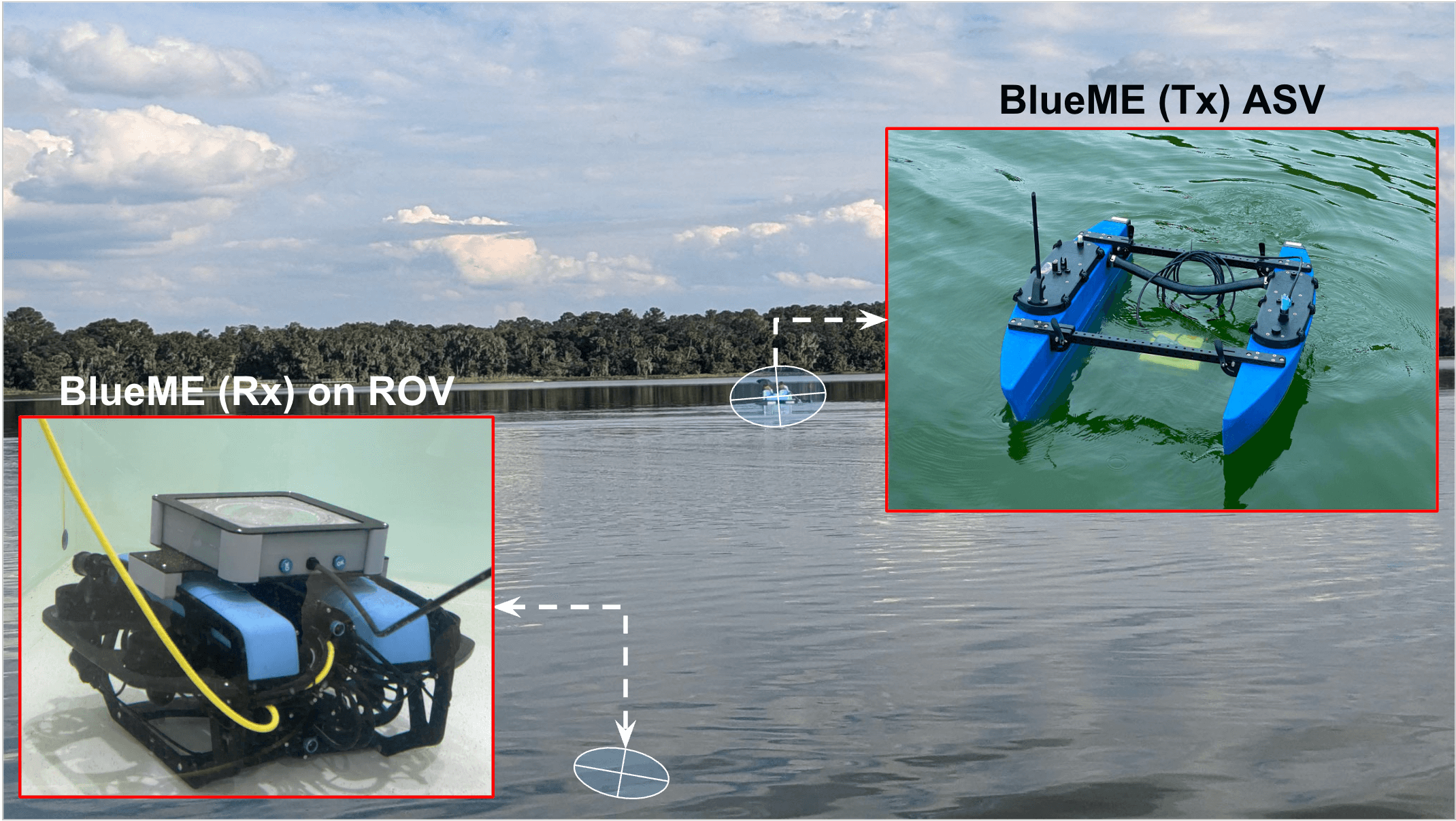} \\
    \vspace{-1mm}
    \text{(a) Freshwater experiments (lake trials)} \\
    \vspace{1mm}

    \includegraphics[width=0.5\textwidth]{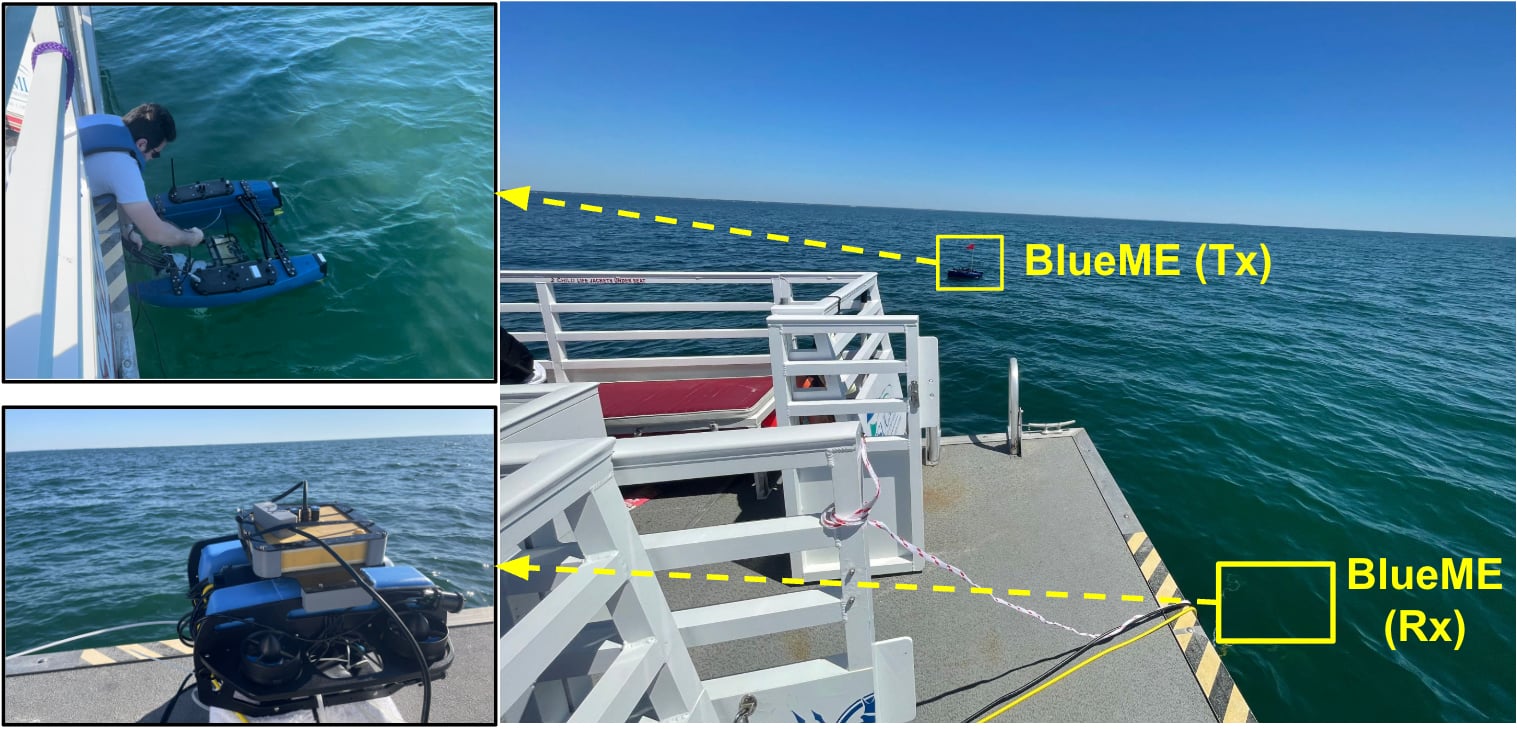} \\
    \vspace{-1mm}
    \text{(b) Saltwater experiments (ocean trials)}
    \caption{Our field experimental setup includes a BlueME transmitter and receiver deployed on an ASV and an underwater ROV for a series of robot-to-robot communication trials.}%
    \label{fig:field_setup}
    \vspace{-3mm}
\end{figure}

Our experimental analyses include (\textbf{i}) evaluation of the BlueME system's magnetoelectric (ME) antenna communication characteristics, and (\textbf{ii}) performance assessment through robotics field trials. First, we validate the proposed antenna's functionality using empirical data and simulation results, providing a basis for further investigation into optimal modulation techniques. We then integrate two BlueME systems on an autonomous surface vehicle (ASV) and an underwater remotely operated vehicle (ROV) to evaluate communication range and robustness, as illustrated in Fig.~\ref{fig:field_setup}. To this end, we conduct a series of field experimental trials in open water to demonstrate the feasibility of the proposed BlueME system for robot-to-robot communication.

\vspace{-1mm}
\subsection{Setup: Measuring Antenna Impedance}
\vspace{-1mm}
Before assembling the full antenna arrays, each ME antenna was tested individually to determine its resonant frequency and impedance characteristics. The objective was to verify that the empirical measurements matched our simulation results and to assess the combined response of the antennas when configured in arrays. Each antenna was placed in a shallow plastic vat filled with oil to replicate the dielectric environment in the enclosure. The antennas were then positioned within the array frame using bias magnets to ensure that the magnetic field conditions matched the assembled array, as their resonant response depends on the magnetic bias levels, which can vary due to the arrangement of magnets in the array frame.

\begin{figure*}[t]
    \vspace{-2mm}
    \centering
    \includegraphics[width=0.325\linewidth]{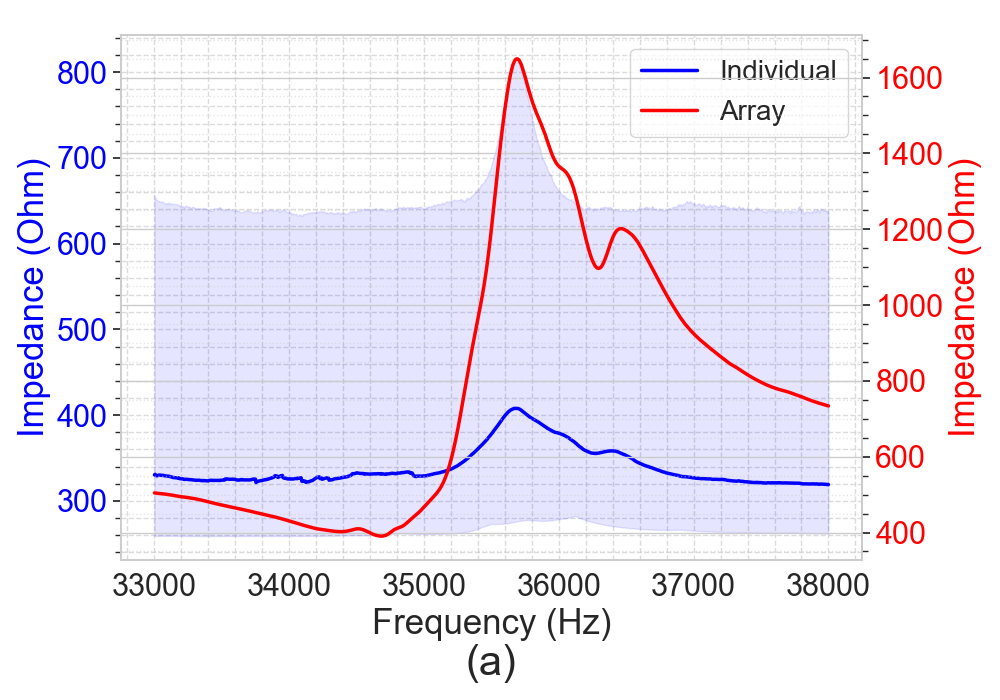}
    \includegraphics[width=0.325\linewidth]{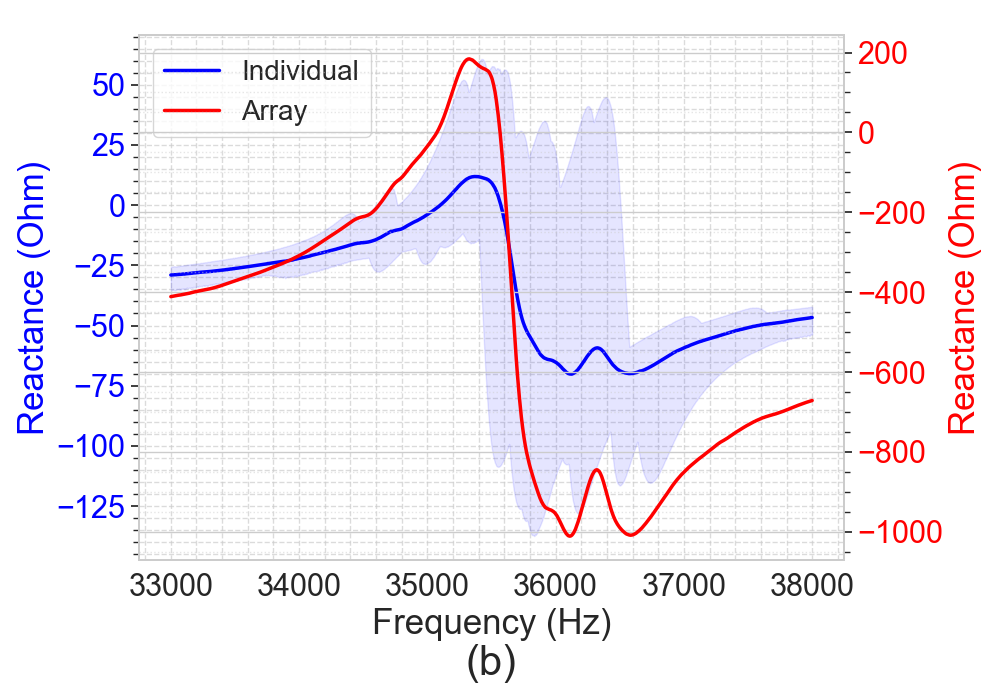}
    \includegraphics[width=0.325\linewidth]{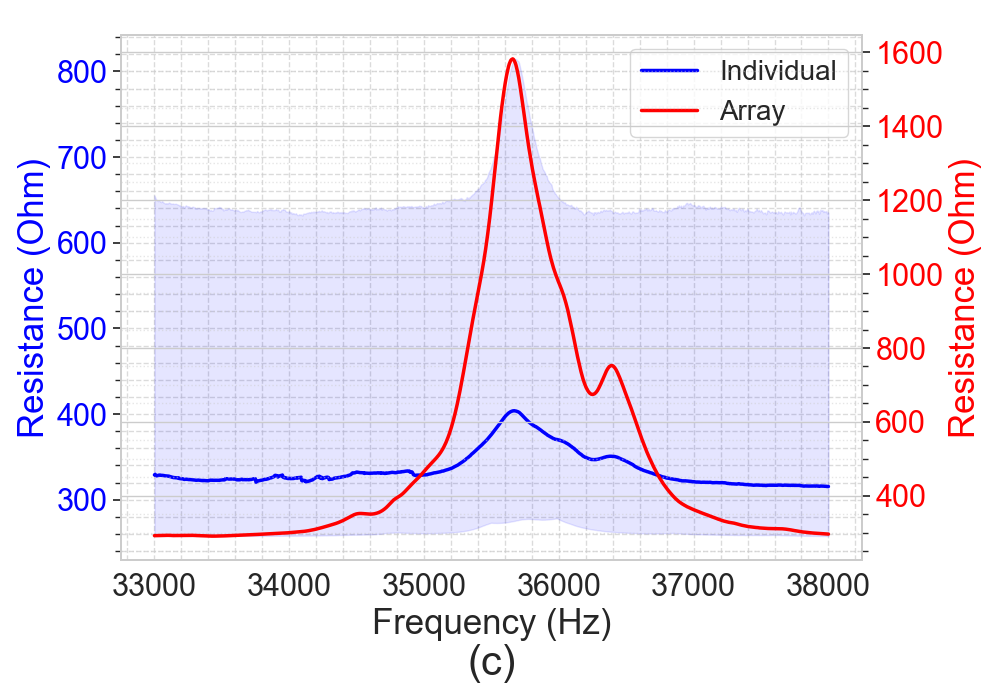} \\[2mm]
    \includegraphics[width=0.325\linewidth]{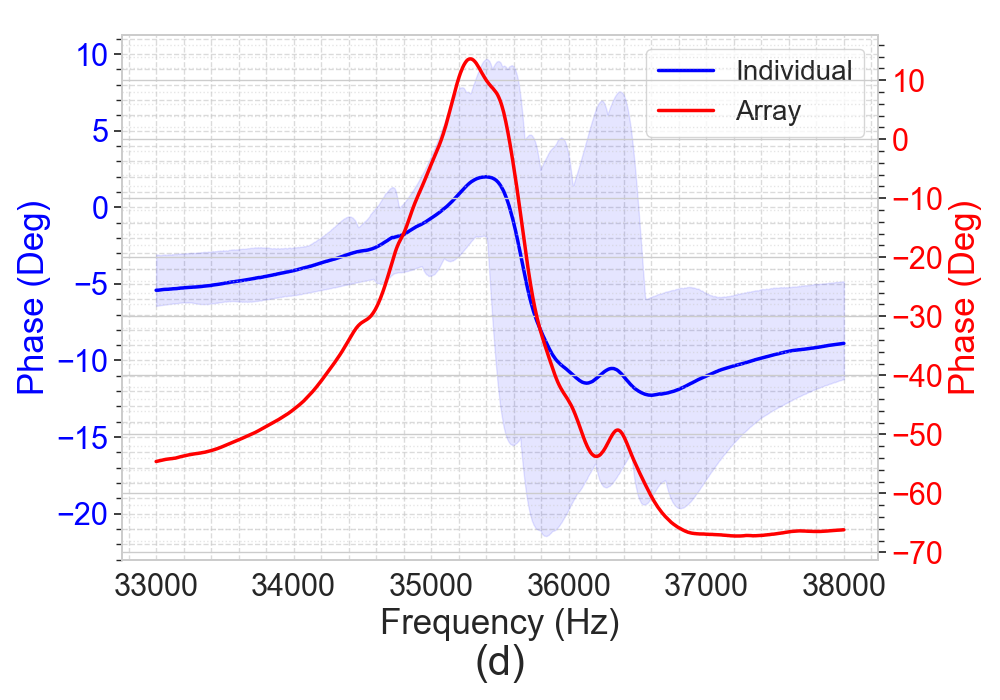} 
    \includegraphics[width=0.325\linewidth]{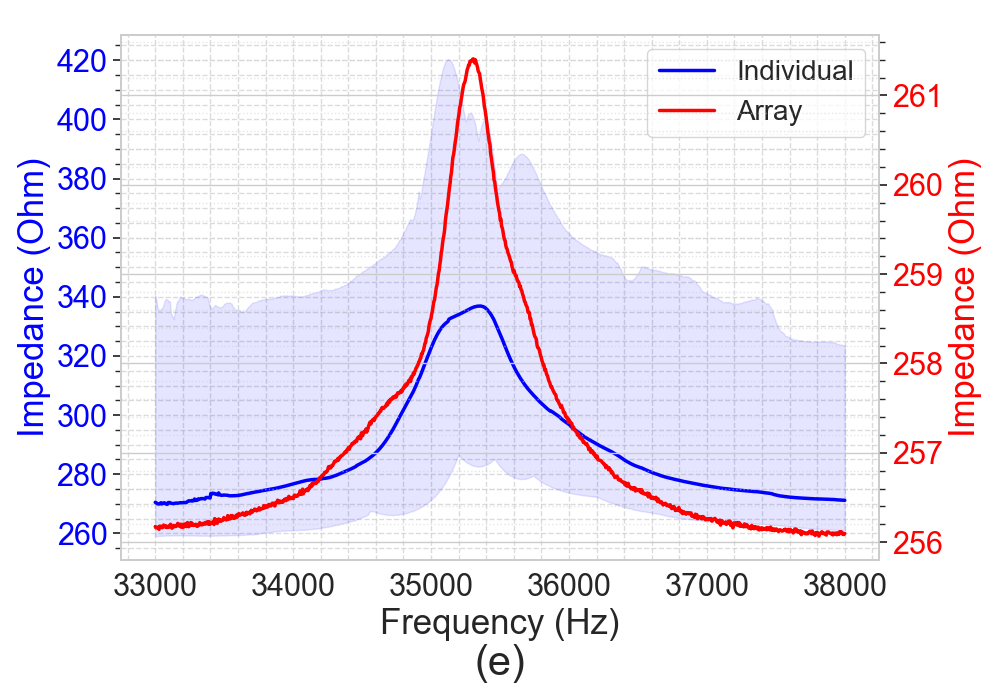}
    \includegraphics[width=0.325\linewidth] {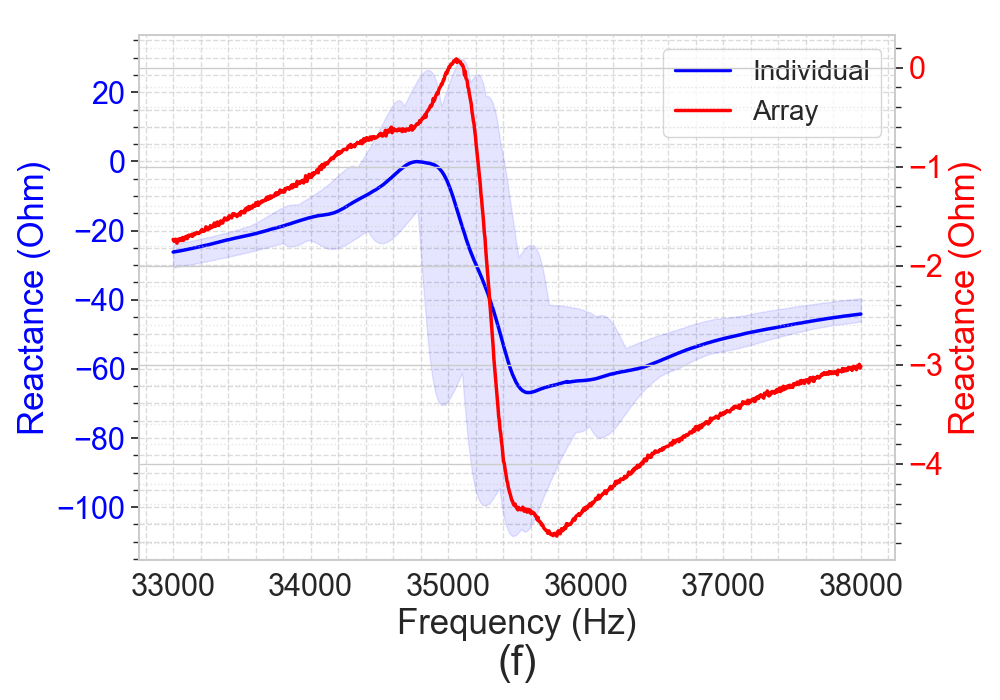} \\[2mm]
    \includegraphics[width=0.4\linewidth]{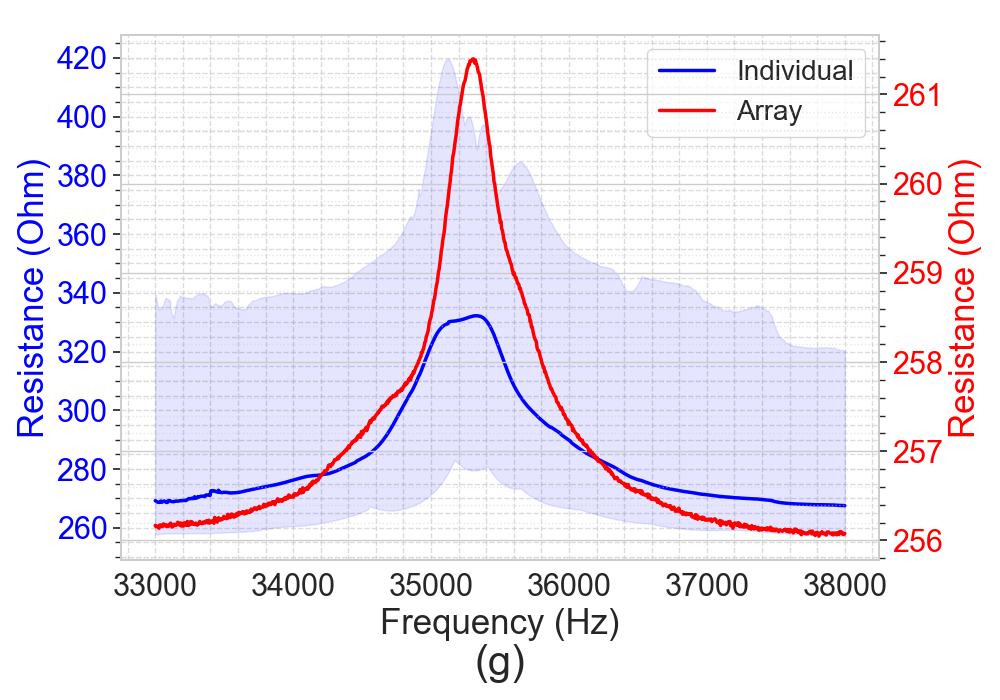} 
    \includegraphics[width=0.4\linewidth]{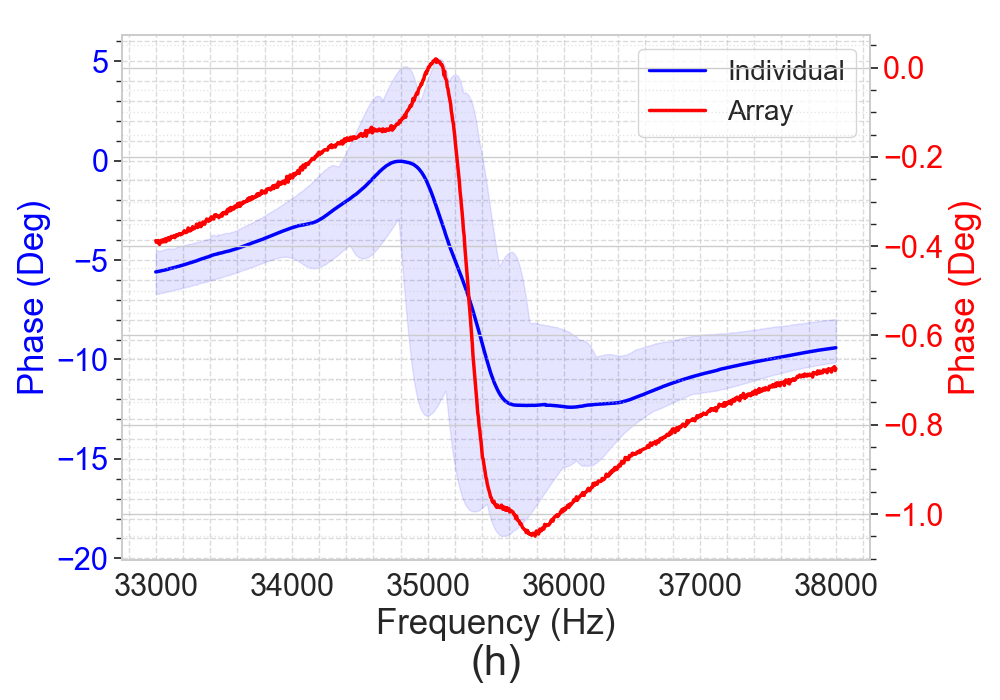} 
    
    \caption{Results of impedance analyses for the BlueME system's Tx-Rx communication in freshwater: (a--d) Rx impedance, reactance, resistance, and phase; (e--h) Tx impedance, reactance, resistance, and phase. The blue shaded areas indicate the bounds across all antennas tested \textbf{individually}, with the blue line showing the average value. The red line shows the result with \textbf{all antennas} connected in series (for Rx) or parallel (for Tx). All results are obtained from impedance analyzer sweeps performed on a DAD3 board from \SI{33}{\kilo\hertz} to \SI{38}{\kilo\hertz}.}
    \label{tx_rx}
\end{figure*}

An impedance sweep from \SI{31}{\kilo\hertz} to \SI{41}{\kilo\hertz} was performed using a DAD3 board connected to an impedance measurement attachment and a BNC adapter. In these tests, the antennas were connected via short wires to a terminal and BNC oscilloscope probes. The impedance magnitude and phase were recorded for each antenna individually.

After individual tests, the antennas were wired together to form the transmitter and receiver arrays. The transmitter antennas were connected in parallel, while the receiver antennas were connected in series. The arrays were then tested as a whole, with impedance measurements taken under the same conditions as the individual antennas. The goal of this setup was to fabricate the ME antennas with sufficiently tight tolerances to validate that wiring the antennas in parallel or series effectively increases bandwidth and/or resonant response \cite{dong_analysis_2023}.

\subsection{Setup:  Underwater Communication Range}
\noindent
\textbf{Lake Trials:} We first conducted proof-of-concept experiments by evaluating the impedance of the proposed system in a water tank measuring $3$\,m (L)$\times$$1.8$\,m (W)$\times$$1.5$\,m (D). The range tests were conducted using two ME antenna arrays in a freshwater lake. The receiver array was mounted on an underwater ROV (BlueROV2) and positioned at a depth of approximately \SI{3}{\meter} below the surface, whereas the transmitter array was mounted on an ASV (BlueBoat) with the transmitter submerged approximately \SI{23}{\centi\meter} below the surface. The transmitter array was driven by a Class-D amplifier powered by a \SI{24}{\volt} supply, which was generated using a boost converter connected to a \SI{12}{\volt} LiFePO$_4$ battery. The input signal to the amplifier varied from \SI{100}{\milli\volt} to \SI{1.5}{\volt}. A small monitoring board was placed in line with the amplifier output, consisting of current sense resistors and an AC-coupled $1$:$10$ voltage divider to measure the current through and voltage across the antenna array.

\begin{figure*}[t]
    \centering
    \includegraphics[width=0.325\linewidth]{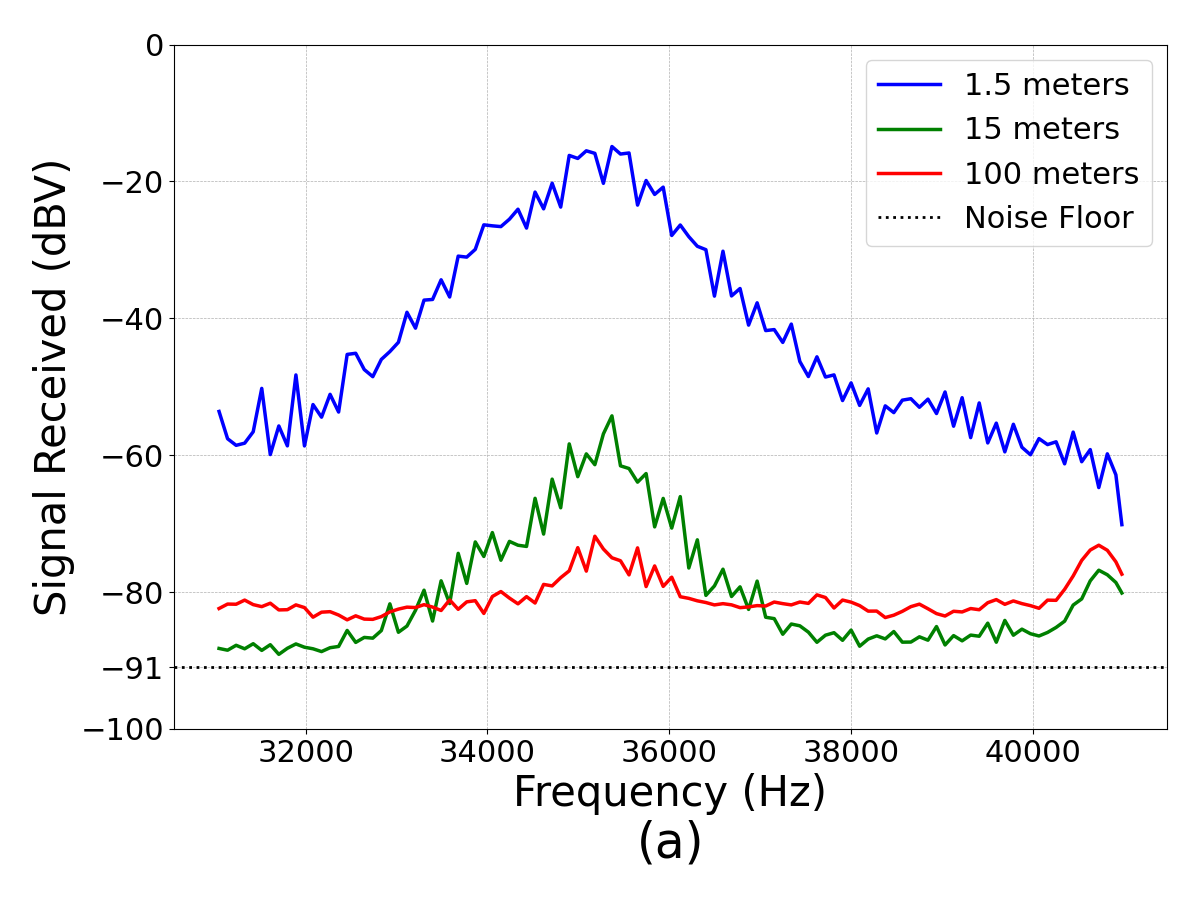}
    \includegraphics[width=0.325\linewidth]{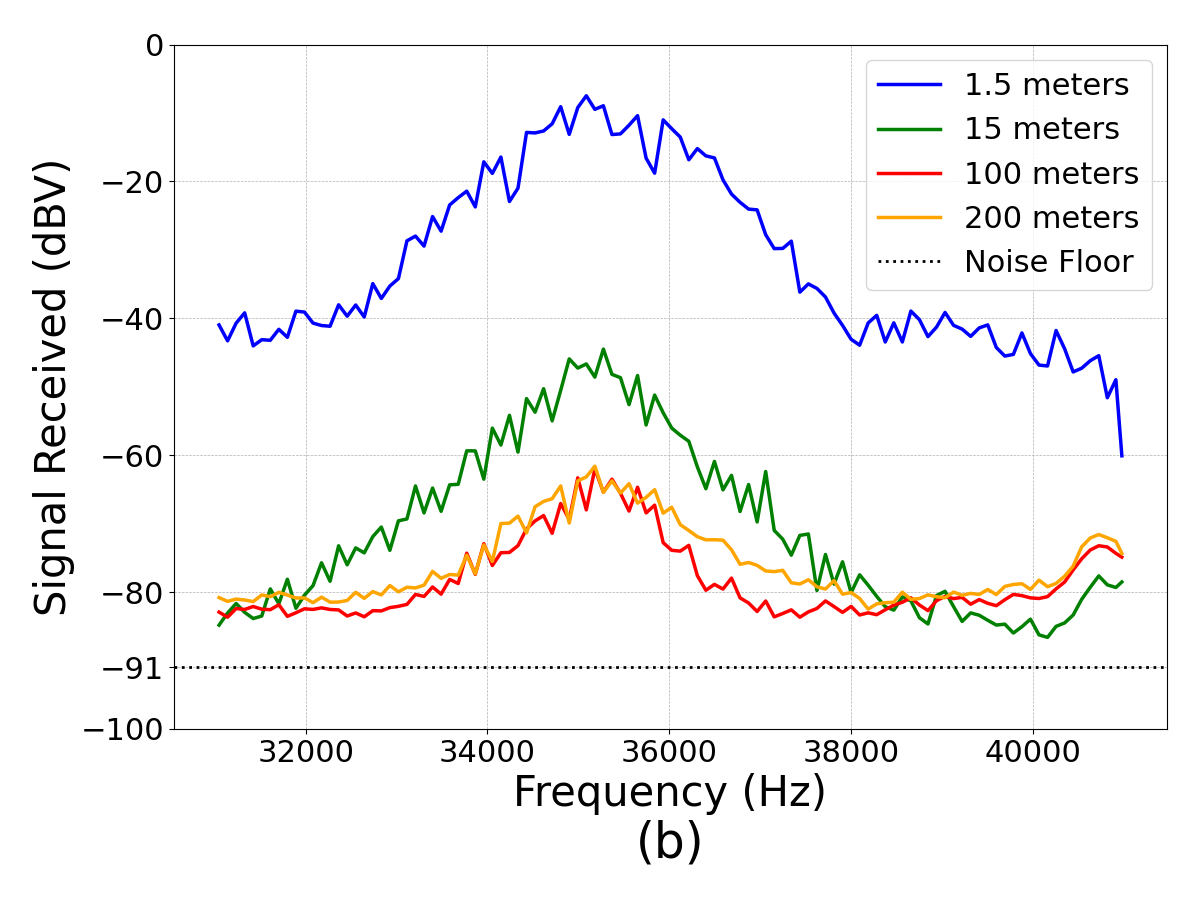}
    \includegraphics[width=0.325\linewidth]{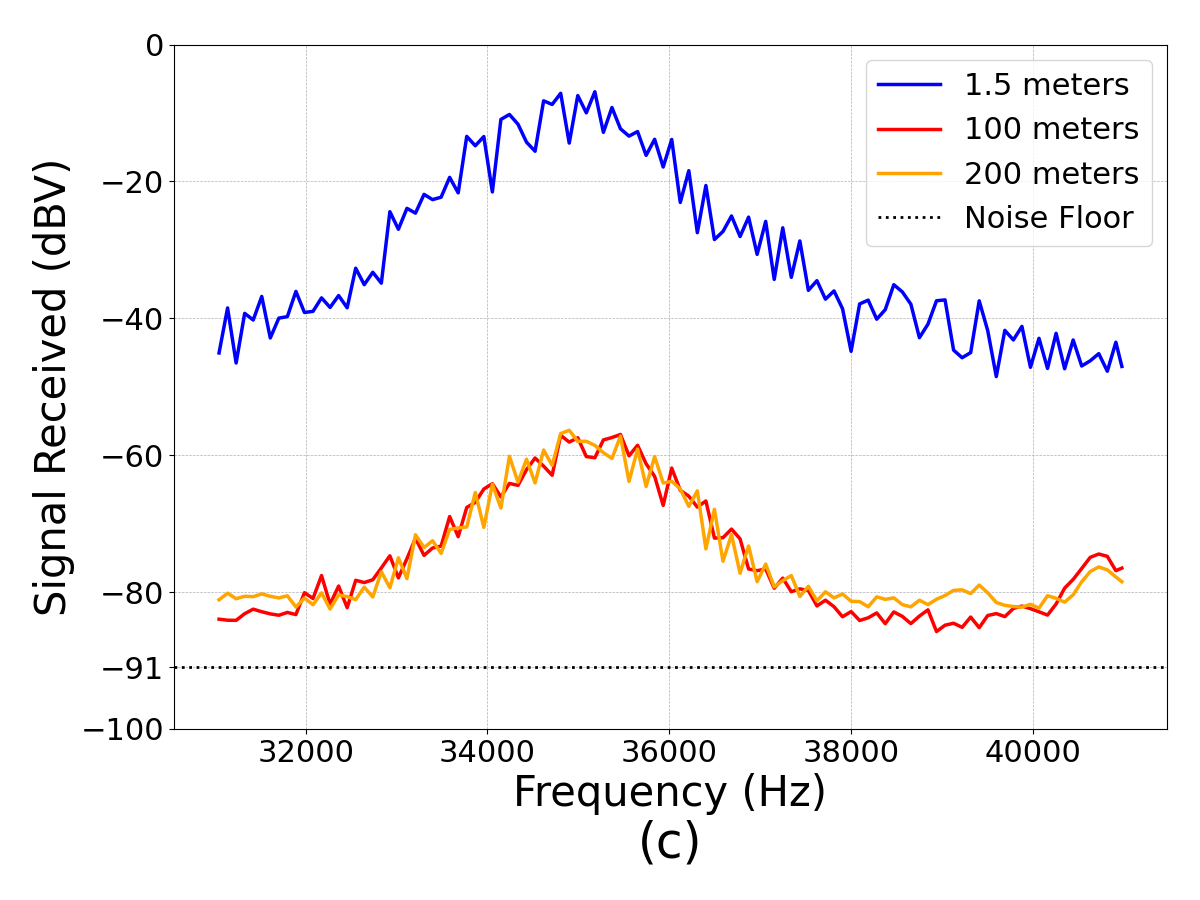}%
    \vspace{-2mm}
    \caption{BlueME signal strength over distance for varied Tx amplifier inputs: (a) \SI{100}{\milli\volt}; (b) \SI{750}{\milli\volt}; and (c) \SI{1.5}{\volt} in freshwater. The plots show the peak RMS amplitude observed on the Rx as a linear frequency sweep is performed on the Tx over a period of $5$ seconds at several distance intervals. The average noise floor level is set at $60$\,dB of amplification when no signal was being transmitted; it is depicted with a dotted horizontal line. For all tests, a low-noise amplifier ({\tt AlphaLab LNA10}) added approximately $60$\,dB of gain prior to sampling the signal; the output amplitude of the transmitter amplifier varied based on the power supply voltage of \SI{24}{\volt}.}
    \label{fig:signal_distance}
\end{figure*}

The receiver array (on the ROV) was connected to an {\tt AlphaLab LNA10} low-noise amplifier (LNA) set to \SI{60}{\decibel} gain with a low-pass filter cutoff at approximately \SI{70}{\kilo\hertz} to mitigate aliasing effects. The output of the LNA was connected to a DAD3 board for data acquisition at a sampling rate of \SI{1}{\mega S/s}. The LNA and DAD3 were housed in a waterproof enclosure mounted on the ROV. Both transmitter and receiver arrays were connected using \SI{5}{\meter} RG-6 coaxial cables with BNC compression fittings.

For the range tests, the transmitter array conducted frequency sweeps from \SI{31}{\kilo\hertz} to \SI{41}{\kilo\hertz} at varying drive strengths. The receiver array collected data continuously, using the FFT peak hold feature of the WaveForms software to record the received signal amplitude across the frequency range. Initial tests were performed in freshwater at distances of \SI{1.5}{\meter}, \SI{15}{\meter}, \SI{100}{\meter}, and \SI{200}{\meter}. The ASV's position during each trial was estimated via GPS with an accuracy of approximately \SI{0.5}{\meter} and used to determine the transmitter–receiver separation (see Fig.~\ref{fig:field_setup}). The maximum test distance was set to \SI{200}{\meter} based on the available test area at the lake. 

\vspace{1mm}
\noindent
\textbf{Ocean Trials:} {For more comprehensive and continuous testing, the experiments were repeated in saltwater off the west coast of Florida (Fig.~\ref{testing_points}). The Gulf test sites ranged from depths of approximately $27'$–$35'$ ($8.2$–$10.7$ m); range measurements were performed for horizontal distances. As shown in Fig.~\ref{fig:field_setup}, the Tx (mounted on ASV) drifted continuously away, with the Tx assembly suspended 3 m below the water surface, while the Rx (mounted on ROV) operated near the ship at depths of $27'$-$35'$. GPS data from the ASV and the ship were used to determine the TX–RX separation, which was measured up to \SI{730}{\meter}.}

\section{Performance Evaluation: Freshwater}

\subsection{Impedance Measurement Results}
\vspace{-1mm}
The impedance measurements of the individual antennas are shown in Fig.~\ref{tx_rx}; the results match our simulations and proof-of-concept experiments, thus validating the proposed design. We observe that the parallel configuration of transmitter antennas resulted in an impedance close to the parallel sum of the individual antenna impedances within the target frequency range. Similarly, the series configuration of the receiver antennas yielded an impedance close to the sum of the individual impedances. These results demonstrate that wiring the antennas in parallel or series effectively adjusts the overall impedance and resonant response, which can improve bandwidth and signal strength; see Fig.~\ref{fig:signal_distance}.

Furthermore, some antennas exhibited smaller resonant peaks near the primary resonant frequency of $35$--\SI{36}{\kilo\hertz}. We hypothesize that this is due to minor variations in the fabrication process or internal micro-cracks. Notably, certain antennas displayed approximately one order of magnitude lower impedance at resonance and higher quality factors than average, indicating stronger resonant responses. Conversely, some antennas had approximately one order of magnitude higher resonant impedance and lower quality factors than average, suggesting weaker resonant responses but wider bandwidths. These variations highlight the importance of consistent fabrication processes and the potential for optimization through mass manufacturing and binning.

\vspace{-1mm}
\subsection{Freshwater Communication Performance}
\vspace{-1mm}
For the range tests, we initially anticipated that reactive near-field behavior would dominate the antenna response, resulting in rapid signal attenuation with distance. However, the measured results exhibited a slower decay rate than expected from purely reactive near-field coupling, with signals observed from \SI{1.5}{\meter} to \SI{200}{\meter}. These observations indicate that part of the transmitted energy propagated beyond the reactive near field, possibly through radiative or interface-guided modes along the water–air or water–seabed boundaries~\cite{smolyaninov_surface_2021}. The maximum test distance in freshwater was set to \SI{200}{\meter} based on the available test area; signal detection beyond this range was subsequently confirmed in ocean trials (see Sec.~\ref{sec:saltwater}).

This performance can be attributed to several factors. First, the electromagnetic wavelength in water at \SI{36}{\kilo\hertz} is approximately \SI{170}{\meter}, much shorter than in air due to water's higher permittivity and conductivity. The wavelength in a conductive medium is given by $\lambda = {2\pi}/{\beta}$, where $\beta$ is the phase constant. In conductive media, $\beta$ is influenced by permittivity, permeability, conductivity, and angular frequency. The reduced wavelength improves the radiation efficiency of electrically small antennas. Additionally, the ME antennas, when configured in arrays, are within each other's near fields, effectively forming a larger equivalent magnetic dipole with increased radiation ~\cite{dong2022vlf}. Antenna misalignment had a more pronounced effect at closer ranges. This can be attributed to near-field interactions causing destructive interference at close range and/or signal propagation along more diverse paths at greater distances.

Moreover, we found that the rate of attenuation decreased with increasing distance. This may be due to electromagnetic waves propagating along paths with lower attenuation (\eg, the lake bed, surface wave)~\cite{smolyaninov_development_2020} or due to reflections at shallow surfaces~\cite{che_re-ev_new}. Unlike acoustic propagation, electromagnetic waves' significantly higher propagation velocity in water results in wavelengths of orders of magnitude longer than acoustic waves at comparable frequencies. Consequently, typical path length differences produce smaller phase differences between multipath components, rendering interference effects negligible. Additionally, reflections from obstacles and natural features are minimal since objects of size comparable to or smaller than the electromagnetic wavelength tend to produce weak scattering, with incident waves primarily diffracting around or penetrating through such objects rather than producing strong reflections.

\begin{figure*}[t]
    \centering
    \includegraphics[width=0.48\linewidth]{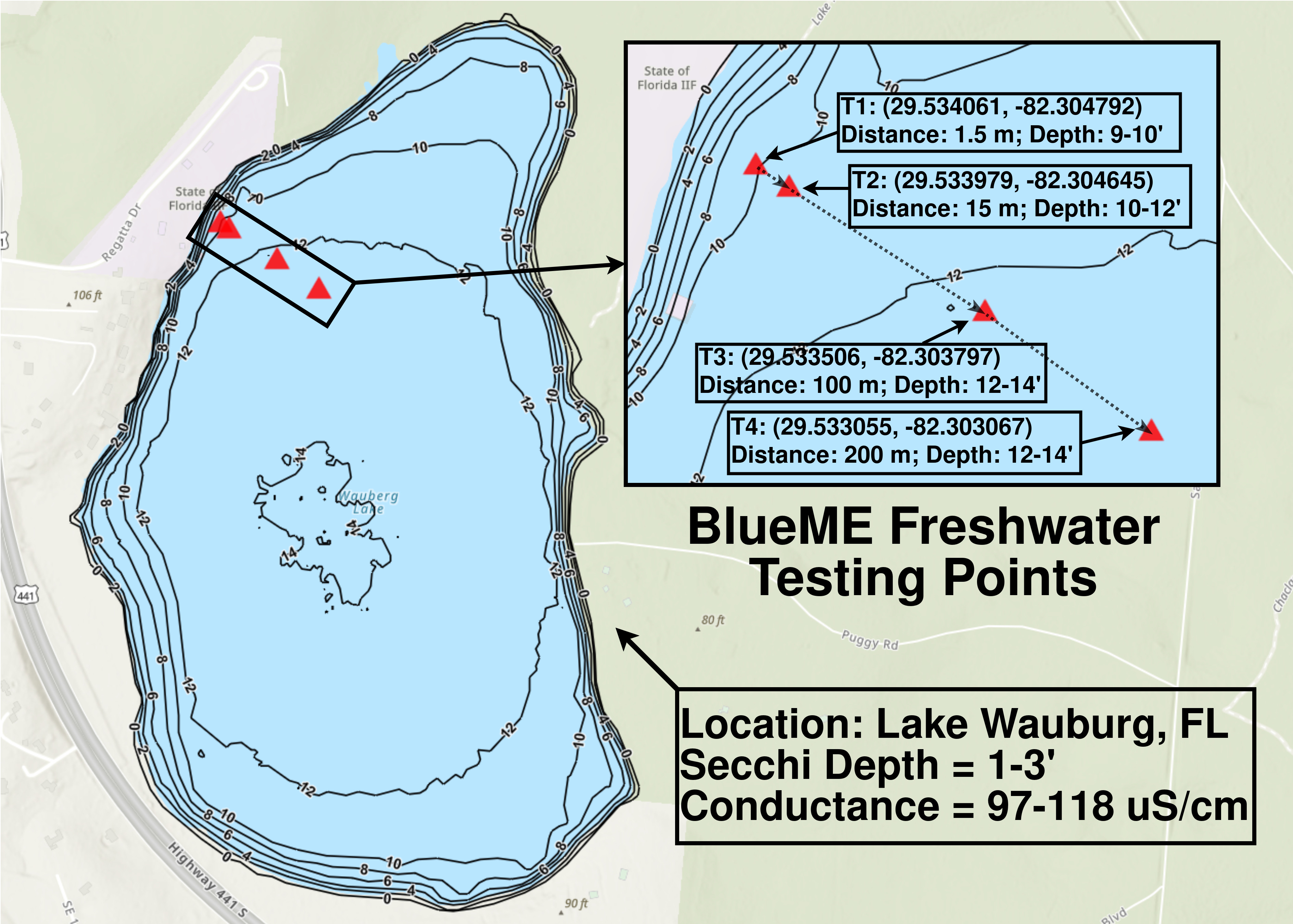}
    \includegraphics[width=0.48\linewidth]{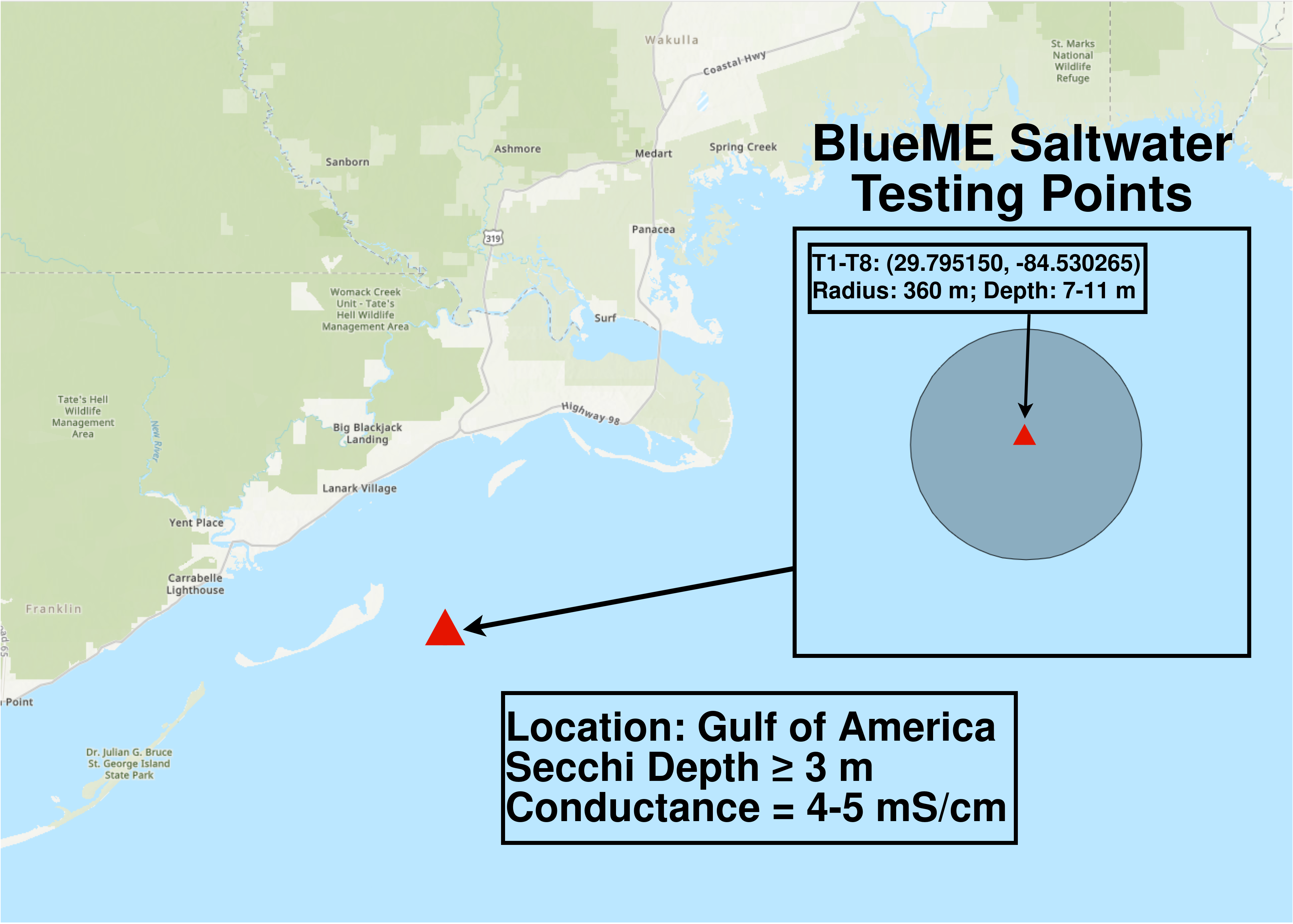}
    \vspace{-1mm}
    \caption{The deployment locations and relevant information from our field tests. (\textbf{Left}) Lake Wauburg, FL, where the freshwater experiments were conducted; depth contours are shown in feet, with measurement points labeled T1–T4. The water depths ranged from 1 to 3 ft, and water conductance ranged from 97 to 118 µS/cm; data is from the monitoring data collected by UF IFAS LAKEWATCH~\cite{university_of_florida_alachua_2022}. Tests were performed along the indicated transect, beginning with a transmitter-receiver separation of 1.5 m near the dock and extending to 200 m across. (\textbf{Right}) Ocean trial locations at the West Gulf Coast near St. Teresa, FL; tests were performed at 8-11 m (27–35 ft) depths spanning multiple sites that are about 1 Km offshore.      }
    \label{testing_points}
    \vspace{-1mm}
\end{figure*}

\subsection{Analyzing the Theory of Operation} \label{sec:calc_wavelength}
The performance characteristics of ME antenna arrays submerged underwater are not extensively explored in the literature. As such, we make several observations in our analysis.

\medskip
\noindent \textbf{Wavelength Changes}. The antenna's observed rate of signal drop-off appears to change across distances. This may be explained by the change in electromagnetic wavelength from air to fully submerged in water, which is also a more conductive medium. To this end, we calculate the expected wavelength of an electromagnetic plane wave in an underwater environment. As shown by~\cite{moore_radio_1967}, the electromagnetic propagation constant can be calculated as:
\begin{equation}
    \gamma = \sqrt{j\omega\mu\sigma} = \sqrt{\frac{\omega\mu\sigma}{2}}(1 + j) = \sqrt{\pi f \mu \sigma}(1 + j),
\end{equation}
where $\mu$ is the water medium's permeability, $\sigma$ denotes conductivity, and $\omega = 2\pi f$ is the angular frequency. Given that the displacement current may be neglected at sufficiently low frequencies~\cite{moore_radio_1967}, $\gamma$ can be expressed as:
\begin{equation}
    \gamma = \alpha + j\beta = \sqrt{\frac{\omega\mu\sigma}{2}} (1 + j) = \beta (1 + j),
\end{equation}
where $\alpha$ and $\beta$ are the attenuation and phase constants, respectively, and $\alpha = \beta$ in this case. Subsequently, the expression for wavelength in terms of $\beta$ and $\sigma$ is:
\begin{align}
    \beta &= \sqrt{\frac{\omega\mu\sigma}{2}} = \sqrt{\pi f \mu \sigma}, \\
    \Rightarrow \lambda &= \frac{2\pi}{\beta} = \frac{2\pi}{\sqrt{\pi f \mu \sigma}}.
\end{align}

The specific conductivity $\sigma$ of Lake Wauburg in Gainesville, Florida ($29.52653^{\circ}$N, $82.30464^{\circ}$W) is approximately $97$\,\si{\micro\siemens\per\centi\meter} as measured over a 16-year averaging period~\cite{university_of_florida_alachua_2022}, which equals $0.0097$\,\si{\siemens\per\meter}. Using the formula above, we calculate the wavelength of an electromagnetic wave to be between approximately \SI{182}{\meter} and \SI{158}{\meter} for frequencies from \SI{31}{\kilo\hertz} to \SI{41}{\kilo\hertz}, respectively. Thus, the underlying frequency has a wavelength of approximately \SI{170}{\meter}, compared to approximately \SI{8327}{\meter} in air.

\begin{figure}[t]
    \centering
    \includegraphics[width=\linewidth]{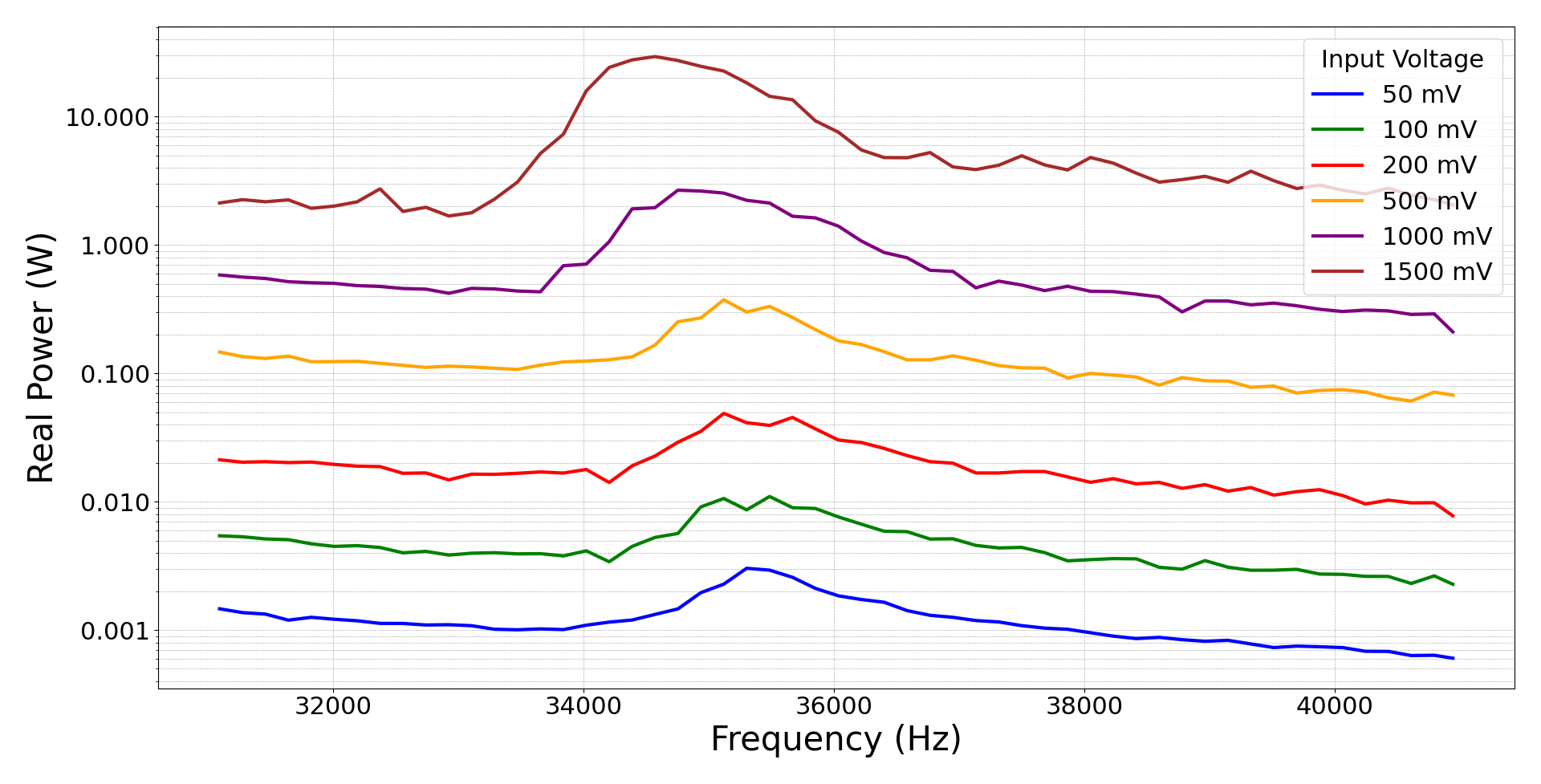}%
    \vspace{-1mm}
    \caption{Measured real (in-phase) power consumption versus frequency sweep of the transmitter circuit for input signal voltages from $50$\,mV to $1500$\,mV (peak amplitude). The amplifier has an input voltage of $\pm3.9$ Vpp and produces an output voltage range of $\pm24$ Vpp (on data from freshwater experiments).}
    \label{power_voltage}
    \vspace{-2mm}
\end{figure}

\medskip
\noindent 
\textbf{Far-Field Effects}. For electrically small antennas, the reactive near-field region extends to $r < \frac{\lambda}{2\pi}$ from the antenna, which is roughly \SI{27}{\meter} in our case. The radiative near-field extends to approximately one wavelength away~\cite{dong_analysis_2023}. The transition zone between the near-field and far-field regions lies between $\lambda$ and $2\lambda$. Therefore, the far-field region of the antenna begins from $2\lambda \approx \SI{340}{\meter}$ and beyond.

\medskip
\noindent 
\textbf{Radiation Resistance}. The radiation resistance characterizes the efficiency of an antenna in radiating power~\cite{dong_analysis_2023, burnside2023axial}; it is calculated by the following expression:
\begin{equation}
    R_{\text{rad}} \approx \eta \frac{\left(\beta L\right)^2}{24\pi} \approx \eta \frac{\pi}{6} \left(\frac{L}{\lambda}\right)^2,
\end{equation}
where $\eta$ is the wave impedance of the medium, $\beta$ is the phase constant, $L$ is the antenna length, and $\lambda$ is the wavelength. In water, the wave impedance is $\eta = \frac{Z_0}{\sqrt{\epsilon_r}}$, where $Z_0 = 120\pi$\,\si{\ohm} is the free-space impedance and $\epsilon_r \approx 80$ is the relative permittivity of water. Thus, $R_{\text{rad}}$ becomes:
\begin{equation}
     R_{\text{rad}} \approx \frac{Z_0}{\sqrt{\epsilon_r}} \frac{\pi}{6} \left(\frac{L}{\lambda}\right)^2. 
\end{equation}
Assuming that $L$ and all other constants remain identical between air and water, we find 
\begin{equation}
    R_{\text{rad}} \propto \frac{1}{\lambda^2 \sqrt{\epsilon_r}}.
\end{equation}
Although the relative permittivity of water is significantly higher than air ($80$:$1$), the substantial decrease in wavelength (from \SI{8327}{\meter} in air to \SI{170}{\meter} in water) results in a net increase in radiation resistance per antenna by a factor of approximately $ \left( \frac{8327}{170} \right)^2 \times \frac{{1}}{\sqrt{80}} \approx 267$. This leads to an increase in the total radiated power per antenna by a similar factor.

\medskip
\noindent
\textbf{Single Antenna vs Antenna Array.} Increasing number of antennas on the transmitter side ($1$ to $15$) produces a quadratic increase in net radiated power, as shown by Dong~\etal~\cite{dong2022vlf}:
\begin{equation}
    P_{\text{rad}} = \frac{\mu_0 \omega^4}{12 \pi c^3}|m_0|^2.
\end{equation}
Since the constants remain the same across array configurations and $m_0$ is linearly proportional to the number of ME antenna elements in the transmitter array $N_t$, we have 
\begin{equation}
P_{\text{rad}} \propto |m_0|^2 \Rightarrow P_{\text{rad}} \propto {N_t}^2.
\end{equation}
Thus, an increase in total radiated power for a $3\times5$ array on the transmitter side is a factor of $15\times15=225$, assuming negligible mutual coupling losses.

Increasing the number of antennas on the receiver side also increases the receiver sensitivity linearly, assuming all antennas are identical, \ie, a factor of $225 \times 15 = 3{,}375$ improvement in the overall link budget. Consequently, using an array of $15$ antennas underwater compared to a single antenna in air theoretically yields an overall link budget improvement by a factor of $3{,}375 \times 267 = 901{,}125$, which corresponds to an increase of approximately $119$\,dB -- a factor not specifically discussed in the literature~\cite{dong2022vlf,dong2020portable,du2023very,xu2019low, xiang2016subsea}. In practice, however, resonant frequency mismatches and increased mutual coupling at closer antenna spacings and higher drive strengths will degrade phase coherence within the transmitter array~\cite{dong_analysis_2023}, reducing the actual achievable improvement from the array configuration.

\begin{figure*}[t]
    \vspace{-2mm}
    \centering
    \includegraphics[width=0.325\linewidth]{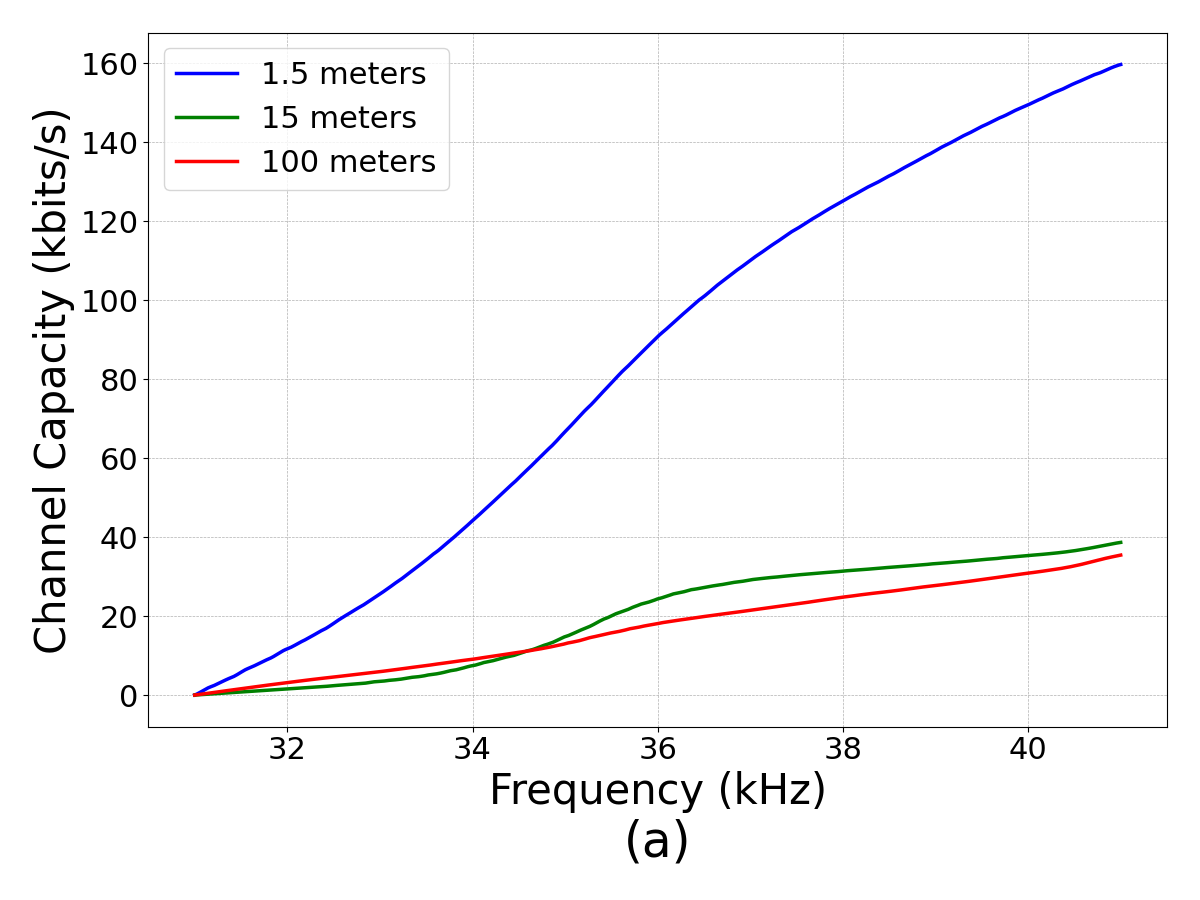} 
    \includegraphics[width=0.325\linewidth]{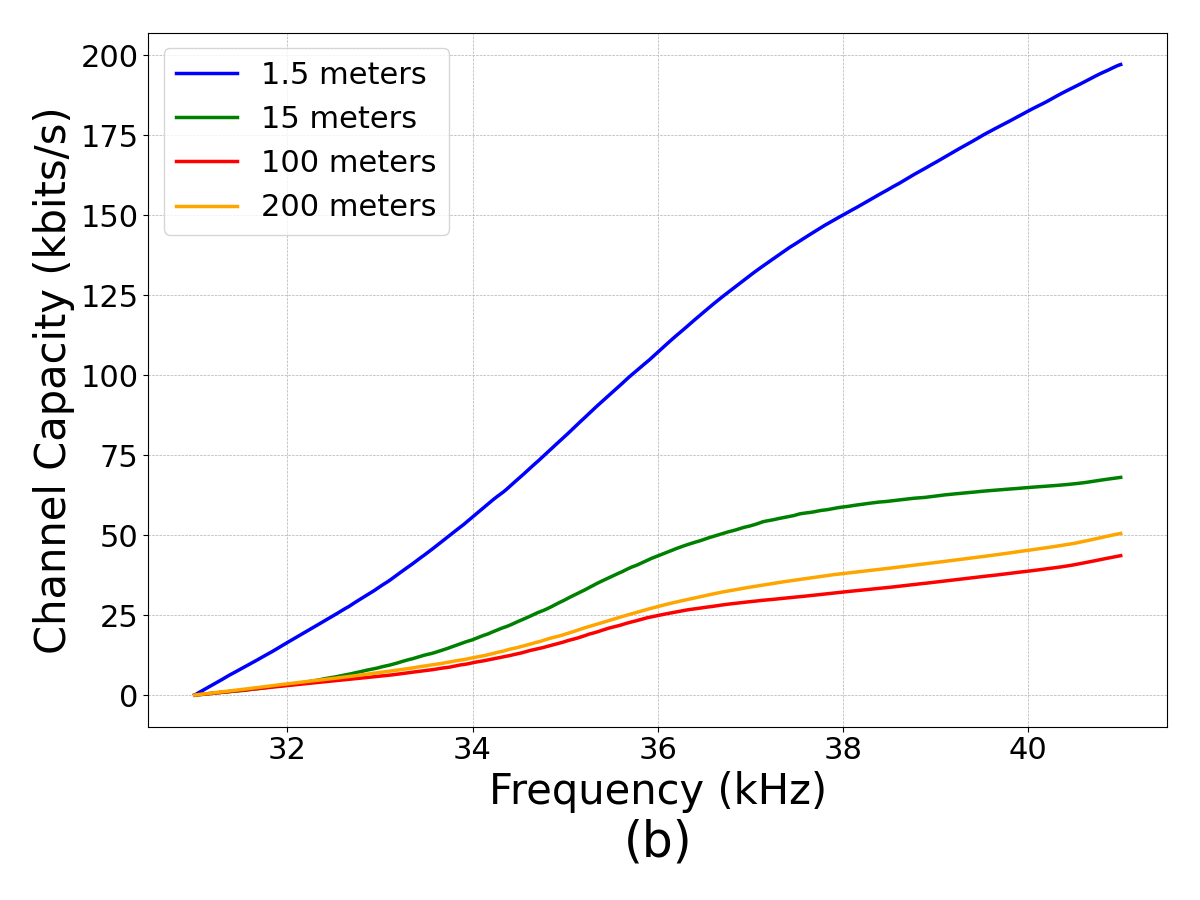} 
    \includegraphics[width=0.325\linewidth]{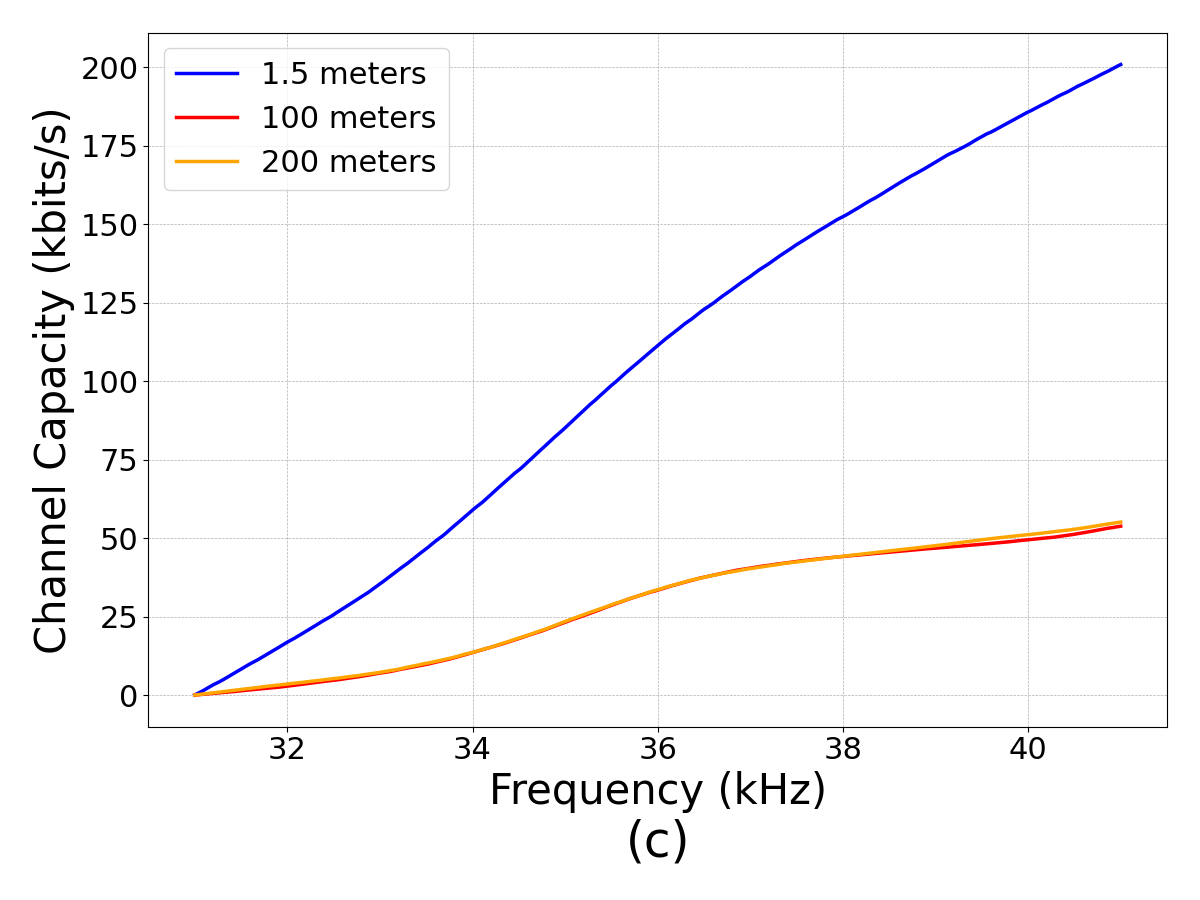}%
    \vspace{-1mm}
    \caption{Channel capacity analysis derived from the BlueME signal strength measurements (on data from freshwater experiments): (a) \SI{100}{\milli\volt}; (b) \SI{750}{\milli\volt}; and (c) \SI{1.5}{\volt}. The plots show the cumulative channel capacity calculated at each frequency point using the measured signal-to-noise ratio, with a constant noise floor of \SI{-91}{\deci\bel\volt}. Each point on the curve represents the theoretical tight upper bound on the error-free data rate achievable up to that frequency, assuming optimal encoding. To find the channel capacity within any specific frequency band, one can take the difference between the cumulative capacity values at the upper and lower frequency bounds.}
    \label{fig:capacity_distance}
    \vspace{-1mm}
\end{figure*}

\vspace{-1mm}
\subsection{Power Footprint and Robustness}
\vspace{-1mm}
The proposed BlueME system consumes minimal power, with optimal performance achieved at approximately \SI{1}{\watt} of power consumption (see Fig.~\ref{power_voltage}). Nonlinear effects such as spring-softening in the piezoelectric materials and spring-hardening in the Metglas layers start to become significant at higher drive strengths, leading to shifts and distortions in the resonant frequency with decreased quality factors \cite{chu_multilayered_2023}. In Fig.~\ref{power_voltage}, increased drive power shifts the resonance peak to lower frequencies and distorts the response shape. Shown earlier in Fig.~\ref{fig:signal_distance}, the received signals at various transmitter power levels and distances, a downward frequency shift is also observed. However, since the receiver experiences only the coupled power from the transmitter, these power levels are much lower, resulting in minimal nonlinear distortion at the receiver. Thus, the observed frequency shift primarily reflects the transmitter's nonlinear behavior combined with the receiver's nearly constant resonant response. These effects can be exploited to adjust for resonance mismatches between RX/TX arrays by varying the drive strength. We also observed that the antenna performance was not significantly affected by obstacles, water turbidity, or multipath propagation, which are major challenges for traditional acoustic and optical modalities.

\vspace{-1mm}
\subsection{Channel Capacity}
\vspace{-1mm}
The development of an optimal modulation scheme is beyond the scope of this paper; however, we can derive theoretical channel limits from our measurements using the Shannon-Hartley theorem~\cite{de2024shannon}. For a simple channel with additive white Gaussian noise, the Shannon-Hartley theorem gives the channel capacity $C$ as:
\begin{equation}
C = B \log_2\left(1 + \frac{S}{N}\right) = B \log_2\left(1 + \frac{V_{S}^2}{V_{N}^2}\right)
\end{equation}
where $B$ is the channel bandwidth, $S$ is the received signal power, and $N$ is the noise power. The signal and noise powers can be calculated from their respective RMS voltages squared. Based on measurements of the LNA noise floor, the noise spectra within our frequency range of interest (31 kHz to 41 kHz) can be approximated as Gaussian.
For our frequency-selective underwater channel, we must consider the varying SNR across different frequencies. This leads to a more general form of the capacity formula:
\begin{equation}
C = \int_0^B \log_2\left(1 + \frac{S(f)}{N(f)}\right) df.
\end{equation}
Here, $S(f)$ and $N(f)$ represent the frequency-dependent signal and noise power spectral densities. Since our measurements are taken at discrete frequency points (shown in Fig.~\ref{fig:signal_distance}), we can express this as:
\begin{equation}
C = \sum_{n=0}^{N-1} \log_2\left(1 + \frac{S(n{\Delta f})}{N(n{\Delta f})}\right) {\Delta f}
\end{equation}
where $n$ is the frequency index, $N$ is the total number of frequency points, and ${\Delta f}$ is the frequency resolution of our measurements (${\Delta f} = B/N$). Using this discretized form, we calculate the cumulative theoretical channel capacity from our signal strength measurements, with results shown in Fig.~\ref{fig:capacity_distance}.

\section{Performance Evaluation: Saltwater}
\label{sec:saltwater}
Antenna performance was further evaluated in saltwater trials conducted off the west coast of Florida along the Gulf (see Fig.~\ref{testing_points}). The primary objective was to examine the effect of higher conductivity (\(\sigma \approx 5\,\mathrm{S/m}\)) on achievable communication range. The experimental setup mirrored that of the freshwater tests, with all equipment housed in watertight enclosures and mounted on the BlueBoat (Tx) and BlueROV2 (Rx). Minor modifications included the integration of a LattePanda single-board computer for autonomous control and the replacement of the original Class-D amplifier with a smaller unit to accommodate spatial constraints. During testing, all systems were powered by onboard batteries and operated fully submerged, except for an Ethernet tether used to stream real-time data from the receiver. For the saltwater trials, the LNA gain was reduced to 40\,dB from 60\,dB in the freshwater tests to prevent saturation, which lowered the measured noise floor but did not measurably affect the overall SNR.

Autonomous frequency sweeps between \(30\,\mathrm{kHz}\) to \(40\,\mathrm{kHz}\) were conducted at one-minute intervals by the transmitter. The BlueBoat carrying the transmitter was allowed to drift freely, with its position tracked using GPS. The receiver was positioned approximately \(15\,\mathrm{m}\) from a stationary pontoon boat and recorded peak-hold FFT measurements each time new GPS coordinates were logged.

\begin{figure}[t]
    \centering
    \includegraphics[width=\linewidth]{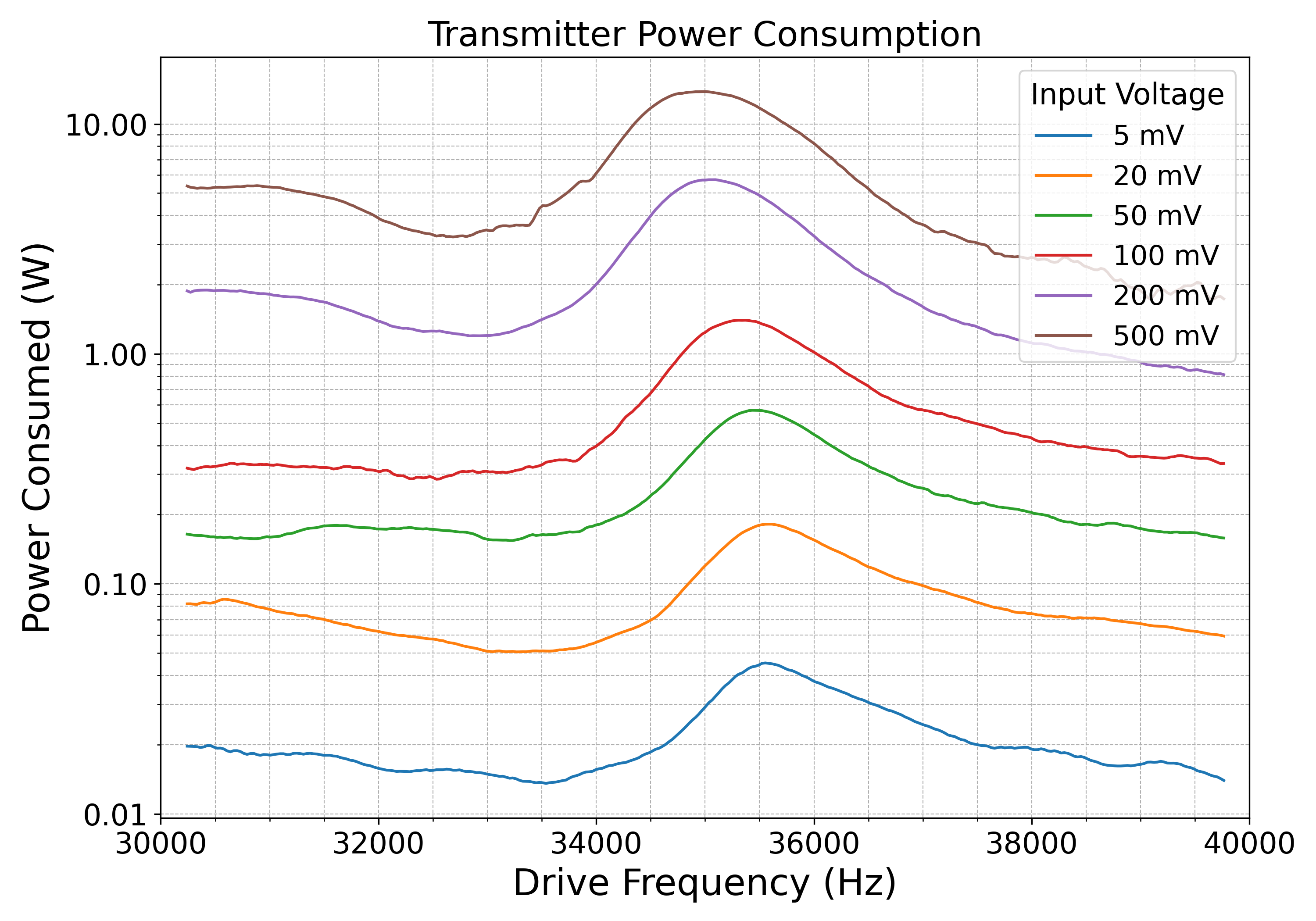}
    \caption{The measured transmitter power consumption versus frequency (in saltwater). The input voltage amplitudes varied between \(5\,\mathrm{mV_{pp}}\) and \(500\,\mathrm{mV_{pp}}\), while the frequency sweep was performed from \(30\,\mathrm{kHz}\) to \(40\,\mathrm{kHz}\).}
    \label{tx_power_sweep}
\end{figure}

An initial transmitter power sweep (shown in Fig.~\ref{tx_power_sweep}) confirmed negligible differences in antenna characteristics relative to the freshwater experiments (shown earlier in Fig.~\ref{power_voltage}), validating the consistency of power characteristics across test conditions. Subsequent experiments involved transmitter sweeps at fixed distances under varying input power levels, as well as tests at fixed transmitter power with varying transmitter–receiver separation. From these measurements, the received SNR was computed, and the corresponding spectral characteristics were analyzed, as presented in Figs.~\ref{rx_distance_snr_sweep} and \ref{rx_power_sweep}.

\subsection{Wavelength and Field Regions}
The near-field, transition, and far-field boundaries were estimated using the same method applied in the freshwater analysis. Conductivity at the site was \(\sigma = 48.18 \pm 2.41~\mathrm{mS/cm}\) and the water temperature ranged from 20–25\(^\circ\)C~\cite{mercado_fsucml_2025}. For a center frequency of \(f = 36\,\mathrm{kHz}\) and seawater conductivity of \(\sigma = 4.818\,\mathrm{S/m}\), the wavelength is calculated as \(\lambda_{\text{sw}} \approx 7.59\,\mathrm{m}\). The corresponding theoretical boundaries for the antenna regions are defined as:
\begin{itemize}
    \item Reactive near-field (\(r < \lambda/2\pi\)): \(r < 1.21\,\mathrm{m}\)
    \item Radiative near-field (\(\lambda/2\pi < r < \lambda\)): \(1.21\,\mathrm{m} < r < 7.59\,\mathrm{m}\)
    \item Transition region (\(\lambda < r < 2\lambda\)): \(7.59\,\mathrm{m} < r < 15.19\,\mathrm{m}\)
    \item Far-field (\(r > 2\lambda\)): \(r > 15.19\,\mathrm{m}\)
\end{itemize}

\begin{figure*}[t]
    \centering
    \includegraphics[width=0.49\linewidth]{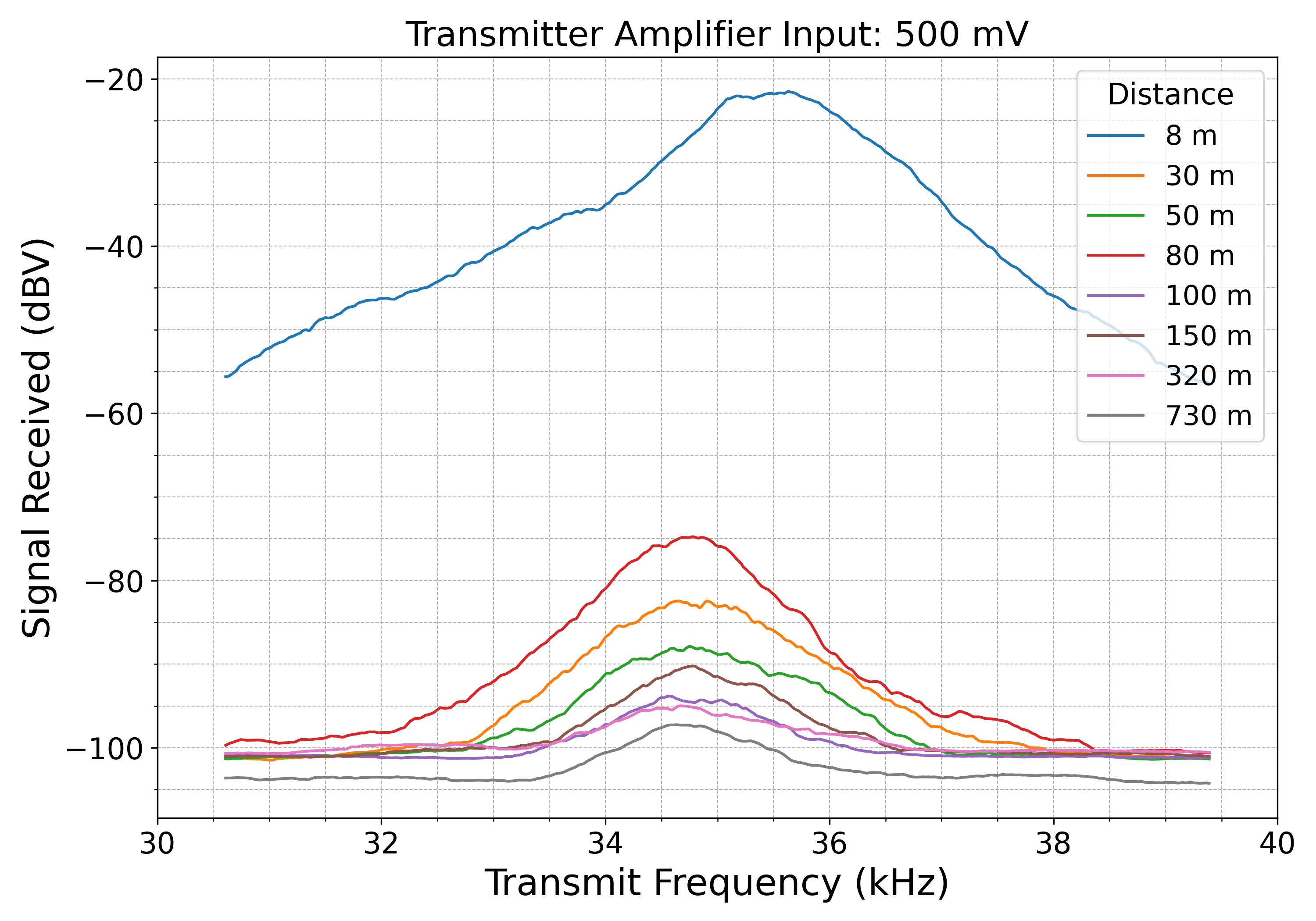}
    \includegraphics[width=0.49\linewidth]{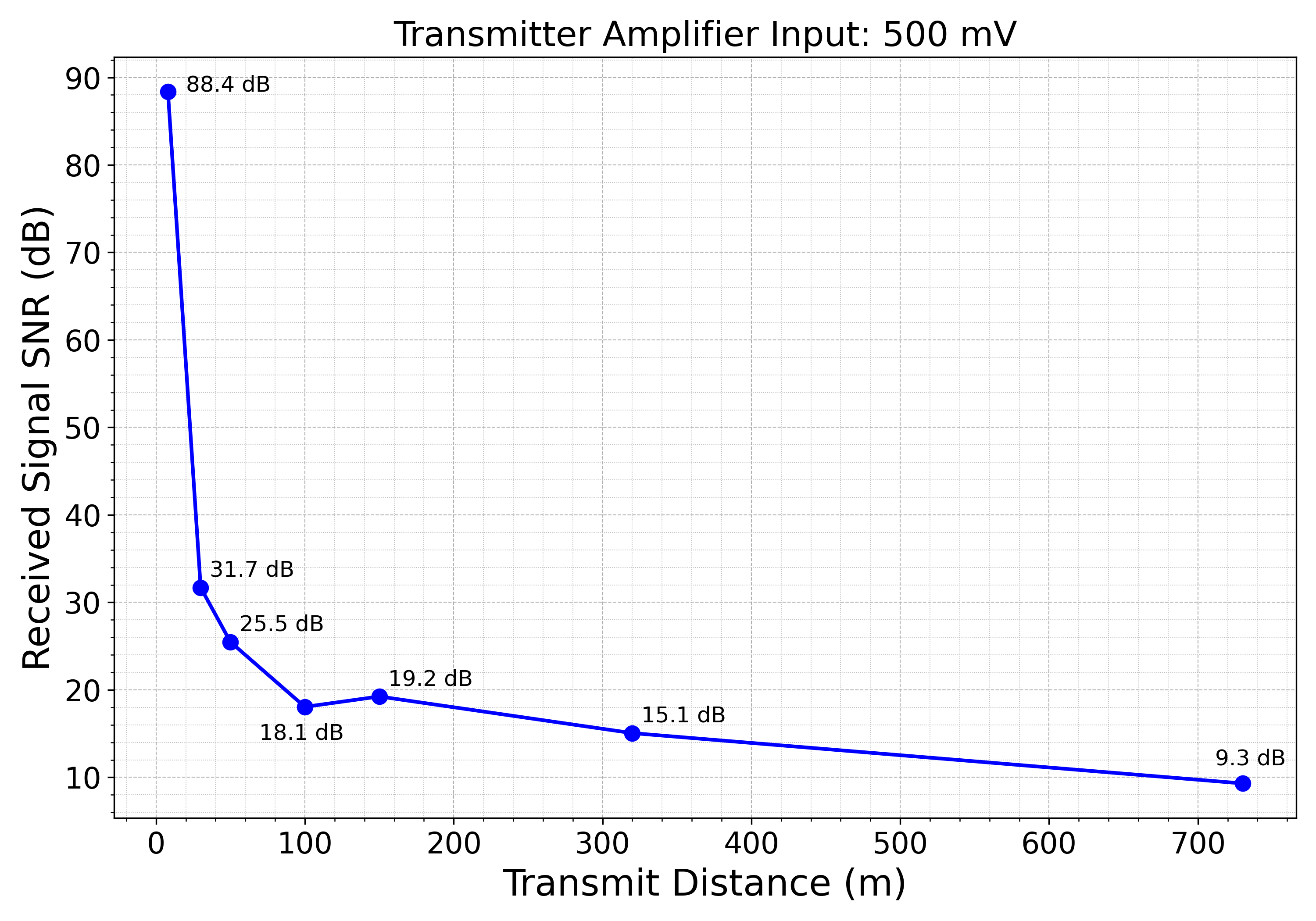}
    \vspace{-1mm}
    \caption{Measured received signals from the saltwater field trials with transmitter output power \(10\,\mathrm{W}\) and input amplitude \(500\,\mathrm{mV_{pp}}\), measured over separations from \(8\,\mathrm{m}\) to \(730\,\mathrm{m}\). (\textbf{Left}) Received spectra at each separation; and (\textbf{Right}) Peak received SNR as a function of distance.}
    \label{rx_distance_snr_sweep}
    \vspace{-1mm}
\end{figure*}

The measured SNR at distances below approximately \(15\,\mathrm{m}\) exhibited a steeper decay than the nominal \(1/r^3\)–\(1/r^6\) dependence typically associated with reactive or radiative near-field behavior. At larger separations (\(r > 15\,\mathrm{m}\)), the attenuation trend approached an approximate \(1/r^1\) dependence, suggesting that signal propagation may have occurred along paths not confined to the bulk medium, such as along the water–air or water–seabed interfaces~\cite{smolyaninov_surface_2021,smolyaninov_superlensing_2023}. A localized deviation from this overall trend was observed at a transmitter–receiver spacing of \(80\,\mathrm{m}\) (Fig.~\ref{rx_distance_snr_sweep}), where the peak received signal level was approximately \(15\,\mathrm{dB}\) higher than that measured at \(50\,\mathrm{m}\). A similar deviation was noted at \(150\,\mathrm{m}\), where the peak signal level exceeded that at \(100\,\mathrm{m}\) by roughly \(5\,\mathrm{dB}\). A reduction in the measured noise floor was observed during the \SI{730}{\meter} test, caused by a drop in the LNA supply voltage as its battery approached full discharge. All measurements were performed under identical experimental conditions, with the submerged transmitter naturally drifting and rotating during the tests.

\subsection{Radiation Resistance} 
The same relationship used in the freshwater case can be extended to saltwater by considering the higher conductivity of the medium. Starting from the general dependence
\begin{equation}
    R_{\mathrm{rad}} \propto \frac{1}{\lambda^{2}\sqrt{\varepsilon_r}},
    \qquad
    \lambda = \frac{2\pi}{\sqrt{\pi f \mu \sigma}},
\end{equation}
and substituting for $\lambda$, the radiation resistance becomes directly dependent on the conductivity of the surrounding medium. Assuming the frequency, permeability, and relative permittivity remain constant between freshwater and saltwater, this simplifies to
\begin{equation}
    R_{\mathrm{rad}} \propto \frac{f\,\mu\,\sigma}{4\pi\,\sqrt{\varepsilon_r}} \;\Rightarrow\; R_{\mathrm{rad}} \propto \sigma.
\end{equation}
Substituting representative conductivities for freshwater ($\sigma_{\mathrm{fw}} = 9.7\times10^{-3}\,\mathrm{S/m}$) and saltwater ($\sigma_{\mathrm{sw}} \approx 5\,\mathrm{S/m}$) gives the expected ratio as
\begin{equation}
    \frac{R_{\mathrm{rad,sw}}}{R_{\mathrm{rad,fw}}}
    = \frac{\sigma_{\mathrm{sw}}}{\sigma_{\mathrm{fw}}}
    = \frac{5}{9.7\times10^{-3}}
    \approx 5.2\times10^{2}.
\end{equation}
This result follows directly from the same scaling principles but reflects the influence of saltwater’s greater conductivity on the radiation resistance of the antenna.

\begin{figure}[t]
    \centering
    \includegraphics[width=0.98\linewidth]{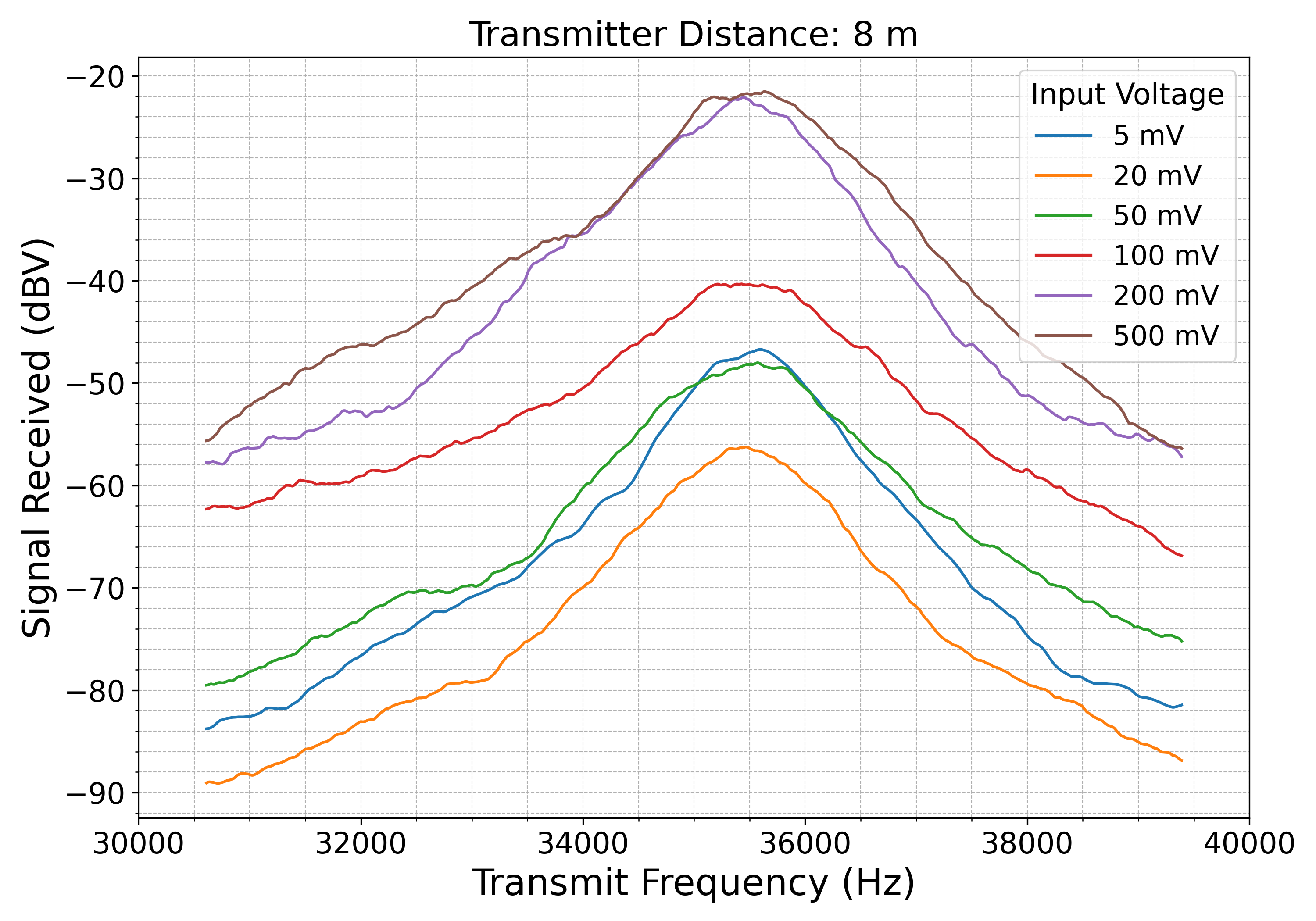}
    \caption{The received spectral responses at varying transmitter input voltages (in saltwater). The transmitter–receiver separation was fixed at \(8\,\mathrm{m}\), while the input voltage amplitude varied from \(5\,\mathrm{mV_{pp}}\) to \(500\,\mathrm{mV_{pp}}\).}
    \label{rx_power_sweep}
\end{figure}

\subsection{Power Consumption and Nonlinear Effects}
The power consumption characteristics measured during the saltwater tests (Fig.~\ref{tx_power_sweep}) were consistent with those obtained earlier in freshwater (Fig.~\ref{power_voltage}). The same antenna assemblies were used in both experiments, and the response curves exhibited comparable shapes and resonance features. Nonlinear behavior became evident at higher drive levels, with the resonance peak broadening and shifting asymmetrically toward lower frequencies. In the received-signal measurements (Fig.~\ref{rx_power_sweep}), the overall signal amplitude increased with drive strength up to approximately \(200\,\mathrm{mV_{pp}}\), beyond which additional input power did not produce proportionally higher received levels. A deviation from this trend was observed at an input of \(5\,\mathrm{mV_{pp}}\), where the peak received signal level exceeded that measured at \(20\,\mathrm{mV_{pp}}\) and \(50\,\mathrm{mV_{pp}}\).

\vspace{-1mm}
\subsection{Modulation and Practical Considerations}
\vspace{-1mm}
When an ME antenna is driven at high amplitudes, its effective stiffness becomes amplitude dependent, causing the resonance to deviate from the Lorentzian form and exhibit Duffing-type nonlinear behavior~\cite{chu_voltage-driven_2019,ivana_kovacic_duffing_2011,sadeghi_modeling_2023,zhou_nonlinear_2016}. When a modulated signal is applied, these nonlinearities generate harmonic and intermodulation distortion~\cite{xu_nonlinear_2014,shi_theoretical_2015}. Amplitude-varying modulation formats are particularly susceptible to spectral regrowth and amplitude-to-phase conversion. Constant-envelope, noncoherent formats such as MSK, CPFSK, and BFSK are therefore preferred for communication; alternative approaches such as direct antenna modulation (DAM) and nonlinear antenna modulation (NAM) have also been explored to intentionally utilize the nonlinear response within the modulation process~\cite{qiao_portable_2022,dong_acoustically_2024}.

ME antennas inherently support multiple resonant modes beyond the fundamental, corresponding to higher-order mechanical and electromechanical harmonics of the piezoelectric element~\cite{swain_engineering_2020}. In the present system, additional resonances appear above approximately 70 kHz and extend to several hundred kHz. Operation outside the fundamental band can excite these higher modes if the transmit spectrum includes harmonic components or spectral leakage from the modulation process. Maintaining spectral confinement helps suppress spurious excitation of these bands, although such measures are unnecessary when operation is restricted to a single frequency range.

The radiated field of an ME antenna originates from the time-varying magnetic flux density produced by magnetostrictive motion, which undergoes a phase transition of nearly $180^{\circ}$ through resonance~\cite{lei_theoretical_2025}. This frequency-dependent phase transition produces a corresponding frequency-dependent group delay in the radiated signal. The overall group delay of the communication link reflects the combined phase responses of the Tx and Rx antennas. For narrowband operation ($B \ll \Delta f = f_0/Q$), this delay remains effectively constant and waveform distortion is negligible, whereas for wider bandwidths ($B \sim \Delta f$), phase nonuniformity can distort the transmitted signal unless characterized and compensated at both the transmitter and receiver. Given these antenna characteristics, the desirable properties of a suitable modulation scheme are: (i) constant envelope, to minimize nonlinear conversion; (ii) spectral compactness, to avoid excitation of adjacent resonant modes; and (iii) resilience to phase dispersion, or compatibility with digital equalization schemes.





\begin{figure}[t]
    \centering
    \includegraphics[width=\linewidth]{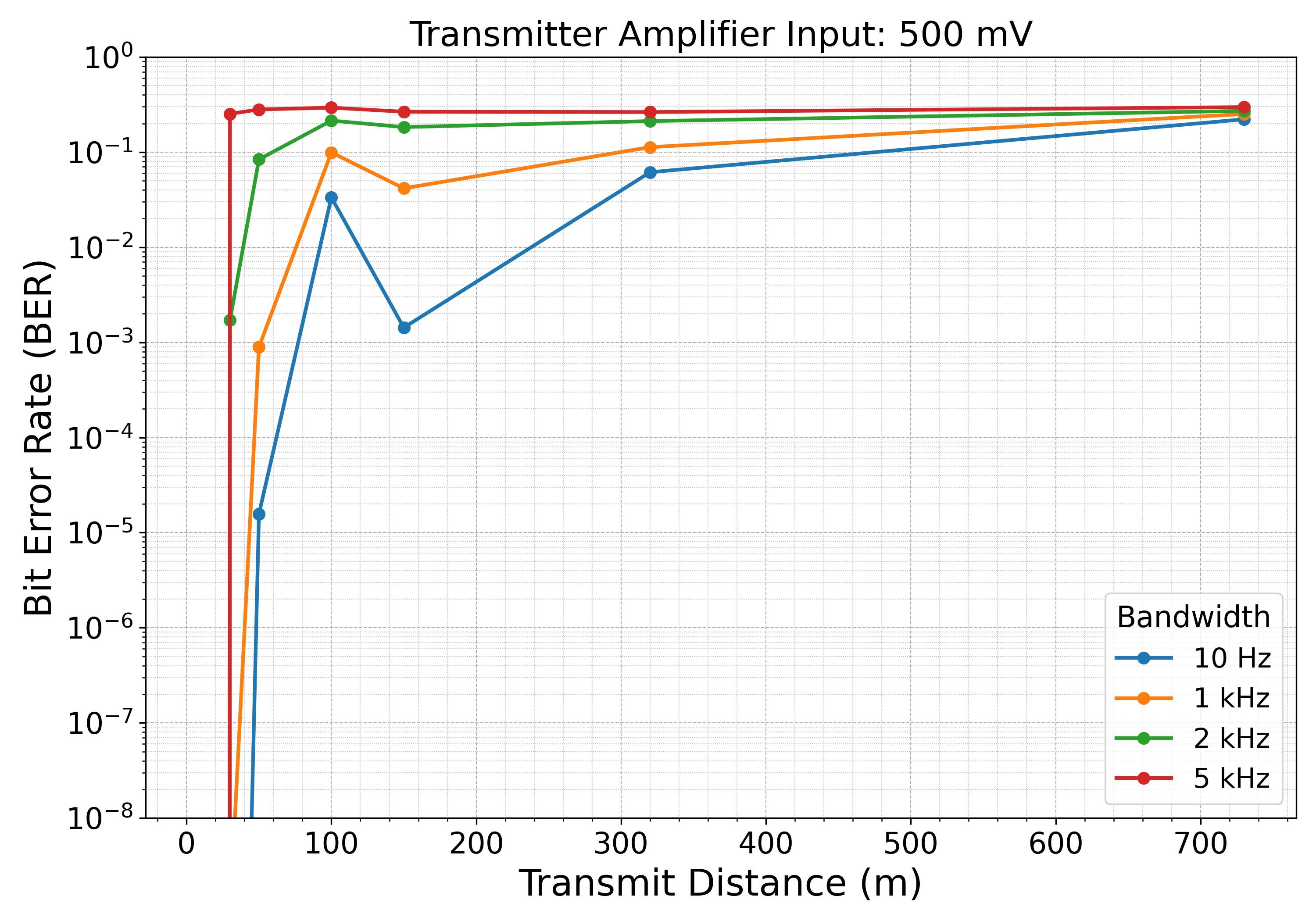}%
    \vspace{-1mm}
    \caption{Theoretical bit-error-rate (BER) as a function of distance for uncoded, noncoherent BFSK. Calculations are based on received signal data collected during the ocean trial for three selected tone spacings (bandwidths). Each case operates at its maximum possible bitrate, which is equal to the bandwidth (1 Hz = 1 bit/s). Each BFSK tone is symmetrically offset by half the corresponding bandwidth around the center frequency. The center frequency was set to 34,629.26 Hz, determined as the average of the peak-amplitude frequencies measured across all test distances.}
    \label{ber_distance_bw}%
    \vspace{-1mm}
\end{figure}

Binary frequency shift keying (BFSK) is a representative modulation scheme satisfying these requirements when operation is confined to the fundamental resonance band. In BFSK, binary symbols are conveyed by switching the carrier between two frequencies, $f_1$ and $f_2$, ideally within the resonance bandwidth. Because the envelope remains constant and the frequency deviation can be selected to stay within the array’s linear operating range, BFSK avoids AM-PM conversion and minimizes distortion. Under small-signal conditions where each antenna element operates well below its nonlinearity threshold, theoretical performance can be analyzed using the standard BER--SNR relationship.

Noncoherent BFSK compares the received energy in two frequency bands, so the decision statistic depends only on signal magnitude and is insensitive to carrier phase. This phase independence makes it well-suited to ME antennas, where the radiated and detected phase varies across the resonance bandwidth unless explicitly characterized and compensated at the transmitter and receiver. Tone separation can therefore span a large fraction of the resonant band without affecting detection. The corresponding BER in additive white Gaussian noise (AWGN)~\cite{proakis_digital_2008,simon_haykin_communication_2021,barry_digital_2004} is computed as: 
\begin{equation}
P_b = \mathrm{BER} = \tfrac{1}{2}\exp\!\left(-\tfrac{E_b}{2N_0}\right).
\end{equation}
The maximum theoretical bit rate for BFSK is determined by the minimum frequency separation required to maintain symbol orthogonality. For noncoherent detection, orthogonality is achieved when the two tones differ by an integer number of cycles over one bit interval \(T_b\), such that $\Delta f = \frac{1}{T_b} = R_b$. More generally, the energy-per-bit to noise-density ratio can be expressed in terms of the SNR, the bit rate \(R_b\), and the tone spacing \(\Delta f\) as
\begin{equation}
\frac{E_b}{N_0} = \mathrm{SNR} \times \frac{R_b}{\Delta f},
\end{equation}
where \(\mathrm{SNR}\) is taken as the minimum SNR of the two selected tones. Substituting this into the noncoherent BFSK bit-error rate equation can be reduced as follows:
\begin{equation}
P_b = \tfrac{1}{2} \exp\!\left[-\tfrac{1}{2}\frac{E_b}{N_0}\right]  = \tfrac{1}{2} \exp\!\left[-\tfrac{\mathrm{SNR}}{2}\frac{R_b}{\Delta f}\right].
\end{equation}

Given the measured SNR as a function of frequency for various transmission distances (Fig.~\ref{rx_distance_snr_sweep} and Fig.~\ref{fig:signal_distance}), the maximum achievable bit rate for a desired \(P_b\) can be estimated by selecting two tones near the resonance frequency that satisfy the BFSK tone-spacing condition and exhibit approximately equal SNR. For a chosen BER threshold, the maximum tone separation that satisfies this criterion can be used to determine \(R_b\), which can then be mapped to a range via the corresponding SNR–distance relationship. This analysis, assuming an uncoded channel, is illustrated in Fig.~\ref{ber_distance_bw}.

\begin{figure}[t]
    \centering
    \includegraphics[width=\linewidth]{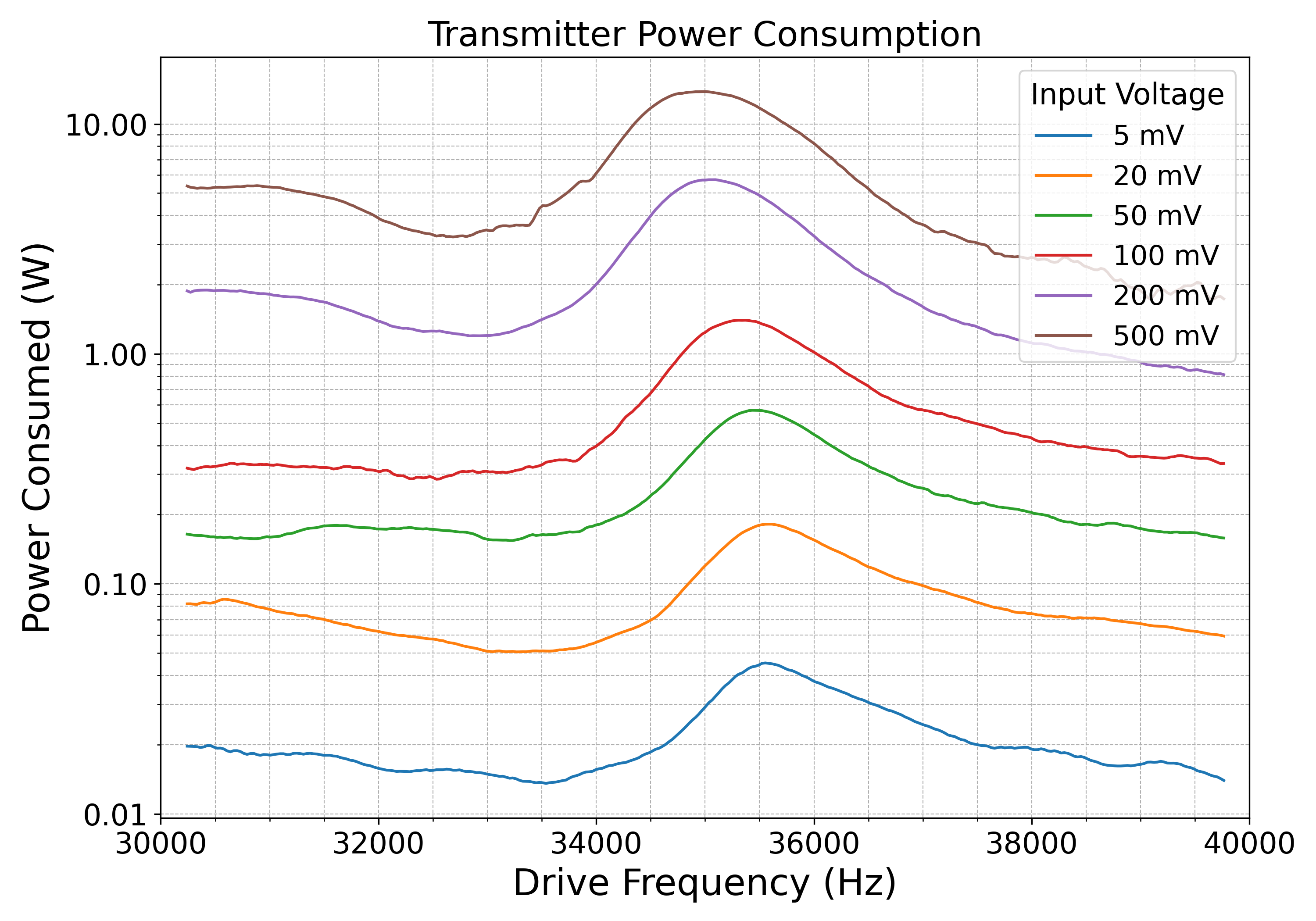}
    \caption{The measured transmitter power consumption versus frequency (in saltwater). The input voltage amplitudes varied between \(5\,\mathrm{mV_{pp}}\) and \(500\,\mathrm{mV_{pp}}\), while the frequency sweep was performed from \(30\,\mathrm{kHz}\) to \(40\,\mathrm{kHz}\).}
    \label{tx_power_sweep}
\end{figure}

\section{Analysis \& Discussion}
\vspace{-1mm}
\subsection{Comparison with Other Underwater Communication Modalities}
\vspace{-1mm}
The proposed BlueME system addresses several underwater communication challenges that commonly affect acoustic, optical, and higher-frequency radio approaches. We now provide a theoretical grounding for the observed properties, cost implications, and operational trade-offs; the summary is presented in Table~\ref{tab:comparison-table}. The tabulated data rates, costs, and ranges are derived from typical systems available in academia and industry. The listed values represent approximate order-of-magnitude maxima commonly reported for each method, serving as representative baselines for typical devices and operating conditions.

\begin{table*}[h]
\centering
\caption{A comparison of various underwater wireless communication technologies across their key performance metrics, implementation characteristics, and other practical considerations. Note that the data rates, communication ranges, and cost values represent approximate order-of-magnitude ranges reported in the literature.}
\label{tab:comparison-table}
\vspace{-1mm}
\resizebox{\textwidth}{!}{%
\begin{tabular}{|c||l|l|c|c|c|c|}
\hline
\textbf{System} &
  \multicolumn{1}{c|}{\textbf{Advantages}} &
  \multicolumn{1}{c|}{\textbf{Disadvantages}} &
  \textbf{\begin{tabular}[c]{@{}c@{}}Data Rates\end{tabular}} &
  \textbf{\begin{tabular}[c]{@{}c@{}}Range\end{tabular}} &
  \textbf{\begin{tabular}[c]{@{}c@{}}Cost (USD)\end{tabular}} &
  \textbf{Examples} \\ \hline \hline

Acoustic &
  \begin{tabular}[c]{@{}l@{}}• Unaffected by turbidity/conductivity\\ • Signals propagate along thermoclines\\ • Low attenuation loss\\ • Minimally affected by alignment\\ • Widely adopted usage\end{tabular} &
  \begin{tabular}[c]{@{}l@{}}• Sensitive to obstructions\\ • High sensitivity to doppler effects\\ • Some sensitivity to biofouling\\ • High risk to human/marine life\\ • Strong environmental interference\\ • High latency\end{tabular} &
  $\sim$10 b/s to 10 kb/s &
  $\sim$100 m to 100 km &
  $\sim$1k to 10k &
  \cite{zia_acoustic_2021,heupel_effects_2008,acoustic3,acoustic4} \\ \hline

Optical &
  \begin{tabular}[c]{@{}l@{}}• Low latency and low power footprint\\ • Lightweight construction\\ • Negligible doppler effects\\ • Widely adopted usage\end{tabular} &
  \begin{tabular}[c]{@{}l@{}}• Sensitive to turbidity and obstacles\\ • High sensitivity to biofouling\\ • Some risk to human/marine life\\ • Strong environmental interference\\ • High sensitivity to alignment\end{tabular} &
  $\sim$1 Mb/s to 1 Gb/s &
  $\sim$10 m to 100 m &
  $\sim$1k to 10k & 
  \cite{zhu_recent_2020,hanson_optical2,hydromea_luma_2024,optical1_georgiosN,optical3}\\ \hline

\begin{tabular}[c]{@{}c@{}}Magnetic\\ Field\end{tabular} &
  \begin{tabular}[c]{@{}l@{}}• Unaffected by turbidity or obstructions\\ • Can work across air-water boundary\\ • Minimal environmental interference\\ • Minimally affected by obstructions\\ • Negligible multipath/doppler effects\\ • Minimal risk to human/marine life\end{tabular} &
  \begin{tabular}[c]{@{}l@{}}• Bulky and dense construction\\ • Affected by the water conductivity\\ • High sensitivity to field alignment\end{tabular} &
  $\sim$100 kb/s to 1 Mb/s &
  $\sim$1 m to 100 m &
  $\sim$100 to 1k &
  \cite{csignum_mi_2024,wang_compact_2016,magnetic1,magnetic2,magnetic3} \\ \hline
\begin{tabular}[c]{@{}c@{}}Electric\\ Field\end{tabular} &
  \begin{tabular}[c]{@{}l@{}}• Lightweight and low cost\\ • Unaffected by turbidity\\ • Low-loss simultaneous power/data\end{tabular} &
  \begin{tabular}[c]{@{}l@{}}• Some risk to human or marine life\\ • High sensitivity to alignment\\ • Large planar area requirements\end{tabular} &
  $\sim$100 kb/s to 1 Mb/s &
  $\sim$100 cm to 1 m &
  $\sim$10 to 100 &
  \cite{da_analysis_2024,wang_bio-inspired_2017,zhao_underwater_2022,elec1,elec2,elec3} \\ \hline
\begin{tabular}[c]{@{}c@{}}Standard RF\\  (MF/HF/VHF)\end{tabular} &
  \begin{tabular}[c]{@{}l@{}}• Unaffected by turbidity\\ • Minimal environmental interference\\ • Minimally affected by obstructions\\ • Minimal risk to human/marine life\end{tabular} &
  \begin{tabular}[c]{@{}l@{}}• Strong attenuation in conductive media\\ • Prone to detuning in deployment\\ • Some sensitivity to alignment\end{tabular} &
  $\sim$100 kb/s to 1 Mb/s &
  $\sim$1 m to 10 m &
  $\sim$100 to 1k &
  \cite{teixeira_evaluation_2015,rad1_tyare_and_diego,rad2_thomas_and_mani,rad3_uribe} \\ \hline
\begin{tabular}[c]{@{}c@{}}SEW-Based RF \\ (MF/HF/VHF)\end{tabular} &
  \begin{tabular}[c]{@{}l@{}}• Unaffected by turbidity\\ • Minimal environmental interference\\ • Minimally affected by obstructions\\ • Minimal risk to human/marine life\end{tabular} &
  \begin{tabular}[c]{@{}l@{}}• Requires near-surface proximity\end{tabular} &
  \begin{tabular}[c]{@{}c@{}}$\sim$1 Mb/s to 100 Mb/s\end{tabular} &
  $\sim$10 m to 100 m&
  $\sim$100 to 1k &
  \cite{Smolyaninov2023_SciRep_SEW,Smolyaninov2024_JOE_SEW} \\ \hline
\begin{tabular}[c]{@{}c@{}}ME-Based RF\\ (VLF/LF/MF)\end{tabular} &
\begin{tabular}[c]{@{}l@{}}• Unaffected by turbidity\\ • Low latency and low power footprint\\ • Minimal environmental interference\\ • Minimally affected by obstructions\\ • Lightweight construction\\ • Negligible multipath/doppler effects\\ • Minimal risk to human/marine life\\ • Compatible with acoustic circuitry\end{tabular} &
  • Affected by water conductivity &
  $\sim$1 kb/s to 100 kb/s &
  $\sim$10 m to 1 km &
  $\sim$10 to 100 &
  \begin{tabular}[c]{@{}l@{}}\textbf{This Work} \\ and \cite{lu_acoustically_2023,du2023very} \\ 
  \cite{dong2022vlf,ME4}\end{tabular}\\ \hline
\end{tabular}%
}
\end{table*}

Acoustic links have been the preferred standard for underwater applications, offering long-range propagation up to $100$\,km with relatively modest power consumption~\cite{zia_acoustic_2021,heupel_effects_2008}. Sound waves couple efficiently into the water and travel long distances with lower attenuation than electromagnetic waves~\cite{acoustic3,acoustic4}. However, acoustic signals experience substantial multipath, Doppler shifts, and boundary reflections, especially in dynamic marine environments. These effects limit typical data throughput to the order of $10$\,b/s--$10$\,kb/s and also raise concerns about potential harm to nearby humans and marine fauna if high-intensity acoustic sources are used.

Optical methods such as laser or LED-based systems can provide data rates in excess of $1$\,Gb/s with low latency~\cite{zhu_recent_2020,hydromea_luma_2024}. However, even moderate turbidity significantly degrades optical transmissions, restricting practical operational ranges to a few tens of meters. Narrow-beam alignment requirements further complicate deployment in many natural waters where visibility can be variable~\cite{hanson_optical2,optical1_georgiosN,optical3}. 

Magnetic-field communication relies on near-field inductive coupling via magnetic coils. These methods are often employed for through-the-earth or short-to-medium-range underwater links and can operate over 1\,m to 100\,m with data rates up to about 1\,Mb/s~\cite{csignum_mi_2024,wang_compact_2016}. Magnetic fields typically penetrate obstacles effectively and offer high reliability where direct line-of-sight may be obstructed. However, large inductors or ferrite materials are often required to generate sufficient field strength, leading to heavier hardware requirements. Accurate alignment between transmitter and receiver coils is also necessary to maximize coupling efficiency \cite{magnetic1,magnetic2,magnetic3}. Typical costs for such systems are on the order of \$100 to \$1\,k USD.

Electric-field communication relies on generating and sensing electric fields in conductive media using capacitive or galvanic coupling~\cite{da_analysis_2024,wang_bio-inspired_2017,elec1,elec2,elec3}. It can be implemented with lightweight and inexpensive electrodes, potentially supporting simultaneous power transfer and data communication. Still, the typical range is limited to about $1$\,m and there are important safety considerations when using higher voltages around marine life or humans~\cite{zhao_underwater_2022}.

Standard RF methods that operate at medium to high-frequency (MF/HF/VHF) exhibit higher data rates, typically on the order of 1\,Mb/s. However, the attenuation in conductive media is much higher, which limits transmission distances to about $1$-$10$\,m~\cite{teixeira_evaluation_2015,rad1_tyare_and_diego,rad2_thomas_and_mani,rad3_uribe}. These higher-frequency systems may also require specialized antenna calibration to address de-tuning effects caused by changing dielectric properties underwater. Recent demonstrations of surface-electromagnetic-wave (SEW)-based RF communication have reported ranges exceeding 100 m and bandwidths of several megahertz by utilizing near-surface propagation along the water-air interface~\cite{Smolyaninov2023_SciRep_SEW,Smolyaninov2024_JOE_SEW}. In these implementations, electromagnetic energy is guided along the boundary layer rather than through the bulk medium, resulting in lower effective attenuation compared to volumetric propagation. This mode of operation represents a distinct configuration from conventional submerged RF systems and is therefore listed as a separate class in Table~\ref{tab:comparison-table}.

In comparison, our ME-based RF systems in the VLF/LF, or medium-frequency (MF) range rely on magnetostrictive-piezoelectric hetero-structures that convert magnetic excitation into electrical signals, and vice versa~\cite{dong2022vlf}. One of their primary advantages is immunity to turbidity and obstructions, which often hinder optical and acoustic communication. The longer electromagnetic wavelengths (on the order of tens to hundreds of meters in freshwater) reduce phase interference and Doppler effects, thereby mitigating multipath fading. Although water conductivity does lead to attenuation at increasing distances, the experimentally validated propagation ranges remain practically useful. Fabrication costs for ME-based antennas at the scales used in our experiments can be on the order of a few USD per element, making them relatively cost-effective. Power requirements are low, and the environmental impact is minimal, as these systems generate negligible acoustic noise and avoid light emissions that could disturb marine ecosystems.

Overall, the comparison in Table~\ref{tab:comparison-table} indicates that no single modality fully addresses all underwater communication requirements. Acoustic systems excel at long-range transmission but are constrained by low data rates. While higher-frequency RF and optical methods can provide larger bandwidths, they operate at severely limited ranges in conductive or turbid waterbody conditions. In contrast, ME-based RF offers a promising balance of moderate-range coverage, moderate data rates, low power consumption, and low risk to marine life. This trade-off analysis suggests that the BlueME system can occupy an attractive niche in scenarios where turbidity or multipath might compromise other modalities while remaining cost-effective and environmentally benign.

\vspace{-1mm}
\subsection{Future Potential and Practicalities}
\vspace{-1mm}
In future implementations, it may be desirable to implement custom FSK schemes for digital communication. FSK will potentially minimize the impact of amplitude-based nonlinear distortions, allowing for reliable data transmission even in the presence of nonlinearities at higher drive strengths. The ability to pack multiple antennas into arrays provides opportunities for scaling communication capabilities. Bandwidth and power can be increased by adding more antennas, and arrays can be optimized for specific operational frequencies or bandwidth requirements.

Overall, the proposed BlueME system is an attractive option for real-time underwater sensor networks and robotics applications. We envision that it could be used for close-range robot-to-robot communication as well as for coordinated planning, relative localization, and control by operating at higher frequencies with shorter wavelengths.
\section{CONCLUSION}
In this paper, we have presented the design, implementation, and experimental validation of a novel ME antenna array system, BlueME, for underwater communication in robotics applications. The BlueME system offers an alternative to traditional acoustic and optical communication methods, addressing challenges in range, efficiency, robustness, and scalability. In testing, the BlueME system achieved reliable communication at distances of over \SI{200}{\meter} with a power consumption of only \SI{1}{\watt} in fresh water and over \SI{700}{\meter} in sea water with a consumption of \SI{10}{\watt}, and indicating its suitability for low-power platforms, such as mobile robots and standalone sensors. Our results also demonstrate that the system remains unaffected by underwater noise, turbidity, and multipath interference.

Future work will focus on optimizing the antenna fabrication process to ensure consistent performance and on tailored modulation schemes that extend the system's capabilities. This includes enabling applications in multi-robot cooperative tasks such as localization, navigation, and mapping, where reliable, low-power underwater communication is essential.
\vspace{-1mm}

\section*{ACKNOWLEDGMENT}
This research is supported in part by the U.S. National Science Foundation (NSF) grants \#$2330416$ and \#$2435009$; and the University of Florida Research grant \#$132763$.

{\small
\bibliographystyle{ieeetr}
\bibliography{references} 

@article{zhou2021survey,
  author={Zhou, Ziye and Liu, Jincun and Yu, Junzhi},
  journal={IEEE/CAA Journal of Automatica Sinica}, 
  title={{A Survey of Underwater Multi-Robot Systems}}, 
  year={2022},
  volume={9},
  number={1},
  pages={1-18},
  note={DOI:10.1109/JAS.2021.1004269}
  }

@article{islam2021robot,
  title={{Robot-to-robot Relative Pose Estimation Using Humans as Markers}},
  author={Islam, Md Jahidul and Mo, Jiawei and Sattar, Junaed},
  journal={Autonomous Robots},
  volume={45},
  number={4},
  pages={579--593},
  year={2021},
  publisher={Springer},
  note={DOI:10.48550/arXiv.1903.00820}
}

@inproceedings{jawhar2020secure,
    title        = {{Secure Communication in Multi-robot Systems}},
    author       = {Jawhar, Imad and Mohamed, Nader and Al-Jaroodi, Jameela},
    year         = 2020,
    booktitle    = {2020 IEEE Systems Security Symposium (SSS)},
    pages        = {1--8},
    organization = {IEEE},
    note={DOI:10.1109/SSS47320.2020.9174264}
    }

@INPROCEEDINGS{trawny_new,
  author={Trawny, Nikolas and Roumeliotis, Stergios I. and Giannakis, Georgios B.},
  booktitle={2009 IEEE International Conference on Robotics and Automation}, 
  title={{Cooperative Multi-robot Localization Under Communication Constraints}}, 
  year={2009},
  volume={},
  number={},
  pages={4394-4400},
  note={DOI:10.1109/ROBOT.2009.5152606}
  }

@article{ghosh1992neural,
  title={{A Neural Network-based Hybrid System for Detection, Characterization, and Classification of Short-duration Oceanic Signals}},
  author={Ghosh, Joydeep and Deuser, Larry and Beck, Steven D},
  journal={IEEE Journal of Oceanic Engineering},
  volume={17},
  number={4},
  pages={351--363},
  year={1992},
  publisher={IEEE},
  note={DOI:10.1109/48.180304}
}

@article{irawan2019robot,
    title        = {{Robot Local Network Using TQS Protocol for Land-to-Underwater Communications}},
    author       = {Irawan, Addie and Abas, Mohammad Fadhil and Hasan, Nurulfadzilah},
    year         = 2019,
    journal      = {Journal of Telecommunications and Information Technology},
    number       = 1,
    pages        = {23--30},
    note={DOI:10.26636/jtit.2019.125818}
}

@article{abdullah2024ego2exo,
  title={{Ego-to-exo: Interfacing Third Person Visuals from Egocentric Views in Real-Time for Improved ROV Teleoperation}},
  author={Abdullah, Adnan and Chen, Ruo and Rekleitis, Ioannis and Islam, Md Jahidul},
  journal={arXiv preprint arXiv:2407.00848},
  year={2024},
  note={DOI:10.48550/arXiv.2407.00848}
}

@article{gu2022communication,
    title        = {{Communication and Cooperation for Spherical Underwater Robots by Using Acoustic Transmission}},
    author       = {Gu, Shuoxin and Zhang, Linshuai and Guo, Shuxiang and Zheng, Liang and An, Ruochen and Jiang, Tao and Xiong, Ai},
    year         = 2022,
    journal      = {IEEE/ASME Transactions on Mechatronics},
    publisher    = {IEEE},
    volume       = 28,
    number       = 1,
    pages        = {292--301},
    note={DOI:10.1109/TMECH.2022.3199598}
}

@inproceedings{jaafar2022overview,
    title        = {{Overview of Underwater Communication Technology}},
    author       = {Jaafar, AN and Ja'afar, H and Pasya, I and Abdullah, R and Yamada, Y},
    year         = 2022,
    booktitle    = {Proceedings of the 12th National Technical Seminar on Unmanned System Technology 2020: NUSYS'20},
    pages        = {93--104},
    organization = {Springer},
    note={DOI:10.1007/978-981-16-2406-3\_8}
}

@inproceedings{dong2022vlf,
    title        = {{VLF Mechanical Antenna Arrays for Underwater Wireless Communications}},
    author       = {Dong, Cunzheng and He, Yifan and Liu, Xiaxin and Sun, Nian Xiang},
    year         = 2022,
    booktitle    = {16th International Conference on Underwater Networks \& Systems},
    pages        = {1--5},
    note={DOI:10.1145/3567600.3568150}
}

@article{slevin2021broadband,
    title        = {{Broadband Electrically Small VLF/LF Transmitter Via Time-Varying Antenna Properties}},
    author       = {Slevin, Edward and Cohen, Morris B and Opalinski, Nathan and Thompson, Lee and Golkowski, Mark},
    year         = 2021,
    journal      = {IEEE Transactions on Antennas and Propagation},
    publisher    = {IEEE},
    volume       = 70,
    number       = 1,
    pages        = {97--110},
    note={DOI:10.1109/TAP.2021.3096950}
}

@inproceedings{kumara2021underwater,
    title        = {{Underwater Communication: A Detailed Review}},
    author       = {Kumara, Suresh and Vatsb, Chanderkant},
    year         = 2021,
    booktitle    = {CEUR Workshop Proceedings},
}

@article{moore_radio_1967,
  title={{Radio Communication in the Sea}},
  author={Moore, Richard K},
  journal={IEEE spectrum},
  volume={4},
  number={11},
  pages={42--51},
  year={1967},
  publisher={IEEE},
  note={DOI:10.1109/MSPEC.1967.5217169}
}

@ARTICLE{che_re-ev_new,
  author={{Che, Xianhui and Wells, Ian and Dickers, Gordon and Kear, Paul and Gong, Xiaochun}},
  journal={IEEE Communications Magazine}, 
  title={{Re-evaluation of RF electromagnetic communication in underwater sensor networks}}, 
  year={2010},
  volume={48},
  number={12},
  pages={143-151},
  note={DOI:10.1109/MCOM.2010.5673085}
}

@article{min2019development,
    title        = {{Development of Underwater Acoustic Communication Technology}},
    author       = {Min, ZHU and Yanbo, WU},
    year         = 2019,
    journal      = {Bulletin of Chinese Academy of Sciences},
    publisher    = {Bulletin of Chinese Academy of Sciences},
    volume       = 34,
    number       = 3,
    pages        = {289--296},
    note={DOI:10.16418/j.issn.1000-3045.2019.03.006}
}

@inproceedings{bourre2013robust,
    title        = {{A Robust OFDM Modem for Underwater Acoustic Communications}},
    author       = {Bourr{\'e}, Arnaud and Lmai, Said and Laot, Christophe and Houcke, S{\'e}bastien},
    year         = 2013,
    booktitle    = {2013 MTS/IEEE OCEANS},
    pages        = {1--5},
    organization = {IEEE},
    note={DOI:10.1109/OCEANS-Bergen.2013.6608003}
}

@inproceedings{freitag_basin-scale_2001,
    title        = {{Basin-scale Acoustic Communication: A Feasibility Study Using Tomography m-sequences}},
    shorttitle   = {Basin-scale acoustic communication},
    author       = {Freitag, L. and Stojanovic, M.},
    booktitle    = {{MTS}/{IEEE} OCEANS 2001. An Ocean Odyssey. Conference Proceedings ({IEEE} Cat. No.01CH37295)},
    volume       = 4,
    pages        = {2256--2261 vol.4},
    note          = {DOI:10.1109/oceans.2001.968349},
    urldate      = {2024-10-29},
    eventtitle   = {
        {MTS}/{IEEE} OCEANS 2001. An Ocean Odyssey. Conference Proceedings
        ({IEEE} Cat. No.01CH37295)
    },
    date         = {2001-11},
}

@inproceedings{watanabe2005design,
    title        = {{A Design of Tiny Basin Test-bed for AUV Multi Agent}},
    author       = {Watanabe, Keisuke and Nakamura, Akira},
    year         = 2005,
    booktitle    = {OCEANS},
    pages        = {1002--1008},
    organization = {MTS/IEEE},
    note={DOI:10.1109/OCEANS.2005.1639885}
}

@article{tuci2008evolving,
    title        = {{Evolving Homogeneous Neurocontrollers for a Group of Heterogeneous Robots: Coordinated Motion, Cooperation, and Acoustic Communication}},
    author       = {Tuci, Elio and Ampatzis, Christos and Vicentini, Federico and Dorigo, Marco},
    year         = 2008,
    journal      = {Artificial Life},
    publisher    = {MIT Press One Rogers Street, Cambridge, MA 02142-1209, USA journals-info~\ldots{}},
    volume       = 14,
    number       = 2,
    pages        = {157--178},
    note={DOI:10.1162/artl.2008.14.2.157}
}

@article{kilfoyle_state_2000,
  title={{The State of the Art in Underwater Acoustic Telemetry}},
  author={Kilfoyle, Daniel B and Baggeroer, Arthur B},
  journal={IEEE Journal of oceanic engineering},
  volume={25},
  number={1},
  pages={4--27},
  year={2000},
  publisher={IEEE},
  note={DOI:10.1109/48.820733}
}

@article{llor2012underwater,
    title        = {{Underwater Wireless Sensor Networks: How do Acoustic Propagation Models Impact the Performance of Higher-level Protocols?}},
    author       = {Llor, Jes{\'u}s and Malumbres, Manuel P},
    year         = 2012,
    journal      = {Sensors},
    publisher    = {Molecular Diversity Preservation International (MDPI)},
    volume       = 12,
    number       = 2,
    pages        = {1312--1335},
    note={DOI:10.1109/WAINA.2014.149}
}

@article{stojanovic_underwater_2009,
  title={{Underwater Acoustic Communication Channels: Propagation Models and Statistical Characterization}},
  author={Stojanovic, Milica and Preisig, James},
  journal={IEEE communications magazine},
  volume={47},
  number={1},
  pages={84--89},
  year={2009},
  publisher={IEEE},
  note={DOI:10.1109/MCOM.2009.4752682}
}

@article{schirripa2020underwater,
    title        = {{Underwater Optical Communications: Overview}},
    author       = {Schirripa Spagnolo, Giuseppe and Cozzella, Lorenzo and Leccese, Fabio},
    year         = 2020,
    journal      = {Sensors},
    publisher    = {Mdpi},
    volume       = 20,
    number       = 8,
    pages        = 2261,
    note={DOI:10.3390/s20082261}
}

@inproceedings{khalighi2014underwater,
    title        = {{Underwater Wireless Optical Communication; Recent Advances and Remaining Challenges}},
    author       = {Khalighi, Mohammad-Ali and Gabriel, Chadi and Hamza, Tasnim and Bourennane, Salah and Leon, Pierre and Rigaud, Vincent},
    year         = 2014,
    booktitle    = {2014 16th international conference on transparent optical networks (ICTON)},
    pages        = {1--4},
    organization = {IEEE},
    note={DOI:10.1109/ICTON.2014.6876673}
}

@article{wang2016long,
    title        = {{A Long Distance Underwater Visible Light Communication System with Single Photon Avalanche Diode}},
    author       = {Wang, Chao and Yu, Hong-Yi and Zhu, Yi-Jun},
    year         = 2016,
    journal      = {IEEE Photonics Journal},
    publisher    = {IEEE},
    volume       = 8,
    number       = 5,
    pages        = {1--11},
    note={DOI:10.1109/JPHOT.2016.2602330}
}

@article{joslin2015demonstration,
    title        = {{Demonstration of Biofouling Mitigation Methods for Long-term Deployments of Optical Cameras}},
    author       = {Joslin, James and Polagye, Brian},
    year         = 2015,
    journal      = {Marine Technology Society Journal},
    publisher    = {Marine Technology Society},
    volume       = 49,
    number       = 1,
    pages        = {88--96},
    note={DOI:10.4031/MTSJ.49.1.12}
}

@article{sun2020review,
    title        = {{A Review on Practical Considerations and Solutions in Underwater Wireless Optical Communication}},
    author       = {
        Sun, Xiaobin and Kang, Chun Hong and Kong, Meiwei and Alkhazragi, Omar
        and Guo, Yujian and Ouhssain, Mustapha and Weng, Yang and Jones, Burton
        H and Ng, Tien Khee and Ooi, Boon S
    },
    year         = 2020,
    journal      = {Journal of Lightwave Technology},
    publisher    = {IEEE},
    volume       = 38,
    number       = 2,
    pages        = {421--431},
    note={DOI:10.1109/JLT.2019.2960131}
}

@article{du2023very,
    title        = {{VLF Magnetoelectric Antennas for Portable Underwater Communication: Theory and Experiment}},
    author       = {
        Du, Yongjun and Xu, Yiwei and Wu, Jingen and Qiao, Jiacheng and Wang,
        Zhiguang and Hu, Zhongqiang and Jiang, Zhuangde and Liu, Ming
    },
    year         = 2023,
    journal      = {IEEE Transactions on Antennas and Propagation},
    publisher    = {Ieee},
    volume       = 71,
    number       = 3,
    pages        = {2167--2181},
    note={DOI:10.1109/TAP.2022.3233665}
}

@article{lu_acoustically_2023,
    title        = {{Acoustically Actuated Compact Magnetoelectric Antenna for Low-Frequency Underwater Communication}},
    author       = {L\"{u}, Xiaozhou and Chen, Xi and Zhang, Weiqiang and Gu, Long and Bao, Weimin},
    volume       = 71,
    number       = 11,
    pages        = {8493--8503},
    note          = {DOI:10.1109/tap.2023.3307711},
    issn         = {1558-2221},
    urldate      = {2024-07-11},
    journal = {{IEEE} Transactions on Antennas and Propagation},
    year={2023},
    publisher={IEEE},
    note={DOI:10.1109/TAP.2023.3307711}
}

@article{luo2024magnetoelectric,
    title        = {{Magnetoelectric Microelectromechanical and Nanoelectromechanical Systems for the IoT}},
    author       = {
        Luo, Bin and Will-Cole, AR and Dong, Cunzheng and He, Yifan and Liu,
        Xiaxin and Lin, Hwaider and Huang, Rui and Shi, Xiaoling and McConney,
        Michael and Page, Michael and others
    },
    year         = 2024,
    journal      = {Nature Reviews Electrical Engineering},
    publisher    = {Nature Publishing Group UK London},
    pages        = {1--18},
    note={DOI:10.1038/s44287-024-00044-7}
}

@article{lv2021investigation,
  title={{Investigation of Underwater Wireless Optical Communications Links with Surface Currents and Tides for Oceanic Signal Transmission}},
  author={Lv, Zhijian and He, Gui and Qiu, Chengfeng and Liu, Zhaojun},
  journal={IEEE Photonics Journal},
  volume={13},
  number={3},
  pages={1--8},
  year={2021},
  publisher={IEEE},
  note={DOI:10.1109/JPHOT.2021.3076895}
}

@article{dong2020portable,
    title        = {{
        A portable very low frequency (VLF) communication system based on
        acoustically actuated magnetoelectric antennas
    }},
    author       = {
        Dong, Cunzheng and He, Yifan and Li, Menghui and Tu, Cheng and Chu,
        Zhaoqiang and Liang, Xianfeng and Chen, Huaihao and Wei, Yuyi and
        Zaeimbashi, Mohsen and Wang, Xinjun and others
    },
    year         = 2020,
    journal      = {IEEE Antennas and Wireless Propagation Letters},
    publisher    = {IEEE},
    volume       = 19,
    number       = 3,
    pages        = {398--402},
    note={DOI:10.1109/LAWP.2020.2968604}
}

@inproceedings{lin_future_2018,
	title = {{Future Antenna Miniaturization Mechanism: Magnetoelectric Antennas}},
	url = {https://ieeexplore.ieee.org/document/8439678},
	note={DOI:10.1109/MWSYM.2018.8439678},
	shorttitle = {Future Antenna Miniaturization Mechanism},
	eventtitle = {2018 {IEEE}/{MTT}-S International Microwave Symposium - {IMS}},
	pages = {220--223},
	booktitle = {2018 {IEEE}/{MTT}-S International Microwave Symposium - {IMS}},
	author = {Lin, Hwaider and Zaeimbashi, Mohsen and Sun, Neville and Liang, Xianfeng and Chen, Huaihao and Dong, Cunzheng and Matyushov, Alexei and Wang, Xinjun and Guo, Yingxue and Gao, Yuan and Sun, Nian-Xiang},
	date = {2018-06},
	note = {{ISSN}: 2576-7216},
}

@article{xu2019low,
    title        = {{A Low Frequency Mechanical Transmitter Based on Magnetoelectric Heterostructures Operated at their Resonance Frequency}},
    author       = {
        Xu, Junran and Leung, Chung Ming and Zhuang, Xin and Li, Jiefang and
        Bhardwaj, Shubhendu and Volakis, John and Viehland, Dwight
    },
    year         = 2019,
    journal      = {Sensors},
    publisher    = {Mdpi},
    volume       = 19,
    number       = 4,
    pages        = 853,
    note={DOI:10.3390/s19040853}
}

@article{xiang2016subsea,
    title        = {{Subsea Cable Tracking by Autonomous Underwater Vehicle with Magnetic Sensing Guidance}},
    author       = {Xiang, Xianbo and Yu, Caoyang and Niu, Zemin and Zhang, Qin},
    year         = 2016,
    journal      = {Sensors},
    publisher    = {Mdpi},
    volume       = 16,
    number       = 8,
    pages        = 1335,
    note={DOI:10.3390/s16081335}
}

@article{angara2024performance,
  title={{Performance Assessment of Underwater-to-air Optical Wireless Communication System with the Effect of Solar Noise and Sea Surface Conditions}},
  author={Angara, Bhogeswara Rao and Shanmugam, Palanisamy and Ramachandran, Harishankar and Sandhani, Chavapati Gouse},
  journal={IEEE Access},
  year={2024},
  publisher={IEEE},
  note={DOI:10.1109/ACCESS.2024.3409424}
}

@article{stojanovic1994phase,
  title={{Phase-coherent Digital Communications for Underwater Acoustic Channels}},
  author={Stojanovic, Milica and Catipovic, Josko A and Proakis, John G},
  journal={IEEE journal of oceanic engineering},
  volume={19},
  number={1},
  pages={100--111},
  year={1994},
  publisher={IEEE},
  note={DOI:10.1109/48.289455}
}

@inproceedings{hao2023survery,
  author={Hao, Lijun and Ao, Jun and Ma, Chunbo},
  booktitle={2023 Cross Strait Radio Science and Wireless Technology Conference (CSRSWTC)}, 
  title={{Survery on Underwater Wireless Communication Technology}}, 
  year={2023},
  volume={},
  number={},
  pages={1-3},
  note={DOI:10.1109/CSRSWTC60855.2023.10426965}
}

@article{headrick2009growth,
    title        = {{Growth of Underwater Communication Technology in the US Navy}},
    author       = {Headrick, Robert and Freitag, Lee},
    year         = 2009,
    journal      = {IEEE Communications Magazine},
    publisher    = {IEEE},
    volume       = 47,
    number       = 1,
    pages        = {80--82},
    note={DOI:10.1109/MCOM.2009.4752681}
}

@inproceedings{pompili2006deployment,
    title        = {{Deployment Analysis in Underwater Acoustic Wireless Sensor Networks}},
    author       = {Pompili, Dario and Melodia, Tommaso and Akyildiz, Ian F},
    year         = 2006,
    booktitle    = {1st ACM International Workshop on Underwater Networks},
    pages        = {48--55},
    note={DOI:10.1145/1161039.1161050}
}

@article{song2019editorial,
  title={{Editorial Underwater Acoustic Communications: Where we Stand and What is Next?}},
  author={Song, Aijun and Stojanovic, Milica and Chitre, Mandar},
  journal={IEEE Journal of Oceanic Engineering},
  volume={44},
  number={1},
  year={2019},
  note={DOI:10.1109/JOE.2018.2883872}
}

@article{stojanovic1996recent,
  title={{Recent Advances in High-speed Underwater Acoustic Communications}},
  author={Stojanovic, Milica},
  journal={IEEE Journal of Oceanic engineering},
  volume={21},
  number={2},
  pages={125--136},
  year={1996},
  publisher={IEEE},
  note={DOI:10.1109/48.486787}
}

@misc{BlueComm,
    title        = {{BlueComm 200}},
    author       = {Sonadyne},
    year         = 2016,
    note         = {Accessed: 22-2-2024},
    howpublished = {\url{https://www.sonardyne.com/products/bluecomm-200-wireless-underwater-link/}},
}

@misc{Luma,
    title        = {{Luma\texttrademark~Modems}},
    author       = {Hydromea},
    year         = 2020,
    note         = {Accessed: 22-2-2024},
    howpublished = {\url{https://www.hydromea.com/underwater-wireless-communication}},
}

@inproceedings{gilbert1966underwater,
    title        = {{Underwater Experiments on the Polarization, Coherence, and Scattering Properties of a Pulsed Blue-Green Laser}},
    author       = {Gilbert, GD and Stoner, TR and Jernigan, JL},
    year         = 1966,
    booktitle    = {Underwater Photo Optics I},
    volume       = 7,
    pages        = {8--14},
    organization = {Spie},
    note={DOI:10.1117/12.971001}
}

@inproceedings{neuner2019multi,
    title        = {{Multi-laser Transmissometer for Ocean Optical Classification and Biofouling Detection}},
    author       = {
        Neuner, Burton and Wang, Andrew and Zlatanovic, Sanja and Jennings, Dan
        and Tran, Nghia and Wiedemeier, Brandon
    },
    year         = 2019,
    booktitle    = {OCEANS},
    pages        = {1--5},
    organization = {IEEE},
    note={DOI:10.1109/OCEANSE.2019.8867257}
}

@article{wang_bio-inspired_2017,
	title = {{A bio-inspired electrocommunication system for small underwater robots}},
	volume = {12},
	issn = {1748-3190},
	url = {https://dx.doi.org/10.1088/1748-3190/aa61c3},
	note = {DOI:10.1088/1748-3190/aa61c3},
	pages = {036002},
	number = {3},
	journaltitle = {Bioinspiration \& Biomimetics},
	shortjournal = {Bioinspir. Biomim.},
	author = {Wang, Wei and Liu, Jindong and Xie, Guangming and Wen, Li and Zhang, Jianwei},
	urldate = {2024-12-24},
	date = {2017-03},
	langid = {english},
	publisher = {{IOP} Publishing}
}

@inproceedings{abdullah2023caveseg,
    title        = {
        {CaveSeg: Deep Semantic Segmentation and Scene Parsing for Autonomous Underwater Cave Exploration}
    },
    author       = {
        Abdullah, Adnan and Barua, Titon and Tibbetts, Reagan and Chen, Zijie
        and Islam, Md Jahidul and Rekleitis, Ioannis
    },
    year         = 2024,
    booktitle    = {IEEE International Conference on Robotics and Automation (ICRA)},
    organization = {IEEE},
}

@article{wang_seawater_2019,
    title        = {{Seawater Short-Range Electromagnetic Wave Communication Method Based on {OFDM} Subcarrier Allocation}},
    author       = {Wang, Jun and Wang, Shilian},
    volume       = 7,
    number       = 10,
    pages        = {63--71},
    note          = {DOI:10.4236/jcc.2019.710006},
    urldate      = {2024-10-10},
    publisher         = {Scientific Research Publishing},
    rights       = {http://creativecommons.org/licenses/by/4.0/},
    journaltitle = {Journal of Computer and Communications},
    date         = {2019-10-10},
    langid       = {english},
}

@article{Wang2024,
    title        = {{Minitype Arrays of Acoustically Actuated Magnetoelectric Antennas for Magnetic Induction Communication}},
    author       = {Shiyu Wang and Gaoqi Dou and Guangming Song},
    year         = 2024,
    journal      = {Actuators},
    volume       = 13,
    number       = 8,
    pages        = 276,
    note          = {DOI:10.3390/act13080276}
}

@article{Fu2023,
    title        = {{Bias-Free Very Low Frequency Magnetoelectric Antenna}},
    author       = {
        Shifeng Fu and Jiawei Cheng and Tao Jiang and Hanzhou Wu and Ze Fang
        and Jie Jiao and Oleg Sokolov and Sergey Ivanov and Mirza Bichurin and
        Yaojin Wang
    },
    year         = 2023,
    journal      = {Applied Physics Letters},
    volume       = 122,
    number       = 26,
    pages        = 262901,
    note          = {DOI:10.1063/5.0158020},
    url          = {https://doi.org/10.1063/5.0158020}
}

@article{mukherjee2023miniaturized,
    title        = {{A Miniaturized, Low-Frequency Magnetoelectric Wireless Power Transfer System for Powering Biomedical Implants}},
    author       = {
        Mukherjee, Dibyajyoti and Rainu, Simran Kaur and Singh, Neetu and
        Mallick, Dhiman
    },
    year         = 2023,
    journal      = {IEEE Transactions on Biomedical Circuits and Systems},
    publisher    = {IEEE},
    note={DOI:10.1109/TBCAS.2023.3336598}
}

@article{Wu2021,
  title={{Local Charge Density Wave and Metal-insulator Transition in Ba (K) Bi (Pb) O3}},
  author={Manh, D Nguyen and Mayou, D and Cyrot-Lackmann, F},
  journal={Physica C: Superconductivity},
  volume={185},
  pages={1611--1612},
  year={1991},
  publisher={Elsevier},
  note={DOI:10.1016/0921-4534(91)90932-O}
}

@article{yang2021progress,
    title        = {{Progress on Very/Ultra Low Frequency Mechanical Antennas}},
    author       = {
        Yang, Shaolong and Xu, Jianchun and Guo, Menghao and Zhang, Bokun and
        Lan, Chuwen and Li, Haihong and Bi, Ke
    },
    year         = 2021,
    journal      = {ES Materials \& Manufacturing},
    publisher    = {Engineered Science Publisher},
    volume       = 16,
    pages        = {1--12},
    note={DOI:10.30919/esmm5f497}
}

@article{Fang2024,
    title        = {{Crafting Very Low Fequency Magnetoelectric Antenna via Piezoelectric and Electromechanical Synergic Optimization Strategy}},
    author       = {
        Fang, Ze and Jiao, Jie and Wu, Hanzhou and Jiang, Tao and Fu, Shifeng
        and Cheng, Jiawei and Oleg, Sokolov and Sergey, Ivanov and and Mirza,
        Bichurin and Li, Fei and Wang, Yaojin
    },
    year         = 2024,
    journal      = {Journal of Materiomics},
    volume       = {Volume Number},
    number       = {Issue Number},
    pages        = {Page Range},
    note          = {DOI:10.1016/j.jmat.2024.05.010}
}

@article{Leung2024,
    title        = {{Self-Biased Magneto-Electric Antenna for Very-Low-Frequency Communications: Exploiting Magnetization Grading and Asymmetric Structure-Induced Resonance}},
    author       = {
        Chung Ming Leung and Haoran Zheng and Jing Yang and Tao Wang and Feifei
        Wang
    },
    year         = 2024,
    journal      = {Sensors},
    volume       = 24,
    number       = 2,
    pages        = 694,
    note          = {DOI:10.3390/s24020694}
}

@article{Popov2008,
    title        = {{Direct and Converse Magnetoelectric Effect at Resonant Frequency in Laminar Piezoelectric-Magnetostrictive Composite}},
    author       = {
        C. Popov and H. Chang and P. M. Record and E. Abraham and R. W.
        Whatmore and Z. Huang
    },
    year         = 2008,
    journal      = {Journal of Electroceramics},
    volume       = 20,
    number       = 1,
    pages        = {53--58},
    note          = {DOI:10.1007/s10832-007-9184-5},
}

@misc{metglas_inc_2605sa1,
    title        = {{{METGLAS}\textregistered{} 2605SA1 and 2605HB1M Alloy General Properties and Characteristics}},
    author       = {Metglas, Inc.},
    howpublished          = {\url{https://metglas.com/magnetic-materials/}},
    note      = {Accessed: 2024-10-30},
}

@misc{mide_technology_pzt5j,
    title        = {{Piezo Material Properties}},
    author       = {Mide Technology},
    howpublished          = {\url{https://support.piezo.com/article/62-material-properties}},
    note      = {Accessed: 2024-10-30}
}

@article{Hosur2021,
    title        = {{A Comprehensive Study on Magnetoelectric Transducers for Wireless Power Transfer Using Low-Frequency Magnetic Fields}},
    author       = {
        Sujay Hosur and Ram Sriramdas and Sumanta Karan and Na Liu and Shashank
        Priya and Mehdi Kiani
    },
    year         = 2021,
    journal      = {IEEE Transactions on Biomedical Circuits and Systems},
    volume       = 15,
    number       = 5,
    pages        = {1079--1092},
    note          = {DOI:10.1109/tbcas.2021.3108637},
}

@article{Karan2024,
    title        = {{Magnetic Field and Ultrasound Induced Simultaneous Wireless Energy Harvesting}},
    author       = {
        Sumanta Kumar Karan and Sujay Hosur and Zeinab Kashani and Haoyang Leng
        and Anitha Vijay and Rammohan Sriramdas and Kai Wang and Bed Poudel and
        Andrew Patterson and Mehdi Kiani and Shashank Priya
    },
    year         = 2024,
    journal      = {Energy and Environmental Science},
    volume       = 17,
    pages        = {2129--2144},
    note          = {DOI:10.1039/d3ee03889k},
}

@misc{Comsol,
    title        = {{COMSOL - Software for Multiphysics Simulation}},
    author       = {COMSOL, Inc.},
    year         = 2016,
    note         = {Accessed: 22-8-2024},
    howpublished = {\url{https://www.comsol.com}},
}

@article{dong_analysis_2023,
    title        = {{Analysis of {Near} {Field} {Mutual} {Coupling} in {Wideband} {Magnetoelectric} {Antennas} {Array}}},
    author       = {
        Dong, Biao and Yan, Zhongming and Zhang, Yong and Han, Tianhao and
        Zhou, Hongcheng and Wang, Yu
    },
    year         = 2023,
    journal      = {Journal of Applied Physics},
    volume       = 134,
    number       = 11,
    pages        = 114103,
    issn         = {0021-8979},
    urldate      = {2024-02-29},
    note={DOI:10.1063/5.0166407}
}

@article{smolyaninov_surface_2021,
	title = {{Surface Electromagnetic Waves at Gradual Interfaces Between Lossy Media}},
	volume = {170},
	issn = {1559-8985},
	url = {http://arxiv.org/abs/2102.08828},
	note={DOI:10.2528/PIER21043006},
	pages = {177--186},
	journaltitle = {Progress In Electromagnetics Research},
	shortjournal = {{PIER}},
	author = {Smolyaninov, Igor I.},
	date = {2021},
	eprinttype = {arxiv},
	eprint = {2102.08828 [physics]},
}

@article{smolyaninov_development_2020,
    title        = {{Development of Broadband Underwater Radio Communication for Application in Unmanned Underwater Vehicles}},
    author       = {Smolyaninov, Igor and Balzano, Quirino and Young, Dendy},
    volume       = 8,
    number       = 5,
    pages        = 370,
    note          = {DOI:10.3390/jmse8050370},
    issn         = {2077-1312},
    url          = {https://www.mdpi.com/2077-1312/8/5/370},
    urldate      = {2024-10-31},
    publisher         = {Multidisciplinary Digital Publishing Institute},
    rights       = {http://creativecommons.org/licenses/by/3.0/},
    journaltitle = {Journal of Marine Science and Engineering},
    date         = {2020-05},
    langid       = {english},
}

@misc{university_of_florida_alachua_2022,
    title        = {{Alachua {LAKEWATCH} Report 2023}},
    author       = {University of Florida},
    urldate      = {2024-10-30},
    note         = {2022-12-09},
    howpublished ={\url{https://lakewatch.ifas.ufl.edu/media/lakewatchifasufledu/reports/lake-reports/Lake-Report-2023-Alachua.pdf}},
}

@article{burnside2023axial,
    title        = {{An Axial Mode Magnetoelectric Antenna: Radiation Predictions via Multiphysics Modeling with Experimental Validations}},
    author       = {
        Burnside, Emily A and Tiwari, Sidhant and Burnside, Scott R and
        Candler, Robert N and Henderson, Rashaunda and Grimm, Schaffer and
        Carman, Gregory P
    },
    year         = 2023,
    journal      = {Journal of Applied Physics},
    publisher    = {AIP Publishing},
    volume       = 134,
    number       = 14,
    note={DOI:10.1063/5.0171973}
}

@article{chu_multilayered_2023,
    title        = {{A Multilayered Magnetoelectric Transmitter with Suppressed Nonlinearity for Portable {VLF} Communication}},
    author       = {
        Chu, Zhaoqiang and Mao, Zhineng and Song, Kaixin and Jiang, Shizhan and
        Min, Shugang and Dan, Wei and Yu, Chenyuan and Wu, Meiyu and Ren,
        Yinghui and Lu, Zhichao and Jiao, Jie and Nan, Tianxiang and Dong,
        Shuxiang
    },
    volume       = 6,
    pages        = {0208},
    note          = {DOI: 10.34133/research.0208},
    urldate      = {2024-02-10},
    publisher         = {American Association for the Advancement of Science},
    journaltitle = {Research},
    date         = {2023-09-15},
}

@article{de2024shannon,
  title={{Shannon--Hartley Channel Capacity for Underwater Wireless Optical Communications}},
  author={De Freitas, Jolyon M},
  journal={ACS Photonics},
  volume={11},
  number={3},
  pages={866--873},
  year={2024},
  publisher={ACS Publications},
  note={DOI:10.1021/acsphotonics.3c00843}
}

@misc{mercado_fsucml_2025,
	title = {{FSUCML} Real-time and Continuous Seawater Monitoring System: Annual Report 2024 - 2025},
	url = {https://marinelab.fsu.edu/research/seawater-monitoring-system/},
	publisher = {{FSU} Coastal and Marine Laboratory},
	author = {Mejia-Mercado, Beatriz and Trexler, Joel and Morris, Cullen},
	date = {2025},
	langid = {english},
}

@article{smolyaninov_superlensing_2023,
	title = {{Superlensing Enables Radio Communication and Imaging Underwater}},
	volume = {13},
	rights = {2023 The Author(s)},
	issn = {2045-2322},
	url = {https://www.nature.com/articles/s41598-023-45663-6},
	note={DOI:10.1038/s41598-023-45663-6},
	pages = {18333},
	number = {1},
	journaltitle = {Scientific Reports},
	shortjournal = {Sci Rep},
	author = {Smolyaninov, Igor I. and Balzano, Quirino and Barry, Mark and Young, Dendy},
	date = {2023-10-26},
	langid = {english},
	publisher = {Publisher: Nature Publishing Group},
}

@article{chu_voltage-driven_2019,
	title = {{Voltage-Driven Nonlinearity in Magnetoelectric Heterostructures}},
	volume = {12},
	url = {https://link.aps.org/doi/10.1103/PhysRevApplied.12.044001},
	note={DOI:10.1103/PhysRevApplied.12.044001},
	pages = {044001},
	number = {4},
	journaltitle = {Physical Review Applied},
	shortjournal = {Phys. Rev. Appl.},
	author = {Chu, Zhaoqiang and Dong, Cunzheng and Tu, Cheng and He, Yifan and Liang, Xianfeng and Wang, Jiawei and Wei, Yuyi and Chen, Huaihao and Gao, Xiangyu and Lu, Caijiang and Zhu, Zengtai and Lin, Yuanhua and Dong, Shuxiang and {McCord}, Jeffrey and Sun, Nian-Xiang},
	date = {2019-10-01},
	publisher = {Publisher: American Physical Society}
}

@collection{ivana_kovacic_duffing_2011,
	edition = {1},
	title = {{The Duffing Equation: Nonlinear Oscillators and their Behaviour}},
	isbn = {978-0-470-97785-9},
	url = {https://onlinelibrary.wiley.com/doi/10.1002/9780470977859},
	publisher = {John Wiley \& Sons, Ltd},
	editor = {{Ivana Kovacic} and {Michael J. Brennan}},
	date = {2011},
	langid = {english},
	note={DOI:10.1002/9780470977859},
	note = {https://onlinelibrary.wiley.com/doi/pdf/10.1002/9780470977859},
}

@article{sadeghi_modeling_2023,
	title = {{Modeling of Magnetoelectric Microresonator Using Numerical Method and Simulated Annealing Algorithm}},
	volume = {14},
	rights = {http://creativecommons.org/licenses/by/3.0/},
	issn = {2072-666X},
	url = {https://www.mdpi.com/2072-666X/14/10/1878},
	note={DOI:10.3390/mi14101878},
	pages = {1878},
	number = {10},
	journaltitle = {Micromachines},
	author = {Sadeghi, Mohammad and Bazrafkan, Mohammad M. and Rutner, Marcus and Faupel, Franz},
	date = {2023-10},
	langid = {english},
	publisher = {Multidisciplinary Digital Publishing Institute},
}

@article{zhou_nonlinear_2016,
	title = {{Nonlinear Resonant Magnetoelectric Coupling Model for Dual-Peak Phenomenon in Magnetoelectric Laminates}},
	volume = {672},
	issn = {0925-8388},
	url = {https://www.sciencedirect.com/science/article/pii/S0925838816304170},
	note={DOI:10.1016/j.jallcom.2016.02.150},
	pages = {292--297},
	journaltitle = {Journal of Alloys and Compounds},
	shortjournal = {Journal of Alloys and Compounds},
	author = {Zhou, Hao-Miao and Li, Meng-Han and Zhou, Yun and Chen, Qing},
	date = {2016-07-05},
}

@article{xu_nonlinear_2014,
	title = {{Nonlinear Harmonic Distortion Effect in Magnetoelectric Laminate Composites}},
	volume = {105},
	issn = {0003-6951},
	url = {https://doi.org/10.1063/1.4887373},
	note={DOI:10.1063/1.4887373},
	pages = {012904},
	number = {1},
	journaltitle = {Applied Physics Letters},
	shortjournal = {Appl. Phys. Lett.},
	author = {Xu, Hao and Pei, Yongmao and Fang, Daining and Wang, Panding},
	date = {2014-07-08},
}

@article{shi_theoretical_2015,
	title = {{Theoretical Study on Nonlinear Magnetoelectric Effect and Harmonic Distortion Behavior in Laminated Composite}},
	volume = {646},
	issn = {0925-8388},
	url = {https://www.sciencedirect.com/science/article/pii/S0925838815300013},
	note={DOI:10.1016/j.jallcom.2015.05.229},
	pages = {351--359},
	journaltitle = {Journal of Alloys and Compounds},
	shortjournal = {Journal of Alloys and Compounds},
	author = {Shi, Yang and Gao, Yuan-Wen},
	date = {2015-10-15},
}

@inproceedings{qiao_portable_2022,
	title = {{A Portable {VLF} Magnetoelectric Antenna with High Communication Rate Based on Direct Antenna Amplitude Modulation}},
	url = {https://ieeexplore.ieee.org/document/10106834},
	note={DOI:10.1109/IMWS-AMP54652.2022.10106834},
	eventtitle = {2022 {IEEE} {MTT}-S International Microwave Workshop Series on Advanced Materials and Processes for {RF} and {THz} Applications ({IMWS}-{AMP})},
	pages = {1--3},
	booktitle = {2022 {IEEE} {MTT}-S International Microwave Workshop Series on Advanced Materials and Processes for {RF} and {THz} Applications ({IMWS}-{AMP})},
	author = {Qiao, Jiacheng and Wu, Jingen and Du, Yongjun and Xu, Yiwei and Hu, Zhongqiang and Liu, Ming},
	date = {2022-11},
	note = {{ISSN}: 2694-2992},
}

@thesis{dong_acoustically_2024,
	title = {{Acoustically Actuated Magnetoelectric Antennas for {VLF} Communication and Magnetic Sensing}},
	url = {https://hdl.handle.net/2047/D20670383},
	institution = {Northeastern University},
	type = {phdthesis},
	author = {Dong, Cunzheng},
	date = {2024},
	langid = {english},
	note={DOI:10.17760/D20670383},
}

@article{swain_engineering_2020,
	title = {{Engineering Resonance Modes for Enhanced Magnetoelectric Coupling in Bilayer Laminate Composites for Energy Harvesting Applications}},
	volume = {13},
	url = {https://link.aps.org/doi/10.1103/PhysRevApplied.13.024026},
	note={DOI:10.1103/PhysRevApplied.13.024026},
	pages = {024026},
	number = {2},
	journaltitle = {Physical Review Applied},
	shortjournal = {Phys. Rev. Appl.},
	author = {Swain, Atal Bihari and Dinesh Kumar, S. and Subramanian, Venkatachalam and Murugavel, Pattukkannu},
	date = {2020-02-11},
	publisher = {American Physical Society}
}

@article{lei_theoretical_2025,
	title = {{Theoretical Analysis on Radiation Performance of Magnetoelectric Antennas Considering Eddy-Current Loss and Quasistatic Hysteresis Effect}},
	volume = {138},
	issn = {0021-8979},
	url = {https://doi.org/10.1063/5.0282234},
	note={DOI:10.1063/5.0282234},
	pages = {044101},
	number = {4},
	journaltitle = {Journal of Applied Physics},
	shortjournal = {J. Appl. Phys.},
	author = {Lei, Baoxin and Li, Yan and Xu, Guokai and Gao, Changlun and Pang, Kun and Ye, Dagui and Shi, Yang and Xiao, Shaoqiu},
	date = {2025-07-22},
}

@book{proakis_digital_2008,
	location = {Boston, Mass.},
	edition = {5. ed},
	title = {{Digital Communications}},
	isbn = {978-0-07-295716-7},
	pagetotal = {1150},
	publisher = {{McGraw}-Hill},
	author = {Proakis, John G. and Salehi, Masoud},
	date = {2008},
	langid = {english},
}

@book{simon_haykin_communication_2021,
	edition = {5},
	title = {{Communication Systems, 5th Edition}},
	isbn = {978-1-118-07296-7},
	url = {https://www.wiley.com/en-us/Communication+Systems%2C+5th+Edition-p-9781118072967R150},
	publisher = {John Wiley \& Sons, Ltd},
	author = {{Simon Haykin} and {Michael Moher}},
	date = {2021-01},
	langid = {english},
}

@book{barry_digital_2004,
	location = {Boston, {MA}},
	edition = {3},
	title = {{Digital Communication}},
	rights = {http://www.springer.com/tdm},
	isbn = {978-1-4613-4975-4 978-1-4615-0227-2},
	url = {http://link.springer.com/10.1007/978-1-4615-0227-2},
	publisher = {Springer {US}},
	author = {Barry, John R. and Lee, Edward A. and Messerschmitt, David G.},
	date = {2004},
	langid = {english},
	note={DOI:10.1007/978-1-4615-0227-2},
}

@article{zia_acoustic_2021,
	title = {{State-of-the-Art Underwater Acoustic Communication Modems: Classifications, Analyses and Design Challenges}},
	volume = {116},
	issn = {1572-834X},
	url = {https://doi.org/10.1007/s11277-020-07431-x},
	note={DOI:10.1007/s11277-020-07431-x},
	shorttitle = {State-of-the-Art Underwater Acoustic Communication Modems},
	pages = {1325--1360},
	number = {2},
	journaltitle = {Wireless Personal Communications},
	shortjournal = {Wireless Pers Commun},
	author = {Zia, Muhammad Yousuf Irfan and Poncela, Javier and Otero, Pablo},
	urldate = {2024-12-24},
	date = {2021-01-01},
	langid = {english},
}

@article{heupel_effects_2008,
	title = {{Effects of Biofouling on Performance of Moored Data Logging Acoustic Receivers}},
	volume = {6},
	rights = {© 2008, by the Association for the Sciences of Limnology and Oceanography, Inc.},
	issn = {1541-5856},
	url = {https://onlinelibrary.wiley.com/doi/abs/10.4319/lom.2008.6.327},
	note={DOI:10.4319/lom.2008.6.327},
	pages = {327--335},
	number = {7},
	journaltitle = {Limnology and Oceanography: Methods},
	author = {Heupel, Michelle R. and Reiss, Katie L. and Yeiser, Beau G. and Simpfendorfer, Colin A.},
	urldate = {2024-12-23},
	date = {2008},
	langid = {english},
}

@INPROCEEDINGS{acoustic3,
  author={Farr, N. and Bowen, A. and Ware, J. and Pontbriand, C. and Tivey, M.},
  booktitle={OCEANS}, 
  title={{An Integrated, Underwater Optical/Acoustic Communications System}}, 
  year={2010},
  volume={},
  number={},
  pages={1-6},
  note={DOI:10.1109/OCEANSSYD.2010.5603510}
}

@ARTICLE{acoustic4,
  author={Luo, Yu and Pu, Lina and Zuba, Michael and Peng, Zheng and Cui, Jun-Hong},
  journal={IEEE Transactions on Emerging Topics in Computing}, 
  title={{Challenges and Opportunities of Underwater Cognitive Acoustic Networks}}, 
  year={2014},
  volume={2},
  number={2},
  pages={198-211},
  note={DOI:10.1109/TETC.2014.2310457}
}

@article{zhu_recent_2020,
title = {{Recent Progress in and Perspectives of Underwater Wireless Optical Communication}},
journal = {Progress in Quantum Electronics},
volume = {73},
pages = {100274},
year = {2020},
issn = {0079-6727},
note={DOI:10.1016/j.pquantelec.2020.100274},
url = {https://www.sciencedirect.com/science/article/pii/S0079672720300288},
author = {Shijie Zhu and Xinwei Chen and Xiaoyan Liu and Guoqi Zhang and Pengfei Tian}
}

@misc{hydromea_luma_2024,
    title        = {{Luma\texttrademark~Modems}},
    author       = {Hydromea},
    year         = 2020,
    note         = {Accessed: 22-10-2024},
    howpublished = {\url{https://www.hydromea.com/underwater-wireless-communication}},
}

@article{hanson_optical2,
    title        = {{High Bandwidth Underwater Optical Communication}},
    author       = {Hanson, Frank and Radic, Stojan},
    year         = 2008,
    journal      = {Applied optics},
    publisher    = {Optica Publishing Group},
    volume       = 47,
    number       = 2,
    pages        = {277--283},
    note={DOI:10.1364/ao.47.000277}
}

@ARTICLE{optical1_georgiosN,
  author={Arvanitakis, Georgios N. and Bian, Rui and McKendry, Jonathan J. D. and Cheng, Chen and Xie, Enyuan and He, Xiangyu and Yang, Gang and Islim, Mohamed S. and Purwita, Ardimas A. and Gu, Erdan and Haas, Harald and Dawson, Martin D.},
  journal={IEEE Photonics Journal}, 
  title={{Gb/s Underwater Wireless Optical Communications Using Series-Connected GaN Micro-LED Arrays}}, 
  year={2020},
  volume={12},
  number={2},
  pages={1-10},
  note={DOI:10.1109/JPHOT.2019.2959656}
}

@inproceedings{optical3,
author = {Judith Bannon Snow and James P. Flatley and Dennis E. Freeman and Mark A. Landry and Carl E. Lindstrom and Jacob R. Longacre and Joshua A. Schwartz},
title = {{Underwater Propagation of High-Data-Rate Laser Communications Pulses}},
volume = {1750},
booktitle = {Ocean Optics XI},
editor = {Gary D. Gilbert},
organization = {International Society for Optics and Photonics},
publisher = {SPIE},
pages = {419 -- 427},
year = {1992},
note={DOI:10.1117/12.140670},
URL = {https://doi.org/10.1117/12.140670}
}

@misc{csignum_mi_2024,
    title = {{CSignum - Underwater Communications}},
    author = {CSignum},
    year         = 2020,
    note         = {Accessed: 22-2-2024},
    howpublished = {\url{https://www.csignum.com/}},
}

@inproceedings{wang_compact_2016,
	title = {{A Compact Low-Power Underwater Magneto-Inductive Modem}},
	isbn = {978-1-4503-4637-5},
	url = {https://dl.acm.org/doi/10.1145/2999504.3001064},
	note={DOI:10.1145/2999504.3001064},
	pages = {1--5},
	booktitle = {Proceedings of the 11th International Conference on Underwater Networks \& Systems (WUWNet'16)},
	publisher = {Association for Computing Machinery},
	author = {Wang, Yibin and Dobbin, Andrew and Bousquet, Jean-François},
	urldate = {2024-12-24},
	date = {2016-10-24},
}

@ARTICLE{magnetic1,
  author={Li, Yuzhou and Wang, Shengnan and Jin, Cheng and Zhang, Yu and Jiang, Tao},
  journal={IEEE Communications Surveys \& Tutorials}, 
  title={{A Survey of Underwater Magnetic Induction Communications: Fundamental Issues, Recent Advances, and Challenges}}, 
  year={2019},
  volume={21},
  number={3},
  pages={2466-2487},
  note={DOI:10.1109/COMST.2019.2897610}
}

@INPROCEEDINGS{magnetic2,
  title={{Contribution of Surface Wave to Horizontal Magnetic Dipole in Three-layered Media}},
  author={Zhang, Lichen and Tang, Jiansheng and Yu, Fujian and Chen, Hongyu},
  booktitle={2017 IEEE International Conference on Signal Processing, Communications and Computing (ICSPCC)},
  pages={1--6},
  year={2017},
  organization={IEEE},
  note={DOI:10.1109/ICSPCC.2017.8242435},
}

@ARTICLE{magnetic3,
  title={{Multiple Frequency Band Channel Modeling and Analysis for Magnetic Induction Communication in Practical Underwater Environments}},
  author={Guo, Hongzhi and Sun, Zhi and Wang, Pu},
  journal={IEEE Transactions on Vehicular Technology},
  volume={66},
  number={8},
  pages={6619--6632},
  year={2017},
  publisher={IEEE},
  note={DOI:10.1109/TVT.2017.2664099},
}

@article{da_analysis_2024,
	title = {{Analysis and Implementation of Underwater Single Capacitive Coupled Simultaneous Wireless Power and Bidirectional Data Transfer System}},
	volume = {71},
	issn = {1557-9948},
	url = {https://ieeexplore.ieee.org/abstract/document/10550917},
	note={DOI:10.1109/TIE.2024.3395764},
	pages = {15674--15684},
	number = {12},
	journal = {{IEEE} Transactions on Industrial Electronics},
	author = {Da, Chaolai and Li, Fang and Wang, Lifang and Tao, Chengxuan and Li, Shufan and Nie, Ming},
    year={2024},
    publisher={IEEE}
}

@InProceedings{elec1,
  title={{Design and Verification of Underwater Electric Field Communication System Based Spread Spectrum Techniques}},
  author={Lu, Tao and Hu, Qiao and Xu, Dan and Feng, Xinglong and Zhang, Yuzhong},
  booktitle={International Conference on Autonomous Unmanned Systems},
  pages={1421--1431},
  year={2022},
  organization={Springer},
  note={DOI:10.1063/1.4992839}
}

@INPROCEEDINGS{elec2,
  author={Momma, H. and Tsuchiya, T.},
  booktitle={OCEANS'76}, 
  title={{Underwater Communication by Electric Current}}, 
  year={1976},
  volume={1},
  number={1},
  pages={631--636},
  note={DOI:10.1109/OCEANS.1976.1154306}
}

@inproceedings{elec3,
author = {Esemann, Tim and Ardelt, Gunther and Hellbr\"{u}ck, Horst},
title = {{Underwater Electric Field Communication}},
year = {2014},
isbn = {9781450332774},
publisher = {Association for Computing Machinery},
url = {https://doi.org/10.1145/2671490.2674561},
note={DOI:10.1145/2671490.2674561},
booktitle = {Proceedings of the 9th International Conference on Underwater Networks \& Systems (WUWNet'14)},
articleno = {9},
numpages = {5},
location = {Rome, Italy}
}

@article{zhao_underwater_2022,
	title = {{Underwater wireless communication via {TENG}-generated Maxwell’s displacement current}},
	volume = {13},
	rights = {2022 The Author(s)},
	issn = {2041-1723},
	url = {https://www.nature.com/articles/s41467-022-31042-8},
	note={DOI:10.1038/s41467-022-31042-8},
	pages = {3325},
	number = {1},
	journaltitle = {Nature Communications},
	shortjournal = {Nat Commun},
	author = {Zhao, Hongfa and Xu, Minyi and Shu, Mingrui and An, Jie and Ding, Wenbo and Liu, Xiangyu and Wang, Siyuan and Zhao, Cong and Yu, Hongyong and Wang, Hao and Wang, Chuan and Fu, Xianping and Pan, Xinxiang and Xie, Guangming and Wang, Zhong Lin},
	urldate = {2024-12-24},
	date = {2022-06-09},
	langid = {english},
	publisher = {Nature Publishing Group},
}

@inproceedings{teixeira_evaluation_2015,
	title = {{Evaluation of Underwater {IEEE} 802.11 Networks at {VHF} and {UHF} Frequency Bands using Software Defined Radios}},
	isbn = {978-1-4503-4036-6},
	url = {https://dl.acm.org/doi/10.1145/2831296.2831313},
	note={DOI:10.1145/2831296.2831313},
	pages = {1--5},
	booktitle = {Proceedings of the 10th International Conference on Underwater Networks \& Systems (WUWNet'15)},
	publisher = {Association for Computing Machinery},
	author = {Teixeira, Filipe and Santos, José and Pessoa, Luís and Pereira, Mário and Campos, Rui and Ricardo, Manuel},
	urldate = {2024-12-23},
	date = {2015-10-22},
}

@article{rad1_tyare_and_diego,
  author={Vilches, Tyare and Dujovne, Diego},
  journal={IEEE Network}, 
  title={{GNUradio and 802.11: Performance Evaluation and Limitations}}, 
  year={2014},
  volume={28},
  number={5},
  pages={27-31},
  note={DOI:10.1109/MNET.2014.6915436}
}

@inproceedings{rad2_thomas_and_mani,
author = {Schmid, Thomas and Sekkat, Oussama and Srivastava, Mani B.},
title = {{An Experimental Study of Network Performance Impact of Increased Latency in Software Defined Radios}},
year = {2007},
isbn = {9781595937384},
publisher = {Association for Computing Machinery},
url = {https://doi.org/10.1145/1287767.1287779},
note={DOI:10.1145/1287767.1287779},
booktitle = {Proceedings of the Second ACM International Workshop on Wireless Network Testbeds, Experimental Evaluation and Characterization},
pages = {59–66},
numpages = {8},
location = {Montreal, Quebec, Canada},
series = {WinTECH '07}
}

@INPROCEEDINGS{rad3_uribe,
  author={Uribe, Carlos and Grote, Walter},
  booktitle={2009 3rd International Conference on New Technologies, Mobility and Security}, 
  title={{Radio Communication Model for Underwater WSN}}, 
  year={2009},
  volume={},
  number={},
  pages={1-5},
  note={DOI:10.1109/NTMS.2009.5384789}
}

@article{Smolyaninov2023_SciRep_SEW,
  author       = {Igor I. Smolyaninov and Quirino Balzano and Mark Barry and Dendy Young},
  title        = {{Superlensing Enables Radio Communication and Imaging Underwater}},
  journal      = {Scientific Reports},
  year         = {2023},
  volume       = {13},
  articleno    = {18333},
  pages        = {18333},
  note          = {DOI:10.1038/s41598-023-45663-6}
}

@article{Smolyaninov2024_JOE_SEW,
  author       = {Igor I. Smolyaninov and Quirino Balzano and Mark Barry},
  title        = {{Transmission of High-Definition Video Signals and Detection of the Objects Underwater Using Surface Electromagnetic Waves}},
  journal      = {IEEE Journal of Oceanic Engineering},
  year         = {2024},
  volume       = {49},
  number       = {2},
  pages        = {566--571},
  note          = {DOI:10.1109/JOE.2023.3335599}
}

@article{ME4,
author = {Wang, Shi-Yu and Dou, Gao-Qi and Yi, Da and Feng, Shi-Min and Tang, Ming-Chun},
title = {{Reconfigurable Super-Low-Frequency Magnetoelectric Antenna for Underwater Frequency-Hopping Communication}},
journal = {IET Microwaves, Antennas \& Propagation},
volume = {18},
number = {12},
pages = {911-916},
note={DOI:10.1049/mia2.12523},
url = {https://ietresearch.onlinelibrary.wiley.com/doi/abs/10.1049/mia2.12523},
eprint = {https://ietresearch.onlinelibrary.wiley.com/doi/pdf/10.1049/mia2.12523},
year = {2024}
}
}

\end{document}